\documentclass[conference]{IEEEtran}
\ifCLASSINFOpdf
\else
\fi
\DeclareMathAlphabet{\mathcal}{OMS}{cmsy}{m}{n}
\usepackage{cite}
\usepackage{amsmath,amssymb,amsfonts}
\usepackage{textcomp}
\usepackage{xcolor}
\usepackage{graphicx}
\usepackage{subcaption}
\usepackage{mathtools}

\DeclarePairedDelimiter\floor{\lfloor}{\rfloor}
\usepackage{empheq}
\usepackage{cases}
\usepackage{algorithm}
\usepackage[noend]{algpseudocode}
\usepackage{verbatim}
\usepackage{multirow}
\usepackage{booktabs}
\usepackage{tikz}
\usepackage{xcolor}
\usepackage{xspace}
\usepackage{tikz}
\usepackage{placeins}
\usepackage{hyperref}
\usepackage{sidecap}
\sidecaptionvpos{figure}{c}

\usepackage{threeparttable}

\usepackage{tabularx}
\usepackage{siunitx}
\usepackage{tikz}
\newcommand*{\priority}[1]{\begin{tikzpicture}[scale=0.11]%
    \draw (0,0) circle (1);
    \fill[fill opacity=0.5,fill=blue] (0,0) -- (90:1) arc (90:90-#1*3.6:1) -- cycle;
    \end{tikzpicture}}

\algrenewcommand\algorithmicfor{\textbf{for}} 
\algrenewcommand\algorithmicend{\textbf{end}} 

\newcommand{\norm}[1]{\displaystyle \left\| #1 \right\|}
\newcommand*\circled[1]{\tikz[baseline=(char.base)]{\node[shape=circle,fill,inner sep=1.1pt] (char) {\textcolor{white}{#1}};}}

\begin{document}
%
\title{\Large \bf RoVISQ: Reduction of Video Service Quality via Adversarial Attacks on \\ Deep Learning-based Video Compression
\vspace{-0.6cm}}

\def\mojan{\textcolor{black}}
\def\seira{\textcolor{black}}
\def\edit{\textcolor{black}}
\def\cameraready{\textcolor{black}}
\def\jungwoo{\textcolor{black}}
\newcommand{\rdo}{RD attack\xspace}
\newcommand{\rdos}{RD attacks\xspace}




%
 
\author{
\parbox{\linewidth}{\centering
Jung-Woo Chang$^*$ \quad 
Mojan Javaheripi$^*$ \quad 
Seira Hidano$^\dagger$ \quad 
Farinaz Koushanfar$^*$ \\
$^*$University of California San Diego \quad $^\dagger$KDDI Research, Inc. \\
\textcolor{black}{
\{\href{mailto:juc023@ucsd.edu}{juc023}, \href{mailto:mojan@ucsd.edu}{mojan}, \href{mailto:farinaz@ucsd.edu}{farinaz}\}@ucsd.edu, \href{mailto:se-hidano@kddi.com}{se-hidano}@kddi.com
}}\vspace{+0.6cm}}


\IEEEoverridecommandlockouts
\makeatletter\def\@IEEEpubidpullup{6.5\baselineskip}\makeatother
\IEEEpubid{\parbox{\columnwidth}{
    Network and Distributed System Security (NDSS) Symposium 2023\\
    28 February - 4 March 2023, San Diego, CA, USA\\
    ISBN 1-891562-83-5\\
    https://dx.doi.org/10.14722/ndss.2023.23165\\
    www.ndss-symposium.org}
    \hspace{\columnsep}\makebox[\columnwidth]{}
}

\maketitle

\begin{abstract}
Video compression plays a crucial role in video streaming and classification systems by maximizing the end-user quality of experience (QoE) at a given bandwidth budget. 
In this paper, we conduct the first systematic study for adversarial attacks on deep learning-based video compression and downstream classification systems. 
Our attack framework, dubbed RoVISQ, manipulates the Rate-Distortion (\textit{R}-\textit{D}) relationship of a video compression model to achieve one or both of the following goals: (1)~increasing the network bandwidth, (2)~degrading the video quality for end-users. We further devise new objectives for targeted and untargeted attacks to a downstream video classification service. Finally, we design an input-invariant perturbation that universally disrupts video compression and classification systems in real time.
Unlike previously proposed attacks on video classification, our adversarial perturbations are the first to withstand compression. 
We empirically show the resilience of RoVISQ attacks against various defenses, i.e., adversarial training, video denoising, and JPEG compression.  
Our extensive experimental results on various video datasets show RoVISQ attacks 
deteriorate peak signal-to-noise ratio by up to 5.6dB and the bit-rate by up to $\sim$ 2.4$\times$ while achieving over 90$\%$ attack success rate on a downstream classifier. \edit{Our user study further demonstrates the effect of RoVISQ attacks on users' QoE. We provide several example attacked videos used in our survey on \url{https://sites.google.com/view/demo-of-rovisq/home}.}

\end{abstract}

\section{Introduction}
\label{sec:intro}
\begin{figure}[t]
\centering
    \begin{subfigure}[b]{\columnwidth}
    \centering
    \includegraphics[width=0.92\linewidth]{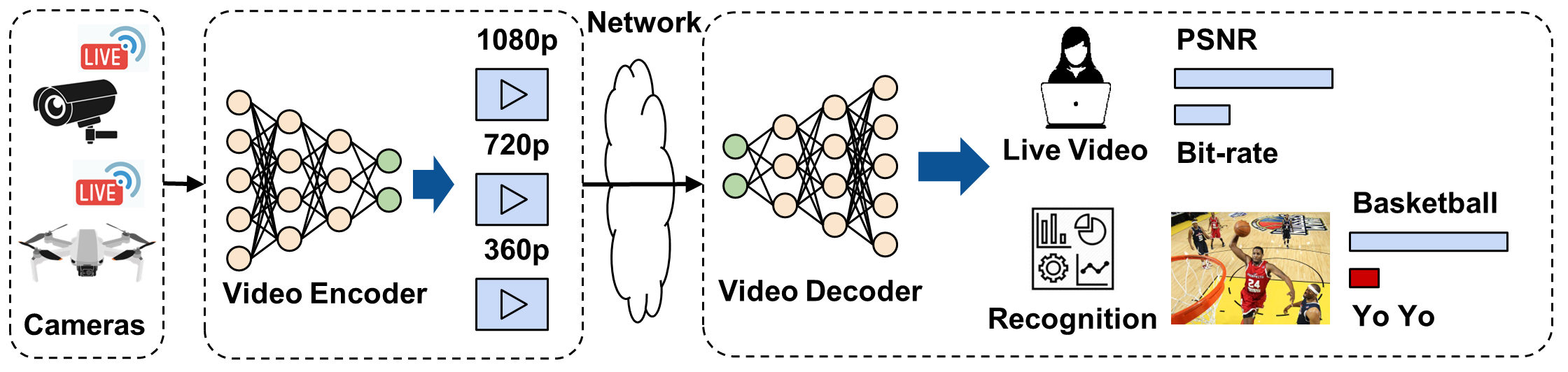}
    \vspace{-0.2cm}
    \caption{Benign}
    \end{subfigure}
    \vspace{-0.3cm}
    \begin{subfigure}[b]{\columnwidth}
    \includegraphics[width=\linewidth]{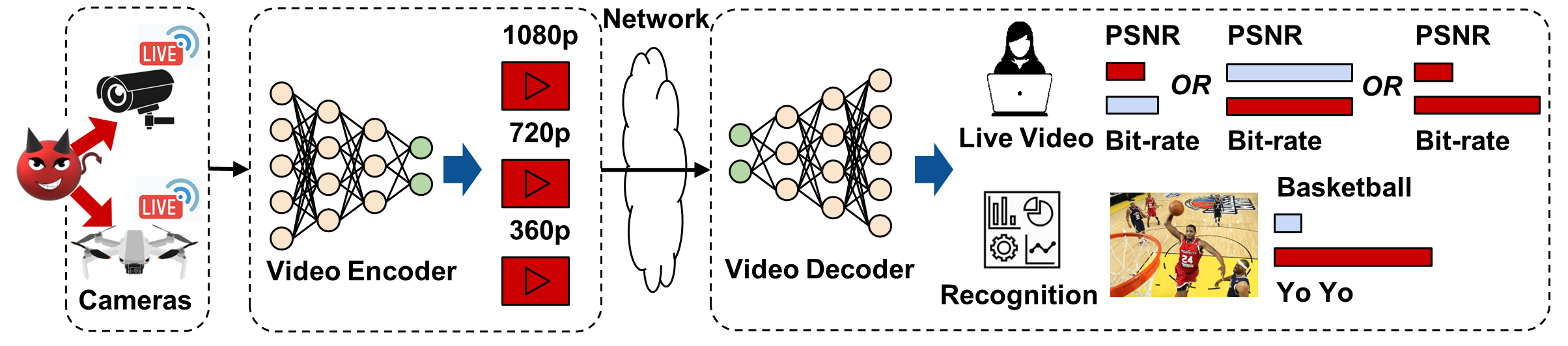}
    \caption{Attacked}
    \end{subfigure}
    \vspace{-0.5cm}
  \caption{High-level view of (a) a live video streaming and classification system, and (b) RoVISQ attacks performed by injecting adversarial perturbations to the front-end video sources. To demonstrate the adversarial effect, we present the peak signal-to-noise ratio (PSNR), bit-rate, and an example outcome of video activity recognition.}
  \label{fig:intro}
 \vspace{-0.7cm}
\end{figure}

Video content accounts for more than 80\% of the internet traffic \cite{Cisco_vc}. Live video traffic has experienced an even higher growth with the advent of streaming services, e.g., remote vehicle control~\cite{lu2020edge}, remote surgery~\cite{hassan2019high}, and UHD streaming~\cite{YouTube,Twitch}. In addition, recent developments in deep learning have given rise to live video analysis services such as activity recognition for health care diagnosis~\cite{healthcare}, video surveillance~\cite{surveillance}, and autonomous driving~\cite{autonomous}. Video compression technologies are a key enabler for the aforesaid media streaming and content analysis applications.
As shown in Figure~\ref{fig:intro}-(a), live video streaming and classification systems typically consist of four main components, i.e., front-end video sources (cameras), video encoder, video decoder, and back-end video subscribers. The video encoder and decoder compress the video for efficient real-time streaming to back-end subscribers. Recent literature proposes Deep Neural Network (DNN) based encoder and decoder for video compression~\cite{Lu_2019_CVPR,Hu_2021_CVPR,Yang_2020_CVPR}. Due to their enhanced performance over conventional codecs~\cite{H.264,H.265}, DNNs are now adopted in the next generation of video compression technologies by the Moving Picture Experts Group (MPEG) standardization group~\cite{Mpeg} and industry~\cite{Qualcomm}.

Signal transformation techniques, e.g., compression, have been trusted to remove the adversarial effect in image recognition~\cite{Jia_2019_CVPR, aydemir2018effects,8953640,8416586}. We empirically show that the state-of-the-art adversarial instances created for video classification~\cite{Wei_aaai_2019,Li_NIPS_2021,Li_2019_NDSS} can also be removed after video compression. Nevertheless, the robustness of the DNN-based video compression pipeline itself against adversarial perturbations remains largely unknown to date. 

In this work, we present RoVISQ, \underline{R}eduction \underline{o}f \underline{Vi}deo \underline{S}ervice \underline{Q}uality, as a new threat to video compression. We show, for the first time, that DNN-based video compression systems can be exploited by adversaries to not only reduce the users' quality of experience (QoE), but also attack downstream video recognition services. The success of RoVISQ attacks is indebted to solving a novel unified optimization problem. The solution crafts adversarial perturbations that simultaneously attack the video compression and the downstream video classification system as shown in Figure~\ref{fig:intro}-(b).


\noindent\underline{\textbf{Proposed Attacks.}} The goal of video compression is to simultaneously minimize the number of bits used for encoding while reducing the distortion. Contemporary DNN-based video compression frameworks~\cite{Lu_2019_CVPR,Yang_2020_CVPR,Hu_2021_CVPR} typically employ a \textit{R}-\textit{D} optimization~\cite{sullivan1998rate, ortega1998rate} that minimizes the distortion (\textit{D}) at a given bit-rate ($R \leq R_{T}$) where $R_{T}$ is the available bit-rate budget. In this paper, we aim to break the \textit{R}-\textit{D} model of a pre-trained video compression model. 

\begin{SCfigure}[50][t]
\centering
\includegraphics[width=0.45\linewidth]{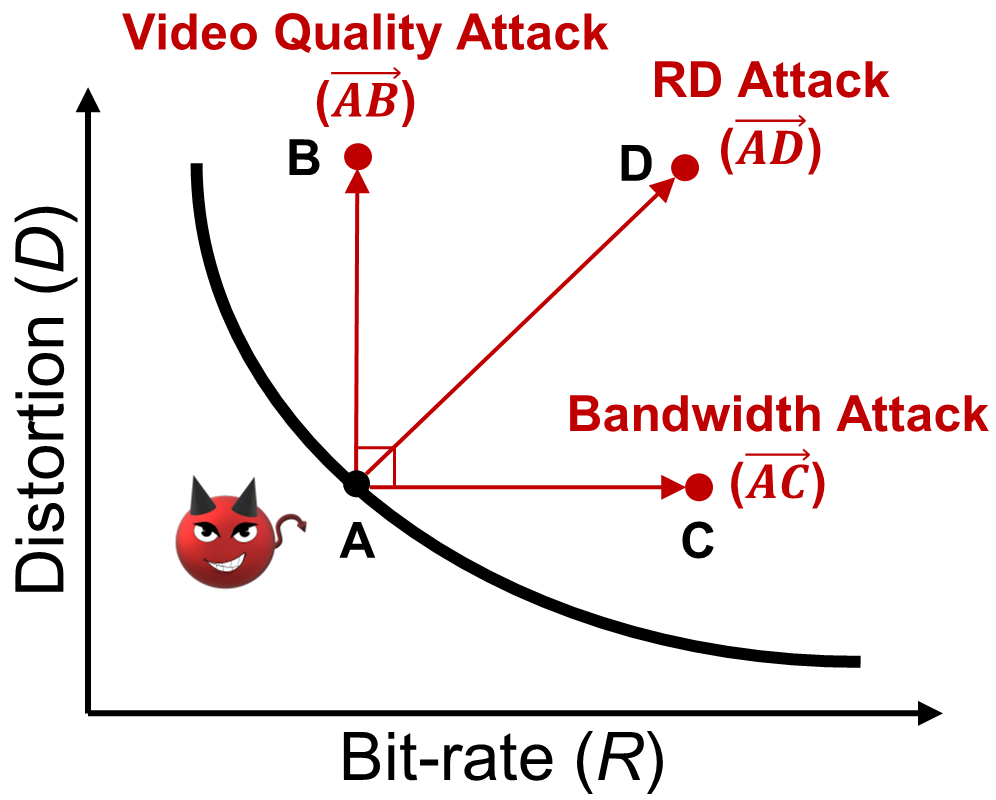}
\vspace{-0.2cm}
\caption{Effect of RoVISQ on the \textit{R}-\textit{D} curve shown for video quality attack($\protect\overrightarrow{AB}$), bandwidth attack($\protect\overrightarrow{AC}$), and \rdo ($\protect\overrightarrow{AD}$).}
\label{fig:intro_rd}
\vspace{-0.4cm}
\end{SCfigure}

We propose three novel attacks on video compression shown in Figure~\ref{fig:intro_rd}: \emph{video quality attack}, \emph{bandwidth attack}, and \emph{\rdo}. The video quality attack increases \textit{D} while keeping \textit{R} constant, whereas the bandwidth attack reduces compression rate (increases \textit{R}) while maintaining \textit{D}. Our \rdo simultaneously increases \textit{R} and \textit{D}. The attacker can choose between aforesaid strategies based on the adversarial goal and characteristics of the video streaming service.
We also propose a \emph{compression-robust classifier attack} that causes a downstream video recognition service to misclassify all actions in the input video. Notably, ours is the first attack on video classification that remains successful when the perturbed video goes through various signal transformations, i.e., video compression, denoising, and frame-by-frame JPEG compression.

RoVISQ attacks are applicable to both online and offline settings. In the offline scenario, videos are locally stored in a camera or smartphone and the attack latency is not constrained. For the challenging online scenario, we preprocess universal perturbations that can attack any video stream in real time. \edit{We assume the attacker has access to open-source video compression/classification models that can be used as a surrogate to train the universal perturbation.}

\noindent\underline{\textbf{Design Challenges and Solutions.}} 
We are faced with several new design challenges not present in prior works that attack only video classification~\cite{Li_2019_NDSS, Pony_2021_CVPR, Li_NIPS_2021, Wei_aaai_2019, Xie_SP_2022}.
Firstly, video compression systems, by design, are extremely robust to noise. State-of-the-art video compression schemes~\cite{Lu_2019_CVPR, Hu_2021_CVPR, Yang_2020_CVPR} group a series of frames into sequences called Group of Pictures (GOP) to allow back-end users to access live video streams at any time. The GOP structure inherently improves robustness against error propagation and sudden scene changes during compression, thereby making the adversarial attack more challenging. We address this problem by modeling the temporal coding structure of the GOP, which is then incorporated in our attack formulation. We validate the effectiveness of our attacks on three types of GOP structures~\cite{Lu_2019_CVPR, Yang_2020_CVPR,Hu_2021_CVPR}.

Secondly, previous attacks~\cite{Li_2019_NDSS, Pony_2021_CVPR, Li_NIPS_2021, Wei_aaai_2019, Xie_SP_2022} on video recognition do not consider video compression in their threat models. Recent work shows that conventional compression algorithms can be used as a defense against adversarial attacks to classification~\cite{Jia_2019_CVPR, aydemir2018effects,8953640,8416586}. Specifically, lossy compression with quantization reduces the spatiotemporal data, thereby easily removing the adversarial perturbations. To provide a compression-resistant perturbation for video recognition, we consider the video compression in our adversarial attack. In the video compression pipeline, the reconstruction error is propagated to the next frame during encoding. Consequently, we design a new model to train the perturbations by using the information from different time steps and combining all information to break the original \textit{R}-\textit{D} optimization. 


Finally, the adversary does not have prior knowledge about the content of the streaming video. While launching the online attacks, the perturbation should be effectively injected at any time in the video encoder. Additionally, different video compression frameworks support various encoding parameters, i.e., compression ratio and GOP structure~\cite{Lu_2019_CVPR, Hu_2021_CVPR, Yang_2020_CVPR}. However, the real time constraints of the online attack prohibit the adversary from customizing the injected noise for the underlying encoding parameters. To tackle the above challenges, we generate a universal adversarial perturbation that is trained via a novel offline process. \edit{Specifically, our training adjusts the universal perturbation to be applicable to various input videos, encoding parameters, and video compression/classification models. We further utilize a temporal transformation function during training which enables us to perform the attack in any order of the video sequence. Notably, our universal perturbations are not only effective for the surrogate video compression/classification models that they are trained on (white-box attack), but they can also degrade the performance of unseen video compression/classification models (black-box attack).}

In summary, our key contributions are as follows:
\begin{itemize}
\setlength\itemsep{0.3em}
\item We perform the first systematic study of adversarial attacks on deep learning-based video compression. We further maliciously exploit the video compression pipeline to direct our attacks towards downstream video recognition systems.

\item We propose four new adversarial attacks that result in high-impact security and QoE consequences. We formalize RoVISQ attacks as well-defined optimization problems that can be solved to obtain perturbations affecting the \textit{R}, \text{D}, and downstream classification. 

\item We construct a well-designed universal perturbation that is invariant to the underlying DNN model, encoding parameters, and input videos. Our universal perturbation is reusable for various video encoders and can be injected at any time during live video streaming and classification. This enables us to conduct an online attack in real time under strict latency constraints.

\item Our adversarial attacks are the first to withstand the inherent denoising performed during video compression. Additionally, we show the resiliency of RoVISQ attacks against various defenses namely, adversarial training, video denoising, and JPEG compression.

\end{itemize}

\section{Preliminaries}
\label{sec:system}

\begin{figure*}[t]
\centering
\includegraphics[width=0.9\textwidth]{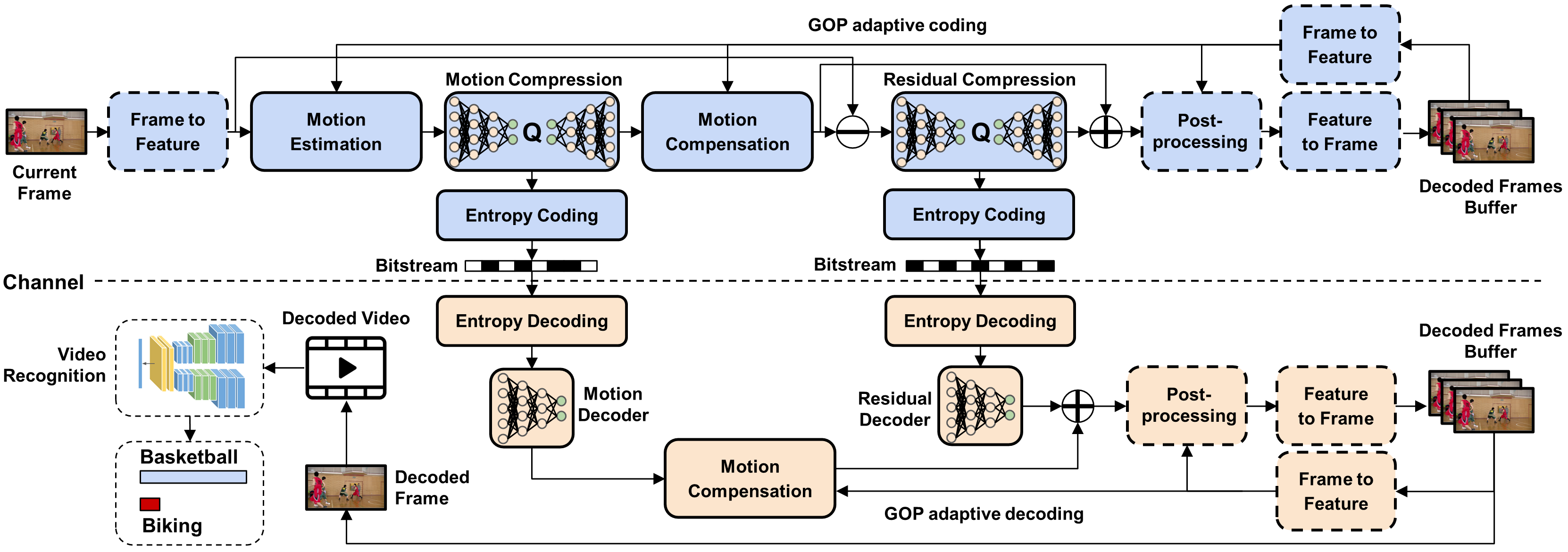}
\caption{Overview of DNN-based video compression and classification frameworks. Blue (resp. orange) color modules compose the video encoder (resp. decoder). Modules with dashed boundaries are used selectively according to the video compression model.}
\label{fig:videocompression}
\vspace{-0.5cm}
\end{figure*}

Figure~\ref{fig:videocompression} illustrates the \edit{commonly used modules in state-of-the-art} DNN-based video compression and classification frameworks. Video compression employs temporal prediction to minimize the difference between the previously decoded frame and the current frame. The encoder sends bitstreams that conform to a specific channel standard to the decoder. The compressed video is then reconstructed from the bitstreams by the decoder. State-of-the-art video compression models can be grouped into two classes, namely, pixel-space and feature-space frameworks. Pixel-space video compression~\cite{Lu_2019_CVPR, Yang_2020_CVPR} resembles traditional video codecs~\cite{H.264, H.265}. More recently, feature-space video compression~\cite{Hu_2021_CVPR} has been proposed which includes additional modules for conversion to the feature-space (shown with dashed border in Figure~\ref{fig:videocompression}). Our adversarial perturbations are applicable to both classes of video compression models. \edit{Note that our perturbations are applied to the input of the video compression system and are generated while targeting all internal modules shown in Figure~\ref{fig:videocompression}, e.g., motion estimation.}


We further include an optional video classifier as the target downstream task after compression. This module predicts the activities performed in the video and is commonly deployed in video analysis services.  The downstream video classification service receives as input the decoded frames from the video compression pipeline. 
In what follows, we explain the internal mechanisms of the video compression modules shown in Figure~\ref{fig:videocompression}. Throughout the paper, we use ``video compression'' and ``video coding'' interchangeably. All DNN-based modules in Figure~\ref{fig:videocompression} are commonly used in SOTA video-compression.

\begin{figure*}[h]
\centering
    \begin{subfigure}[b]{0.32\textwidth}
    \centering
    \includegraphics[width=\linewidth]{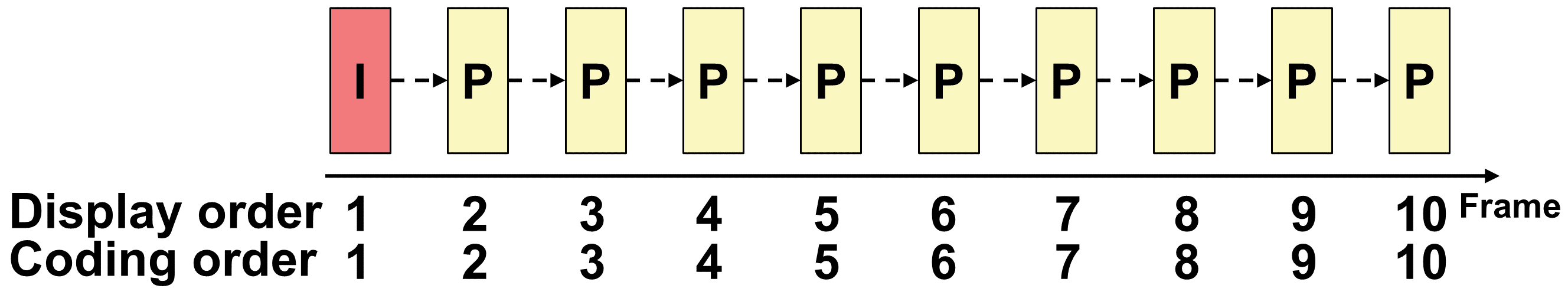}
    \caption{Non-hierarchical}
    \end{subfigure}
    \begin{subfigure}[b]{0.32\textwidth}
    \centering
    \includegraphics[width=\linewidth]{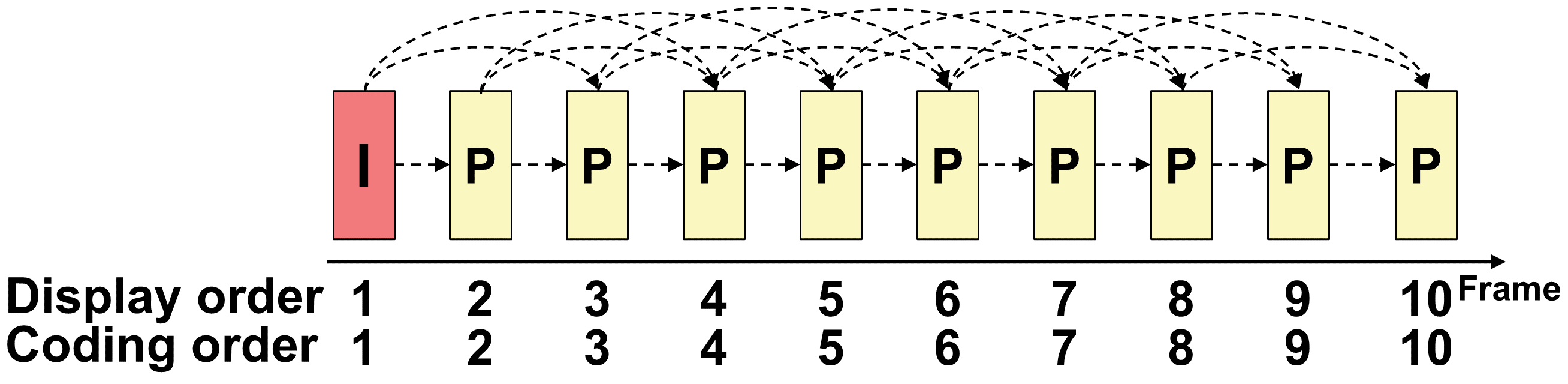}
    \caption{Hierarchical-P}
    \end{subfigure}
    \begin{subfigure}[b]{0.32\textwidth}
    \centering
    \includegraphics[width=\linewidth]{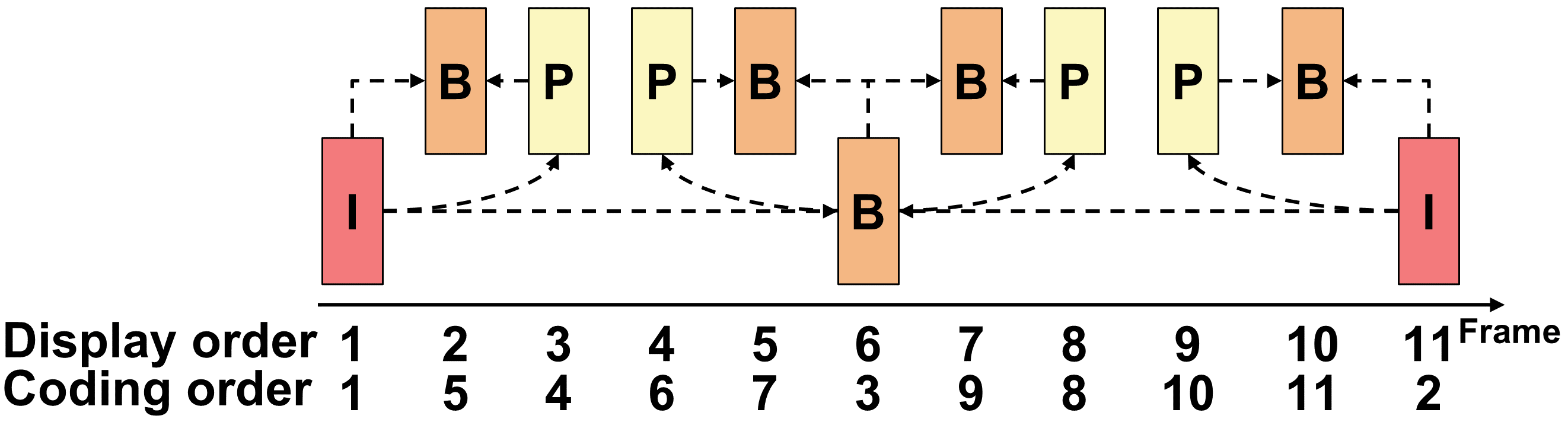}
    \caption{Hierarchical-B}
    \end{subfigure}
    \vspace{-0.3cm}
 \caption{(a) Non-hierarchical~\cite{Lu_2019_CVPR}, (b) hierarchical-P~\cite{Hu_2021_CVPR}, and (c) hierarchical-B~\cite{Yang_2020_CVPR} GOP structures for temporal coding.
}\label{fig:gop}
\vspace{-0.5cm}
\end{figure*}

\noindent\textbf{Temporal Coding Structure.} 
Since back-end viewers can join the live video stream at random points, the bitstreams are transmitted in GOP units for error resiliency. We systematically analyze our attacks on three GOP structures used in state-of-the-art video compression systems, namely, the non-hierarchical~\cite{Lu_2019_CVPR},  hierarchical-P~\cite{Hu_2021_CVPR}, and hierarchical-B~\cite{Yang_2020_CVPR} prediction structures. 
Figure~\ref{fig:gop} shows a high-level view of the GOP structures in our paper. 
A GOP contains different picture types, i.e., an I frame and several P and B frames. Within each GOP, the first frame, called I frame, is coded independently using image compression. A DNN-based image compression algorithm \cite{liu2020unified} is used to efficiently encode the I frame so that the P and B frames can refer to it with high quality. The P and B frames, which occupy most of the GOP, are constructed using video compression to remove temporal redundancy. Specifically, the P frame references the previous picture for the prediction, whereas the B frame uses both previous and forward frames to achieve the highest compression rate. 

\noindent\textbf{Video Encoder} uses a DNN-based motion estimation algorithm to measure the movement between the current frame and the previous reconstructed frame, dubbed the motion map. 
An auto-encoder network is employed to compress the motion vectors with a high compression rate. The encoder also includes a DNN-based motion compensation module which predicts the current frame using the reconstructed motion map and previous frames. The residual frame is then computed using the current frame and the prediction. The residual amount is compressed using a non-linear residual encoder network that efficiently quantizes the redundancy in the latent space. Finally, the quantized latent representations from the motion compression and residual compression are converted to bitstreams via entropy coding~\cite{Ball_2018_ICLR} and sent to the decoder.



\noindent\textbf{Video Decoder.} After receiving the bitstreams from the video encoder, the video decoder recovers the latent representations using entropy decoding. The decoder then applies motion and residual decoders to the latent representations and obtains the reconstructed motion map and residual frame. Similar to the encoder side, the decoder includes a DNN-based motion compensation module that predicts the current frame using information from the previous frames and the reconstructed motion map. The predicted frame and the reconstructed residual frame are added to create the decoded frame, which is then stored in a buffer. The decoder is thus able to reconstruct each frame with a same visual quality as the encoder. The quality of the decoded frame can be further improved through post-processing methods proposed in~\cite{Yang_2020_CVPR, Hu_2021_CVPR}.

\section{Threat Model}
\label{sec:threatmodel}
\subsection{Attack Scenarios} \label{sec:attack_scena}
RoVISQ attacks are targeted towards video data generated by front-end sources (e.g., smartphones and surveillance cameras) that is sent to back-end user(s) through a video compression pipeline. Since prior works on adversarial video~\cite{Li_2019_NDSS, Wei_aaai_2019, Pony_2021_CVPR, Xie_SP_2022} do not consider the role of video compression, their perturbations are mostly eliminated once the video is compressed. As such, the effect of previously proposed attacks on video classification will in fact not reach the target back-end user. To address this problem, our proposed attacks modify the video data at the first stage in the entire video compression and classification systems as shown in Figure~\ref{fig:intro}-(b).

RoVISQ attacks are applicable to autonomous vehicles~\cite{lu2020edge}, remote surgery~\cite{avgousti2018medical}, UHD streaming~\cite{wang2017cruise}, and virtual reality~\cite{ye2019omnidirectional}, where the adversary invokes low-quality or denial-of-service (DoS), e.g., as a competitor or malicious entity.
\edit{Ideally, the attacker's goal is to subvert the video streaming system over a long period of time. 
Therefore, our formulation ensures that RoVISQ perturbations applied to the input video are imperceptible to naked eye as visualized in Appendix Figure~\ref{fig:adv_input_video}. Therefore, our attacks to the input video remain stealthy, making it hard for the streaming service to identify the source of distortion for the users. }
\edit{Our threat model follows the state-of-the-art in image/video adversarial attack literature~\cite{Li_NIPS_2021, carlini2017adversarial,Li_2019_NDSS,Wei_aaai_2019,Xie_SP_2022,carlini2017towards,chen2020hopskipjumpattack}; we assume the attacker uses malware or man-in-the-middle schemes to carry out the attack. We consider both \textit{online} and \textit{offline} scenarios.} 

The online attack directly perturbs the front-end sources as the video is generated in real time. For this scenario, we design well-crafted universal perturbations that can be used to attack any given video sequence at any time step.
\edit{Since our universal perturbations are agnostic to the content of the video, the attacker does not require any knowledge about the streaming data. They can therefore carry out the attack by injecting RoVISQ's pre-trained perturbations using man-in-the-middle methods~\cite{Xie_SP_2022}, e.g., by running Ettercap with ARP poisoning~\cite{pingle2018real}. Specifically, as noted in~\cite{Xie_SP_2022}, a large number of video surveillance systems include unencrypted camera-server communications. Thus, the attacker can capture the stream from the camera network and modify it using our online perturbations before sending it to the media server and subsequently the back-end users. Alternatively, an attacker may alter the streaming video at the application level using malicious software installed during updates~\cite{Li_2019_NDSS,poeplau2014execute}. Signal injection attacks~\cite{kohler2022signal} can also be used to directly apply RoVISQ online perturbations to the camera sensor.} 

We note that video compression is not only used for live streaming, but also for various applications where the data is archived in the cloud. In this context, the target video can be raw data in the original quality or compressed data. If the videos are stored in memory, the attacker can use RoVISQ offline formulation to inject perturbations. Our offline attack scenario is analogous to the assumptions made in adversarial attacks on images~\cite{Szegedy_2013, Goodfellow_2014, Sharif_2020_CCS} and audio signals~\cite{Li_audio_ccs, 8424625}. \edit{In this scenario, the attacker adds small perturbations to a temporarily stored video using pre-installed malware on the streamer's device~\cite{poeplau2014execute}. The malware can be embedded via malicious applications downloaded through social engineering attacks or Internet browser-based vulnerabilities~\cite{kambar2022survey}.}
We consider the following two offline attack scenarios: 1) the adversary accesses the original raw data and perturbs the video, 2) the adversary decodes the compressed data, then inserts the compression-robust perturbation to the decoded video and re-encodes it. 
Through the above offline scenarios, the adversary can craft perturbed videos that degrade the QoE of back-end users.


\subsection{Adversary's Goal}
Throughout this paper, we use the terms ``attacker'' and ``adversary'' interchangeably.
The adversary's goal is to selectively degrade the bit-rate \textit{R} and/or distortion level \textit{D} compared to a pre-trained \textit{R}-\textit{D} curve for the video encoder and decoder. We also optionally target our attacks towards a downstream video recognition model, causing misclassification of the decoded videos. We introduce the following RoVISQ attacks:

\noindent\textbf{Video Quality Attack} increases the distortion \textit{D} at a given bit-rate, therefore, adding unwanted noise to the video content and reducing the visual quality for viewers. This attack is particularly advantageous when the media server administrator is monitoring the network bandwidth in real time. In this scenario, the service provider can detect anomalies in the bit-rate, but the proposed distortion attack remains stealthy.

\noindent\textbf{Bandwidth Attack} is formulated to increase \textit{R} at a given distortion level. As a result, the compression rate of the video encoder degrades and the amount of data transfer on the underlying network channel increases. This prevents legitimate users from successful communication with the streaming server and induces a high latency. 
\edit{Depending on whether or not the service provider uses adaptive bit-rate streaming~\cite{Kim_sigcomm_20,concolato2017adaptive}, the end-users either experience buffering when downloading high-resolution videos due to increased bit-rate or a reduced video resolution at a fixed bit-rate. Thus, by targeting \textit{R}, the adversary can cause video buffering or reduce the resolution of videos viewed by back-end users, which can lead to denial-of-service in real-time high-resolution streaming applications, e.g., remote surgery or vehicle control.}

\noindent\textbf{RD Attack} combines the capabilities of the above two attacks by simultaneously targeting \textit{R} and \textit{D} to cause a high latency and video distortion. As a result, the back-end users suffer from low-quality or denial-of-service. If the media server lowers the video resolution to reduce network traffic, the \rdo is further exacerbated. This is due to the observation made by~\cite{ou2014q}, which show that distortion is more perceptually noticeable at a reduced video resolution.

\noindent\textbf{Compression-Robust Classifier Attack.} Our final attack manipulates the classification result of the decoded video in scenarios where the downstream task uses a DNN-based video recognition. This attack is particularly challenging as the video coding framework inherently invalidates most adversarial examples using DNN-based temporal coding. As such, we carefully craft an optimization problem that generates perturbations that are robust to video coding. We propose two variants of the classifier attack, i.e., targeted and untargeted. Targeted attacks misguide the video classifier to a particular class while untargeted attacks subvert the models to predict any of the incorrect classes.

\subsection{Adversary's Capability and Knowledge}
Video coding standards are jointly determined by various standardization organizations, e.g., the ISO/IEC MPEG, ITU-T Video Coding Experts Group (VCEG), and industry partners (Google, Cisco, Netflix, Qualcomm, etc.) \cite{H.264, H.265, AV1, Mpeg, Qualcomm}. To continue the development of the latest video coding standards, the reference software algorithms are made publicly available and are accompanied by several detailed specification documents~\cite{samelakjoint} to ensure ease of use by clients. 
These open-source software for video compression are directly integrated into research and commercial products. 
Thus, \edit{we assume the attacker has white-box access to an open-source video compression model, i.e., the adversary knows the DNN architecture and weights. We also assume white-box access to an open-source video classifier as explored in prior works \cite{Li_2019_NDSS, Wei_aaai_2019, Pony_2021_CVPR}. These white-box systems are leveraged by the attacker as a surrogate model to create the universal perturbations. The universal perturbations can then be applied to unseen models in a black-box attack scenario, i.e., when the adversary does not know the internals of the underlying DNN models in a video compression/classification system.}

To construct our online attack via universal perturbations, we assume the attacker also has access to a set of public (benign) videos. \edit{As shown in our experiments, even though the training and test datasets for the universal perturbations have different distributions, our attacks are still applicable to the test data. Therefore, the attacker can use any open-source video dataset which is easily accessible.}

\section{Attack Construction}
\label{sec:white-box_construction}
This section explains RoVISQ attacks on video compression and classification for both online and offline scenarios. We first formulate our attacks as well-defined optimization problems following the threat model. We then adapt our formulations to enable offline and online attacks.

\begin{figure}[t]
\centering
\includegraphics[width=\columnwidth]{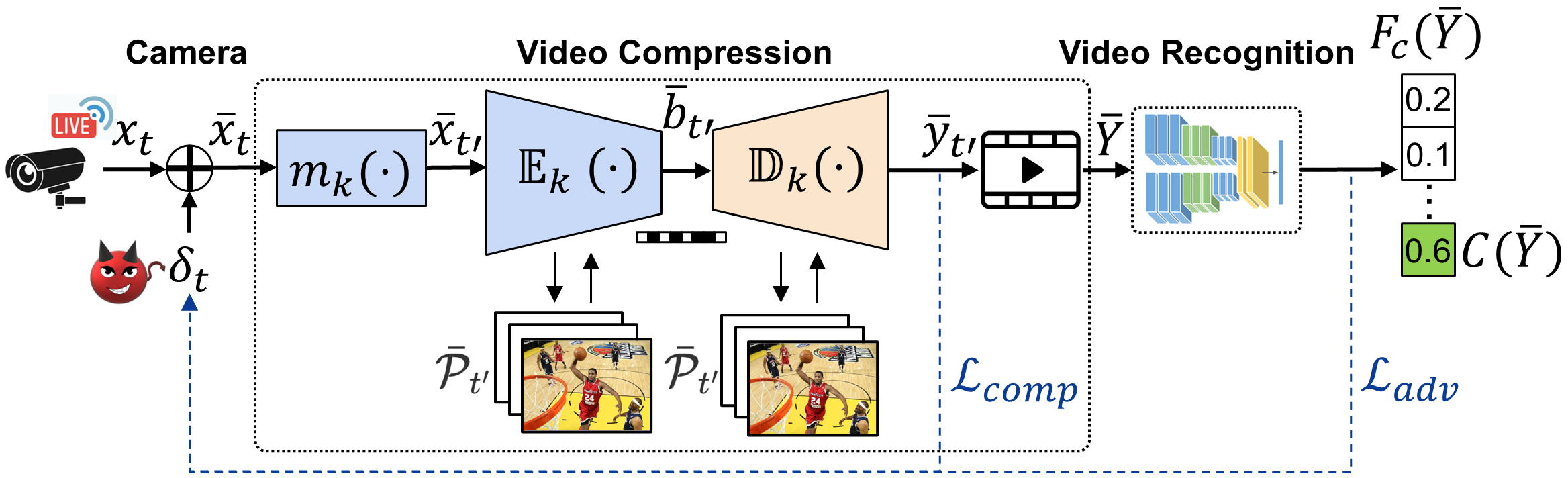}
\caption{Overview of system components for the proposed RoVISQ attacks on video compression and classification. 
}
\label{fig:attack_overview}
\vspace{-0.7cm}
\end{figure}

\subsection{Problem Formulation}
\smallskip
\noindent\underline{\textbf{Conventional Video Compression.}}
Let $X=\{x_{1},\dots, x_{T}\} \in \mathbb{R}^{T\times W \times H \times C}$ denote an original video clip containing $T$ consecutive frames. Here, $x_t$ represents the frame at a time step $t$ which contains $H$ rows, $W$ columns and $C$ color channels.
As shown in Figure~\ref{fig:gop}, DNN-based video compression frameworks~\cite{Yang_2020_CVPR,Lu_2019_CVPR,Hu_2021_CVPR} use different coding orders according to the underlying GOP structure. Each frame in the coding order is mapped to one time step in the input video based on the type of GOP structure $k\in\{\text{non-hierarchical}, \text{hierarchical-P}, \text{hierarchical-B}\}$. We note that this time step, denoted by $t'$, may not be equivalent to the display order $t$, i.e., the order of frames in which the video is captured.
For a given $k$, the $n$-th coding order in the $g$-th GOP is mapped to time step $t'$ using a deterministic function $m_{k}(g,  n)=t'$. Here, $1 \leq n \leq G$ and $0 \leq g \leq \floor{\frac{T}{G}}$ where $G$ denotes the number of frames in the GOP.

Accordingly, we formulate the video encoder as a function $\mathbb{E}_{k}(x_{t'}, \mathcal{P}_{t'}, \lambda) = b_{t'}$ that accepts an input $x_{t'}$ and produces an output bitstream $b_{t'}$. The conversion is performed using several reference frames stored in the decoded frames' buffer: 
$$\mathcal{P}_{t'}= \{y_{m_{k}(g, 1)}, \dots, y_{m_{k}(g, n-1)}\},$$
where $\mathcal{P}_{t'}=\emptyset$ when $n=1$. Here, $\lambda$ controls the trade-off between the bit-rate ($\textit{R}$) and the distortion ($\textit{D}$). The video decoder is formulated as a function $\mathbb{D}_{k}(b_{t'}, \mathcal{P}_{t'}, \lambda)=y_{t'}$ that takes as input the encoded bitstream $b_{t'}$ and reconstructs the decoded frame $y_{t'}$. 
The end-to-end video compression model~\cite{Yang_2020_CVPR, Lu_2019_CVPR, Hu_2021_CVPR} is trained to solve the following \textit{R}-\textit{D} optimization over the $t'$-th frame $x_{t'}$:
\begin{equation} \label{eq:video_compression}
\begin{aligned}
\min_{w} \text{ } R(b_{t'}) + \lambda \cdot D(x_{t'}, y_{t'}), 
\end{aligned}
\end{equation}
where $w$ represents the weights parameters of DNN, $R(\cdot)$ indicates the number of bits allocated for the $t'$-th frame, and $D(\cdot)$ measures the distortion between the input frame $x_{t'}$ and its reconstruction $y_{t'}$. The distortion $D(\cdot)$ is often measured using the mean squared error (MSE)~\cite{Yang_2020_CVPR, Lu_2019_CVPR, Hu_2021_CVPR}.

\noindent\underline{\textbf{Adversarial Attacks on Video Compression.}}
\label{subsec:AdversarialAttacksonVideoCompression}
Let $\Delta =$ $\{\delta_{1}, \dots,\delta_{T}\}$ $\in \mathbb{R}^{T\times W \times H \times C}$ denote a perturbation set for a given video $X$. 
We denote the resulting adversarial video by $\bar{X}=\{\bar{x}_{1},\dots, \bar{x}_{T}\} \in \mathbb{R}^{T\times W \times H \times C}$ which consists of adversarial frames $\bar{x}_{t}=x_{t}+\delta_{t}$. 

Upon receiving the adversarial input $\bar{x}_{t'}$, the encoder outputs an adversarial bitstream $\bar{b}_{t'}$ using perturbed reference frames stored in the adversarial decoded frames' buffer $\bar{\mathcal{P}}_{t'}=\{  \bar{y}_{m_{k}(g, 1)}, \dots, \bar{y}_{m_{k}(g, n-1)}\}$. Finally, the decoder reconstructs the adversarial decoded frame $\bar{y}_{t'}$ from the perturbed bitstream $\bar{b}_{t'}$. Figure~\ref{fig:attack_overview} demonstrates RoVISQ attacks to the video compression modules, along with the corresponding inputs, outputs, and the adversarial perturbation.



Recall the coding of each frame $\bar{x}_{t'}$ is conditioned upon the coding of its preceding frame stored in the (adversarial) decoded frames' buffer $\bar{\mathcal{P}}_{t'}$. This temporal coding constructs a chain of dependency between all adjacent frames in each GOP. To account for the inter-frame dependency (IFD) within a video sequence, our proposed adversary focuses on the coding performance of one GOP unit. Specifically, we construct our attacks based on the following adversarial \textit{R}-\textit{D} model considering the IFD in for the $g$-th GOP:
\begin{equation} \label{eq:QoE_Fators}
\begin{aligned}
\max_{\Delta_g} \quad & \frac{1}{G} \sum^{m_{k}(g, G)}_{t' = m_{k}(g, 1)} (R(\bar{b}_{t'}) + \lambda \cdot D(x_{t'}, \bar{y}_{t'})), \\
\end{aligned}
\end{equation}
where $\Delta_g\in\mathbb{R}^{G\times W\times H\times C}$ is the perturbation for the $g$-th GOP.

We quantify the video compression performance based on two important QoE factors for back-end subscribers: $\{Q_{0},Q_{1}\}$. Here, $Q_{0}$ denotes the end-users' expectations for network bandwidth and $Q_{1}$ represents the expectation for video distortion. Using the perturbation in Equation~\ref{eq:QoE_Fators}, we estimate the QoE factors at intervals of GOP size $G$ as follows\footnote{\edit{Our attacks are easily applicable to compression methods without a GOP structure. In this case, the attacker simply sets GOP size $G=1$.}}:
\begin{align} \label{eq:Entire_QoE_Fators}
    Q_{0}(\mathcal{\bar{B}}_{g}) = & \frac{1}{G} \sum_{\bar{b}_{t'} \in \bar{B}_{g}} R(\bar{b}_{t'}) \\
    \label{eq:Entire_QoE_Fators2}
    Q_{1}(X_{g}, \bar{Y}_{g}) = & \frac{1}{G} \sum_{\bar{y}_{t'} \in \bar{Y}_{g}} D(x_{t'}, \bar{y}_{t'}),
\end{align}
where $X_{g}$, $\bar{\mathcal{B}}_{g}$, and $\bar{Y}_{g}$ denote the set of input frames, adversarial bitstreams, and (adversarial) reconstructed frames over the time steps in the $g$-th GOP, i.e., $t' \in [m_{k}(g,1),m_{k}(g,G)]$. 
We formulate RoVISQ attacks to manipulate the above QoE factors. Note that our definition of $Q_{0},Q_{1}$ captures the IFD through an implicit conditioning on the GOP temporal coding function $m_k(\cdot)$. As such, our perturbations can adapt to any given GOP structure without need for reformulating the attack. Below we explain how each of our attacks are constructed. 

\noindent\textbf{1- Video Quality Attack.}
The adversarial perturbation $\Delta_g$ is determined by solving the following optimization problem:

\begin{equation} \label{eq:video_quality_optimization_problem}
\max_{\Delta_g} \ Q_{1}(X_{g}, \bar{Y}_{g}) \quad\quad
\mathrm{s.t.} \quad \mathbf{E}_{0} < \epsilon_{0}.
\end{equation}
Here, $\mathbf{E}_{0}=\norm{Q_{0}(\mathcal{\bar{B}}_{g}) - Q_{0}(\mathcal{B}_{g})}_{2}$ is the difference between the expected bit-rate of the clean and perturbed videos.

\noindent\textbf{2- Bandwidth Attack.}
The objective of the bandwidth attack is to find $\Delta_g$ that maximizes the bit-rate while keeping the video distortion unaltered as follows:

\begin{equation} \label{eq:bandwidth_optimization_problem}
\max_{\Delta_g} \ Q_{0}(\mathcal{\bar{B}}_{g}) \quad\quad
\mathrm{s.t.} \quad  \mathbf{E}_{1} < \epsilon_{1},
\end{equation}
where $\mathbf{E}_{1}=\norm{Q_{1}(X_{g}, \bar{Y}_{g}) - Q_{1}(X_{g}, {Y}_{g})}_{2}$ is the difference between the expected distortion of clean and perturbed videos. $Y_{g}$ is the set of clean reconstructed frames in the $g$-th GOP.

\noindent\textbf{3- RD Attack.}
$\Delta_g$ is found by solving the following optimization problem that affects both QoE factors at the same time:

\begin{equation} \label{eq:rdo_optimization_problem}
\max_{\Delta_g} \ Q_{0}(\mathcal{\bar{B}}_{g}) + \lambda \cdot  Q_{1}(X_{g}, \bar{Y}_{g}).
\end{equation}


\noindent\underline{\textbf{Compression-Robust Classifier Attacks.}}
Figure~\ref{fig:attack_overview} illustrates a video classification module appended to the video compression pipeline. Video classification relies on a discriminant function $F(\bar{Y})$ that takes as input a perturbed decoded video clip $\bar{Y}=\{\bar{y}_{1},\dots, \bar{y}_{T}\} \in \mathbb{R}^{T\times W \times H \times C}$ and outputs a probability distribution over a set $K$ of class labels. $F_c(\bar{Y})$ indicates the probability of the input video belonging to a specific class $c \in K$. The video classifier $\mathcal{C}$ maps an input $\bar{Y}$ to the class with the maximum probability: $\mathcal{C}(\bar{Y}) = \text{argmax}_{c \in K} F_{c}(\bar{Y})$.
The untargeted attack against $\mathcal{C}$ generates perturbations such that $\mathcal{C}(\bar{Y}) \neq \mathcal{C}(Y)$, whereas the targeted attack aims at $\mathcal{C}(\bar{Y})= c^* (\neq \mathcal{C}(Y))$ for a predetermined incorrect class $c^{*} \in K$, where $Y$ is the clean decoded video clip.
The adversarial loss $\mathcal{L}_{adv}$ for the untargeted and targeted attacks can be written as:
\begin{equation} \label{eq:classification}
\mathcal{L}_{adv} =\begin{cases}
  F_{\mathcal{C}(Y)}(\bar{Y}) - \max\limits_{c \neq \mathcal{C}(Y)}F_{c}(\bar{Y}) & \text{(Untargeted)}\\
  \max\limits_{c \neq c^{*}}F_{c}(\bar{Y}) - F_{c^{*}}(\bar{Y}) & \text{(Targeted)}
\end{cases}
\end{equation}
where $\bar{Y}$ is the perturbed decoded video. Both attacks gain success if and only if $\mathcal{L}_{adv}<0$. To create
versatile perturbations that are robust to compression, we formulate RoVISQ attacks against video classification with consideration for both video compression
and classification. Specifically, the adversarial perturbation $\Delta_g$ is computed using the same formula in Equations~\ref{eq:video_quality_optimization_problem}, \ref{eq:bandwidth_optimization_problem}, \ref{eq:rdo_optimization_problem} with a new constraint added for $\mathcal{L}_{adv}<0$.

\subsection{Offline Attack Methodology}
In the offline attack scenario, injecting the adversarial perturbations is not latency bound. The attacker can therefore use the entire captured video, before creating and adding the adversarial perturbations. To generate the perturbations, the adversary leverages the proposed approximations of QoE to maximize the following adversarial loss function:
\begin{equation} \label{eq:offline1}
\begin{aligned}
\max_{\Delta_g} \quad & \mathcal{L}_{comp}(g) \quad \mathrm{s.t.} \quad \norm{\Delta_{g}}_{\infty} \leq \epsilon_{c}\\
\mathcal{L}_{comp}(g) = &
\begin{cases}
\mathbf{E}_{0} + \lambda \cdot Q_{1}(X_{g}, \bar{Y}_{g}) & \mathrm{if} \ \xi=0 \\
Q_{0}(\mathcal{\bar{B}}_{g}) + \lambda \cdot \mathbf{E}_{1} & \mathrm{if} \ \xi=1 \\
Q_{0}(\mathcal{\bar{B}}_{g}) + \lambda \cdot Q_{1}(X_{g}, \bar{Y}_{g}) & \mathrm{if} \ \xi=2, \\
\end{cases}
\end{aligned}
\end{equation}
where $\xi$ determines the attack type: $\xi=0$ presents the video quality attack, $\xi=1$ is the bandwidth attack, and $\xi=2$ is the \rdo. To ensure the injected noise is imperceptible to humans, the norm of the perturbation is upper bounded by a pre-defined small value $\epsilon_c$ for all attacks. We use the iterative FGSM (I-FGSM)~\cite{kurakin2016adversarial} method to solve Equation~\ref{eq:offline1} as follows:
\begin{equation}\label{eq:fgsm_comp}
\begin{aligned}
    \hat{X}_{n+1} = \mathrm{clip}_{0,1}(\Tilde{X}_{n} + \frac{\epsilon_{c}}{\rho}\mathrm{sign}(\nabla\mathcal{L}_{comp}(g)) \\
    \Tilde{X}_{n+1} = \mathrm{clip}_{-\epsilon_{c},\epsilon_{c}}(\hat{X}_{n+1}-\Tilde{X}_{0})+\Tilde{X}_{0},
\end{aligned}
\end{equation}
where $\rho$ is the total number of iterations. Here, $\Tilde{X}_{0}$ is equal to the original (benign) video $X_{g}$ and the final perturbed video $\bar{X}_{g}$ will be equal to $\Tilde{X}_{\rho}$ from the last iteration of Equation~\ref{eq:fgsm_comp}. 

Equation~\ref{eq:offline1} delivers the perturbations for downgrading the performance (QoE) of the video compression system itself. The adversary can optionally choose to repurpose the RoVISQ attacks to also affect a downstream video recognition system. In such scenarios, we integrate $\mathcal{L}_{adv}$ to simultaneously derive perturbations on video compression and classification. Therefore, we convert Equation~\ref{eq:offline1} to the following objective function:
\begin{equation} \label{eq:offline2}
\max_{\Delta}  \ \mathcal{L}_{total} = \frac{1}{\floor{T/G}+1}\sum^{\floor{T/G}}_{g=0} \mathcal{L}_{comp}(g) - \beta \cdot \mathcal{L}_{adv},
\end{equation}
where $\beta$ adjusts the scale of the two loss functions. We use a similar method as Equation~\ref{eq:fgsm_comp} to solve the above maximization problem. Specifically, we replace the loss function with $\mathcal{L}_{total}$ to generate perturbations for both compression and classification models.

\subsection{Online Attack Methodology}
\label{subsec:uni}

Online adversarial attack is particularly challenging for several reasons. Firstly, the perturbed decoded frames from video encoder and decoder are conditioned on $\lambda$ which controls the compression rate. Since $\lambda$ is determined by the end-user's network traffic, it is difficult for the adversary to infer the value of $\lambda$ in real time. Secondly, the display and coding orders vary according to the GOP structure set for the victim video compression. Even if the adversary knows the GOP structure used by the system, they can not predict the order of the frames in the live video stream. As such, computing the QoE factors in Equations~\ref{eq:Entire_QoE_Fators}, \ref{eq:Entire_QoE_Fators2} are not possible in the online attack. Finally, the adversary may not be able to align the image perturbations with the video sequence, which reduces the attack performance due to a boundary effect \cite{Li_2019_NDSS, Xie_SP_2022}. If not addressed properly, the said challenges hinder that attacker's success and degrade the functionality of the adversarial perturbation.

We thus propose a novel attack that is universally applicable to live video streams using pre-trained perturbations that are generated offline. Our universal perturbations are agnostic to compression ratio, GOP structure, and input, which is suitable for online attack. 
Algorithm~\ref{alg:pseudocode} outlines the training phase of our online attack. In crafting the universal perturbation, we adapt the perturbation to all variations of compression ratio and GOP structures ($\mathbb{G} = \{\text{non-hierarchical}, \text{hierarchical-P}, \text{hierarchical-B}\}$). These variations are captured in a universal loss
$\mathcal{L}_{univ}$, which is used to derive the final perturbation.
Additionally, we address the boundary effect using a temporal transformation function $\Gamma(\cdot)$ for the 3D perturbation with shifting variable $\tau$ \cite{Li_2019_NDSS, Xie_SP_2022}. 


Let us denote the universal adversarial perturbation for a GOP by $\Phi =\{\phi_{1}, \phi_{2}, \dots, \phi_{G}\} \in \mathbb{R}^{G \times W \times H \times C}$. We define a permutation function $\Gamma(\Phi, \tau)$ which produces a cyclic temporal shift of the original perturbation $\Phi$ by an offset $\tau \in [0, G-1]$:
\begin{equation*}
\resizebox{\columnwidth}{!}{
$\begin{aligned}
    \Gamma(\Phi, \tau) =& \{ \phi_{1,\tau}, \phi_{2,\tau} , \dots, \phi_{G,\tau}\}\\
    =& \{\phi_{( \tau \text{ mod } G ) + 1}, \phi_{( (1+\tau) \text{ mod } G ) + 1}, \dots, \phi_{( (G-1+\tau) \text{ mod } G ) + 1}\}.
\end{aligned}$}
\end{equation*}

\setlength{\textfloatsep}{2pt}
\begin{algorithm}[t]
\small
\caption{Online Attack}\label{alg:pseudocode}
\begin{algorithmic}
\State \textbf{Input:} Training data $\mathbb{T}$, Perturbation bound $\epsilon_c$, Attack $\xi$
\State \textbf{Output:} Universal Perturbation $\Phi= \{\phi_{1}, \dots, \phi_{G}\}$
\State $\Phi \gets \emptyset$
\For {$i < \mathrm{MaxIter}$}
    \State ${L}_{univ} \gets 0$
    \For {$X \in \mathbb{T}$}
        \For {$ k \in \mathbb{G}$}
            \For {$\lambda \in \{256,512,1024,2048\}$}
            \State $\mathcal{\bar{B}}_{g} \gets \emptyset, \bar{Y}_{g} \gets \emptyset$
                \For {$g \in \{0, 1, \dots, \floor{\frac{T}{G}}\}$}
                    \State sample $\tau \sim \{0,\dots,G-1\}$
                    \State $\Phi \gets \Gamma(\Phi, \tau)$
                    \State $\bar{X}_{g} \gets X_{g} + \Phi$ 
                    \State $\bar{\mathcal{P}}_{t'} \gets \emptyset$
                    \For {$n \in \{1, \dots, G\}$}
                        \State $t' \gets m_{k}(g, n)$
                        \State $\bar{b}_{t'} \gets \mathbb{E}_{k}(\bar{x}_{t'}, \bar{\mathcal{P}}_{t'}, \lambda)$
                        \State $\bar{y}_{t'} \gets \mathbb{D}_{k}(\bar{b}_{t'}, \bar{\mathcal{P}}_{t'}, \lambda)$
                        \State $\mathcal{\bar{B}}_{g}$.append($\bar{b}_{t'}$), $\bar{Y}_{g}$.append($\bar{y}_{t'}$)
                        \State $\bar{\mathcal{P}}_{t'}$.append($\bar{y}_{t'}$)
                    \EndFor
                    \State $\mathcal{L}_{comp}(g) \gets$ Equation~\ref{eq:offline1} based on $\xi$
                \EndFor
                \State $\mathcal{L}_{total} \gets$ Equation~\ref{eq:offline2}
                \State $\mathcal{L}_{univ} \gets {L}_{univ} + \mathcal{L}_{total}$            
            \EndFor
        \EndFor
    \EndFor
    \State Calculate the mean of $\mathcal{L}_{univ}$
    \State Update $\Phi$ using iterative FGSM~\cite{kurakin2016adversarial} s.t. $\norm{\Phi}_{\infty} \leq \epsilon_{c}$.
\EndFor
\State \textbf{Return:} $\Phi$
\end{algorithmic}
\end{algorithm}

Once the temporal shift is applied to the perturbation, the adversarial video of the $g$-th GOP can be computed as $\bar{X}_{g}=\{x_{1}+\phi_{1,\tau},\dots, x_{G}+\phi_{G,\tau}\}$. Note that $\bar{X}_{g}$ receives all $G$ possible temporal shift transformations during training with a high probability, therefore ensuring the generalizability of $\Phi$. Instead of applying the adversarial perturbation to one video clip, we obtain the universal perturbation by averaging the value across all training videos available to the attacker. When computing the adversarial loss, the adversary can set $\beta$ to zero in Equation~\ref{eq:offline2} to limit the scope of the attack to the compression model only, or a non-zero value to attack a downstream video classification system. In the latter scenario, we consider the case where $\mathcal{L}_{adv}$ is untargeted. We iteratively update $\Phi$ with the I-FGSM method as shown in Equation~\ref{eq:fgsm_comp}.

We note that prior works on video recognition attacks~\cite{Li_2019_NDSS, Xie_SP_2022} do not consider video compression in their threat model for the universal attack. Therefore, they have the disadvantage compared to our formulation that the adversarial perturbation for video recognition is not robust to video compression. Their attack performance against video classifiers drops dramatically as the compression rate increases, as we show in our evaluations (see Figure~\ref{fig:Exp_asr_vcs}). To break down video compression, our perturbations are learned in consideration of the inter-frame dependencies and error propagation.

\section{Attack Evaluation}
\label{sec:experiments_wb}
We first evaluate RoVISQ attacks on video compression in Section~\ref{sec:qoe_attacks}. \edit{We further provide a user study in Section~\ref{sec:user_study} to investigate the effect of RoVISQ attacks on user QoE.} Finally, in Section~\ref{sec:exps_classification}, we evaluate our compression-robust attack against a downstream video classifier by considering the video compression and classification modules in one framework.

\subsection{Experimental Setup} 
\noindent\textbf{RoVISQ Attacks on Video Compression.}
We use the Vimeo-90K dataset~\cite{xue2019video} to train the video compression models. This dataset contains 89,800 video clips with 7 frames each, with a resolution of $448 \times 256$. We crop the video sequences into a resolution of $256 \times 256$ before training. As our victim video compression module, we consider three state-of-the-art models with different structures, namely, DVC~\cite{Lu_2019_CVPR}, HLVC~\cite{Yang_2020_CVPR}, and FVC~\cite{Hu_2021_CVPR}. We implement each model through an open source framework~\cite{pytorchvc} for video compression. We train the benign models with different $\lambda$ values ($\lambda=$256, 512, 1024, 2048) using the loss function in Equation~\ref{eq:video_compression}.

To generate the adversarial perturbations for the offline and online scenarios, we use the loss function in Equation~\ref{eq:offline2} and the pseudo code in Algorithm~\ref{alg:pseudocode} with $\beta=0$. We set I-FGSM iterations to 50 and $\epsilon_c$ to $\frac{10}{255}$, following previous studies \cite{Wei_aaai_2019, Pony_2021_CVPR, Xie_SP_2022, Li_2019_NDSS}. \edit{For the offline attack, we assume the adversary adds perturbations to the raw video before compression is performed (offline attack scenario 1 in Section~\ref{sec:attack_scena}).} For the online attack, we use the Vimeo-90K dataset to train our universal perturbations. To evaluate the attacks, the video sequences in the HEVC datasets~\cite{H.265}, i.e., classes B, C, and D are used. HEVC class B is a high resolution ($1920 \times 1080$) dataset while classes C and D have lower resolutions of $834 \times 480$ and $416 \times 240$, respectively. Following prior work~\cite{Lu_2019_CVPR, Yang_2020_CVPR, Hu_2021_CVPR}, we test the HEVC datasets on the first 100 frames and set $G$ to 10 for non-hierarchical and hierarchical-P GOPs. For hierarchical-B GOP, we set $G$ to 11, and the last frame of each GOP is reused as the first frame of the next GOP. Video quality is measured as peak signal-to-noise ratio (PSNR) and the bit-rate is calculated by bits per pixel (Bpp). 

\noindent\textbf{Baselines on Video Compression.}
We compare RoVISQ attacks on video compression models with adaptive Gaussian noise: $\epsilon_{g}\sim\mathcal{N}(0,\sigma^{2})$. Specifically, since the I frame strongly affects P and B frames in the GOP~\cite{huszak2010analysing}, we evaluate two scenarios, where the I frame either receives higher noise or equal, compared to the P and B frames.
In summary, we use two types of Gaussian noises: (1) Case I ( $\sigma_{I}=\sigma_{P}=\sigma_{B}=\epsilon_c$), (2) Case II ( $\sigma_{I}=2 \cdot \epsilon_c, \sigma_{P}=\sigma_{B}=\epsilon_c$).
We also include the traditional video coding methods, H.264 \cite{H.264}, H.265 \cite{H.265}, as our baseline \edit{and show that RoVISQ attacks also degrade the performance of legacy methods.} We follow the command line in \cite{Lu_2019_CVPR} and use FFmpeg with default mode to implement H.264 and H.265. 


\noindent\textbf{RoVISQ Attacks on Video Classification.} 
We benchmark three state-of-the-art video classification models including I3D \cite{Carreira_2017_CVPR}, SlowFast \cite{Feichtenhofer_2019_ICCV}, and TPN \cite{TPN_2020_CVPR}. We leverage the DVC  framework as the backbone for video compression. \edit{We assume the attacker decodes a video that has already been compressed with DVC, then injects perturbation and re-encodes it (offline attack scenario 2 in Section~\ref{sec:attack_scena}).}
Our attacks are evaluated on the human action recognition dataset UCF-101 \cite{soomro2012ucf101} and the hand gesture recognition dataset 20BN-JESTER (Jester) \cite{materzynska2019jester}. UCF-101 includes 13320 videos from 101 human action categories (e.g., diving, biking,
blow-drying hair, yo yo). Jester includes 27 gestures recorded by crowd-sourced workers (e.g., sliding hand left, sliding two fingers left, zooming in with full hand, zooming out with full hand). 

To generate the offline and online adversarial perturbations, we set $\beta=0.1$ and $\epsilon_c=\frac{10}{255}$~\cite{Wei_aaai_2019, Pony_2021_CVPR, Xie_SP_2022, Li_2019_NDSS} in Equation~\ref{eq:offline2} and Algorithm~\ref{alg:pseudocode}. We set $\mathrm{MaxIter}$ to 50 steps for I-FGSM. \edit{Following the evaluation setup in prior work~\cite{Li_NIPS_2021}, we randomly pick the target class for our targeted attacks. For assessing both targeted and untargeted attacks, we randomly chose one and four videos from each human action category for UCF-101 and Jester datasets, respectively. 
}

\noindent\textbf{Baselines on Video Classification.}
We compare the performance of RoVISQ attacks with the following state-of-the-art adversarial attacks on video recognition: SAP~\cite{Wei_aaai_2019}, GEO-TRAP~\cite{Li_NIPS_2021}, C-DUP~\cite{Li_2019_NDSS}, and U3D~\cite{Xie_SP_2022}. Similar to the threat model for our attacks, we insert the perturbations created by prior work to the inputs of the video compression and classification. The perturbed videos are then compressed and reconstructed by video compression. Finally, they are fed to the victim classifiers. We define the attack success rate as the portion of misclassified test samples for the untargeted attack. For the targeted scenario, the attack success rate is the portion of samples mapped to the adversary's desired class.

\begin{figure*}[ht]
\centering
  \begin{tabular}{@{}c@{}}
    \includegraphics[width=\linewidth]{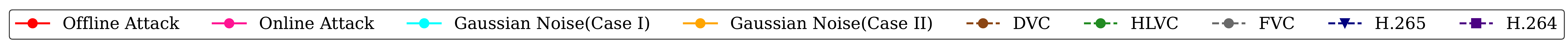} \\  
  \end{tabular}
    \begin{tabular}{@{}ccc@{}}
    \includegraphics[width=0.32\linewidth]{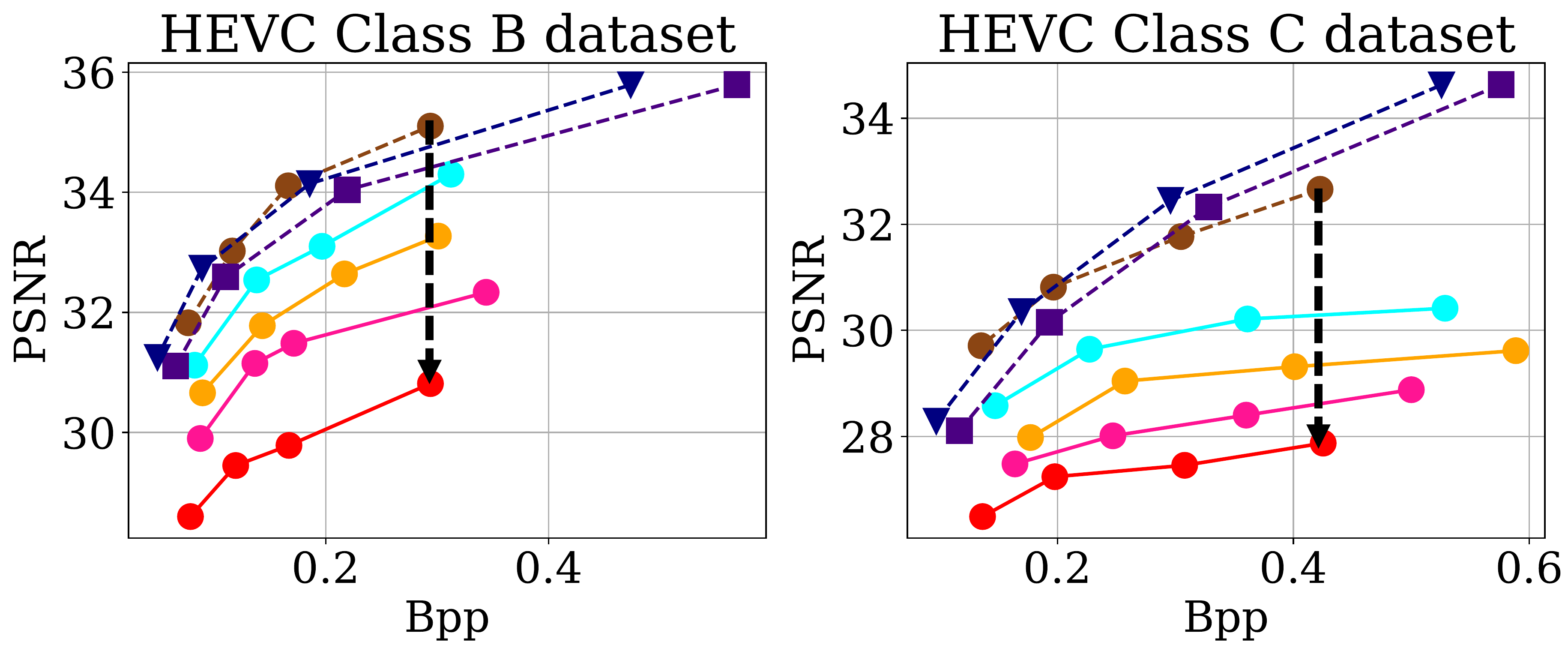}   &
    \includegraphics[width=0.32\linewidth]{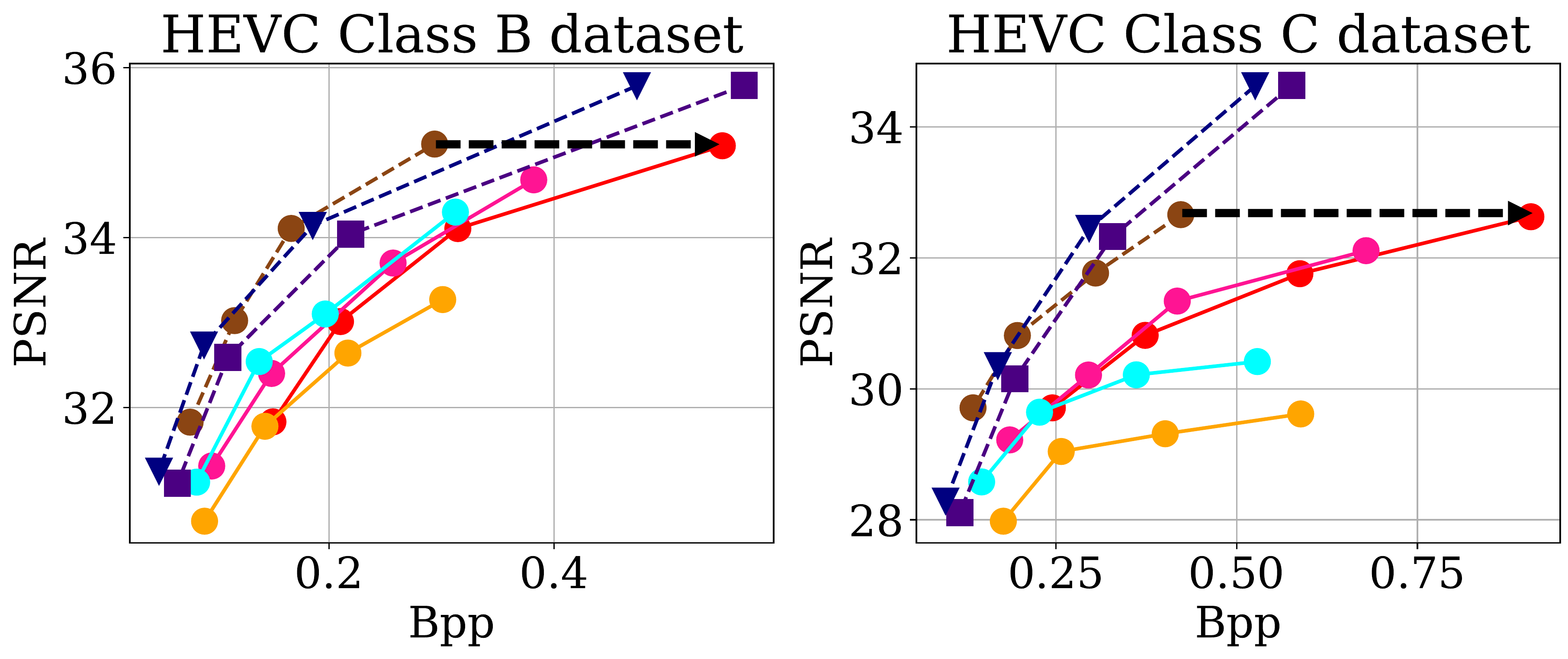}    &
    \includegraphics[width=0.32\linewidth]{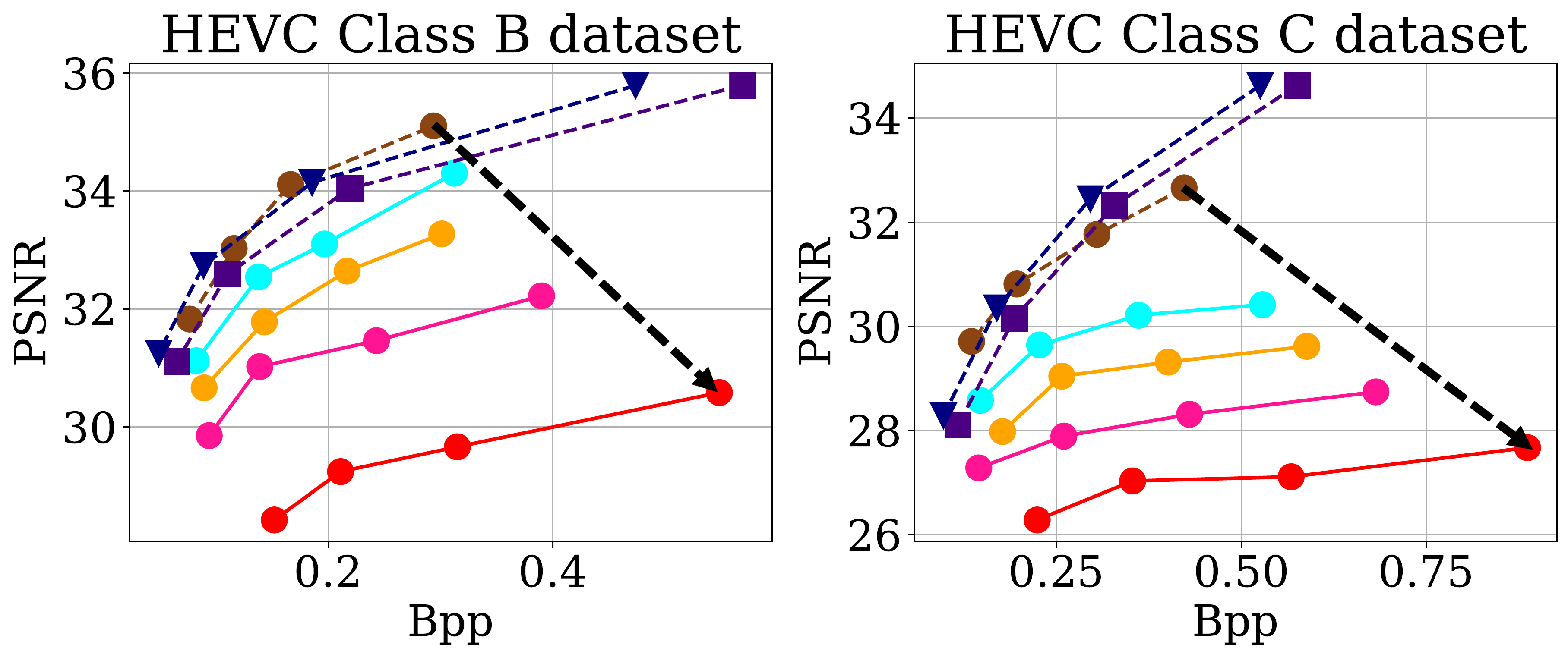} \\
    \footnotesize{Video Quality Attack}&
    \footnotesize{Bandwidth Attack}&
    \footnotesize{RD Attack} \\
    \end{tabular}
    \begin{tabular}{@{}c@{}}
    (a) DVC~\cite{Lu_2019_CVPR}\\
    \end{tabular}
    \begin{tabular}{@{}ccc@{}}
    \includegraphics[width=0.32\linewidth]{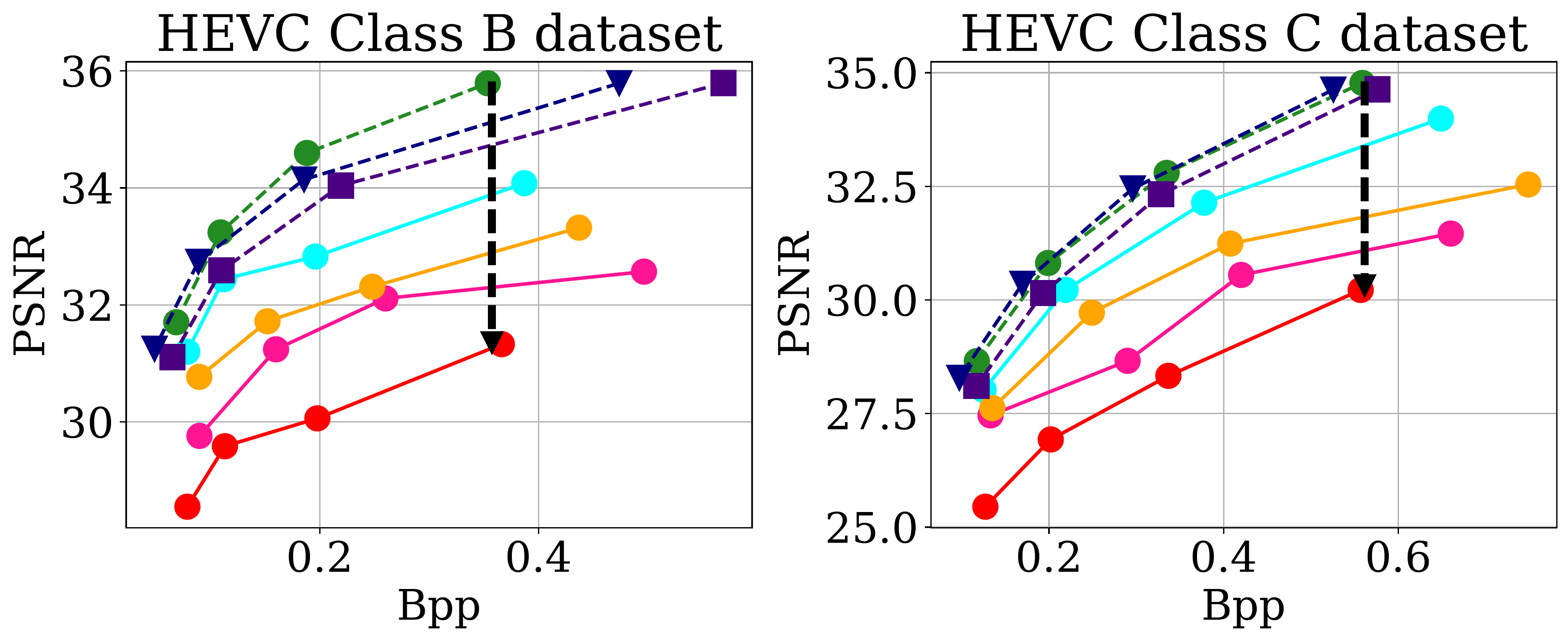}   &
    \includegraphics[width=0.32\linewidth]{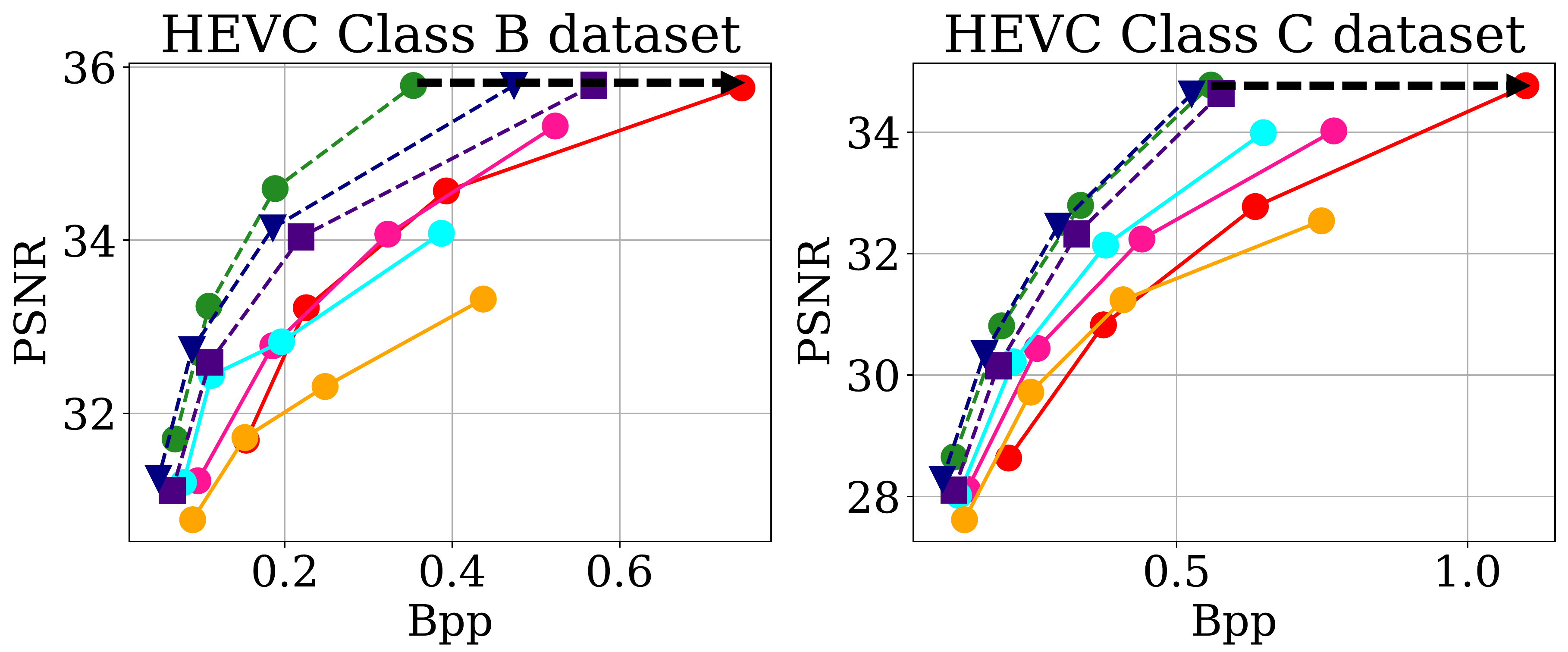}    &
    \includegraphics[width=0.32\linewidth]{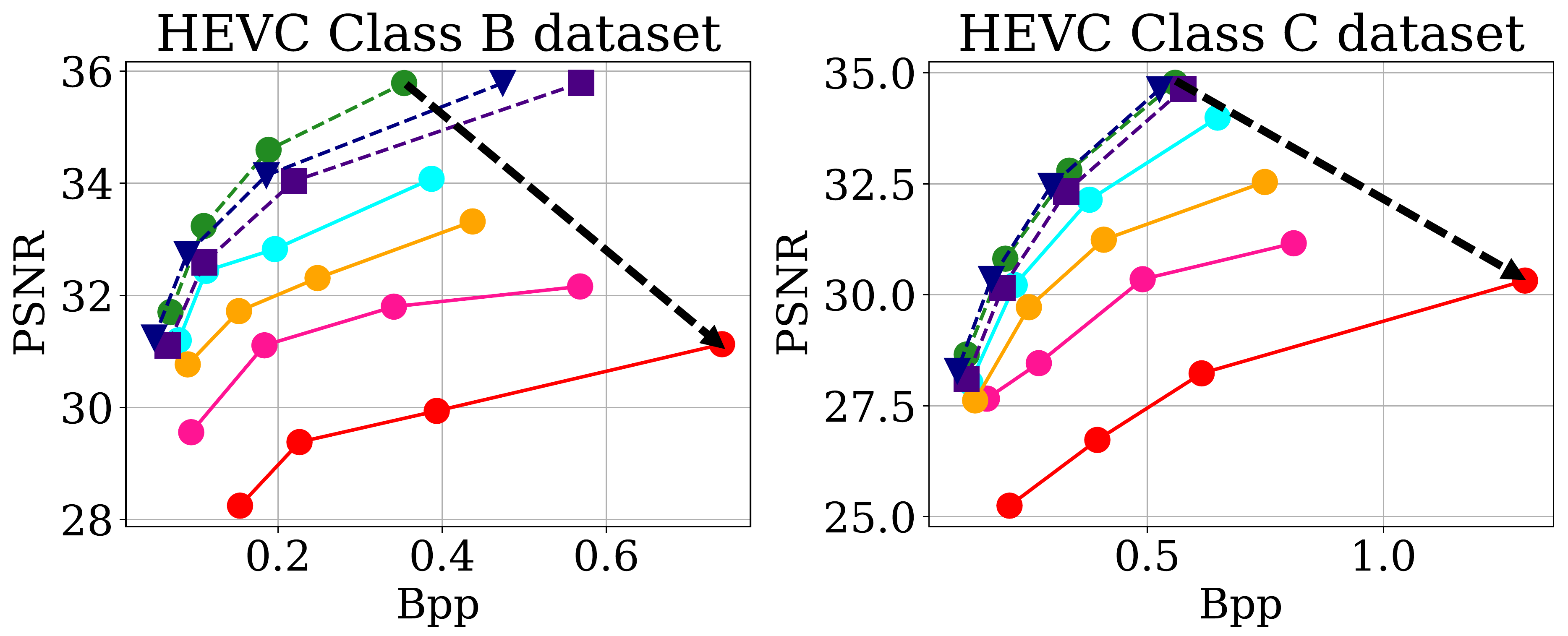} \\
    \footnotesize{Video Quality Attack}&
    \footnotesize{Bandwidth Attack}&
    \footnotesize{RD Attack} \\
    \end{tabular}
    \begin{tabular}{@{}c@{}}
    (b) HLVC~\cite{Yang_2020_CVPR}\\
    \end{tabular}
    \begin{tabular}{@{}ccc@{}}
    \includegraphics[width=0.31\linewidth]{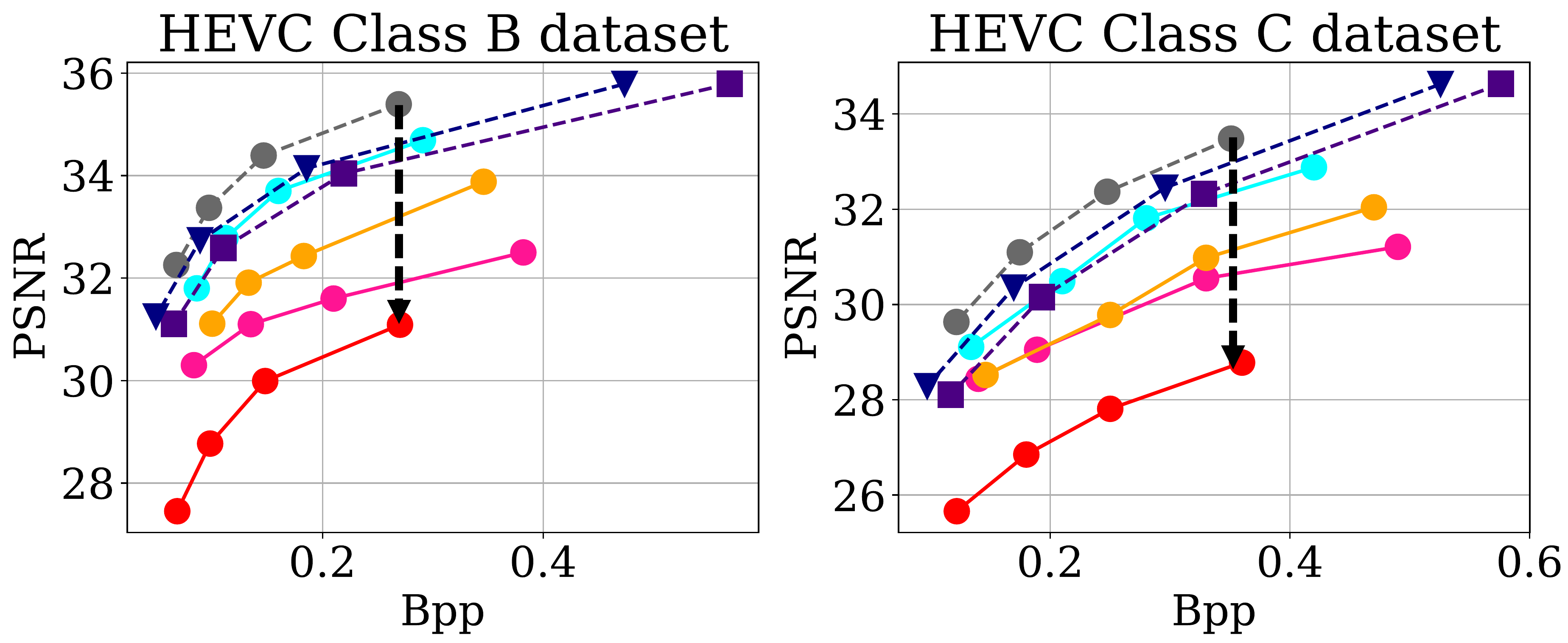} &
    \includegraphics[width=0.31\linewidth]{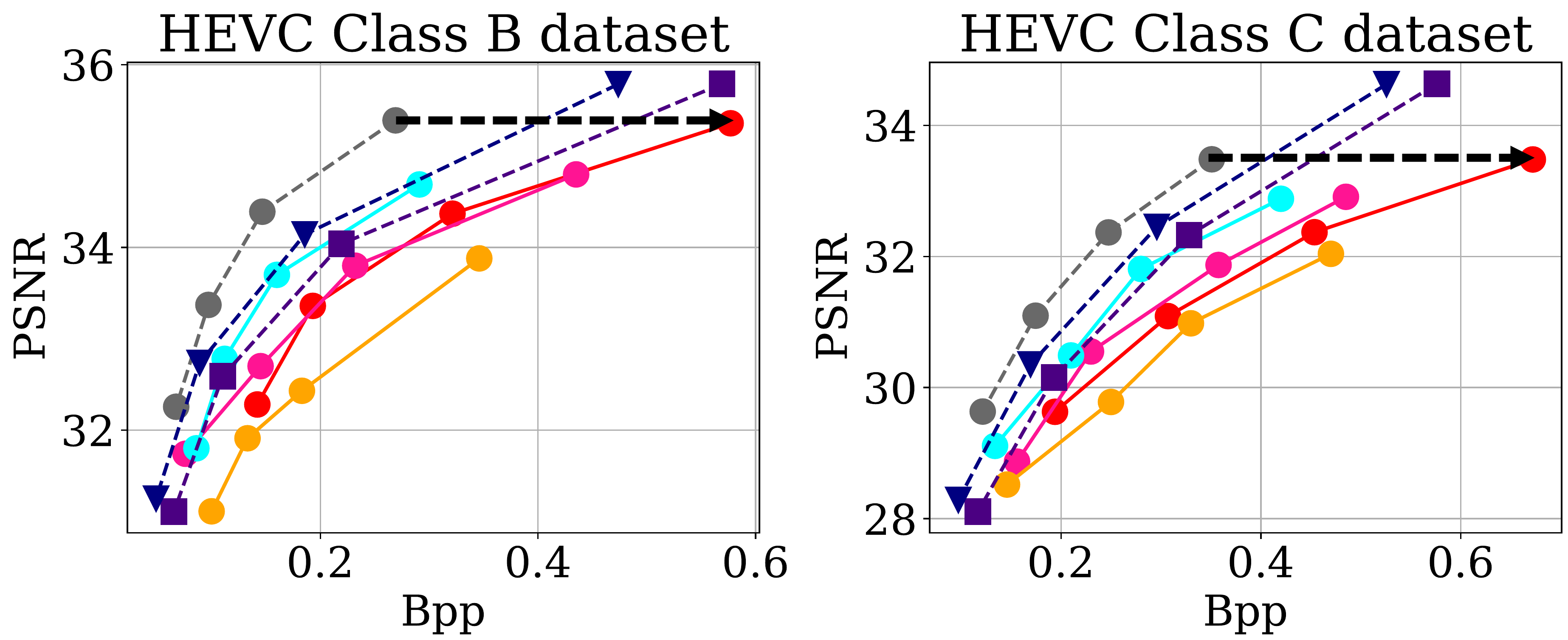}   &
    \includegraphics[width=0.31\linewidth]{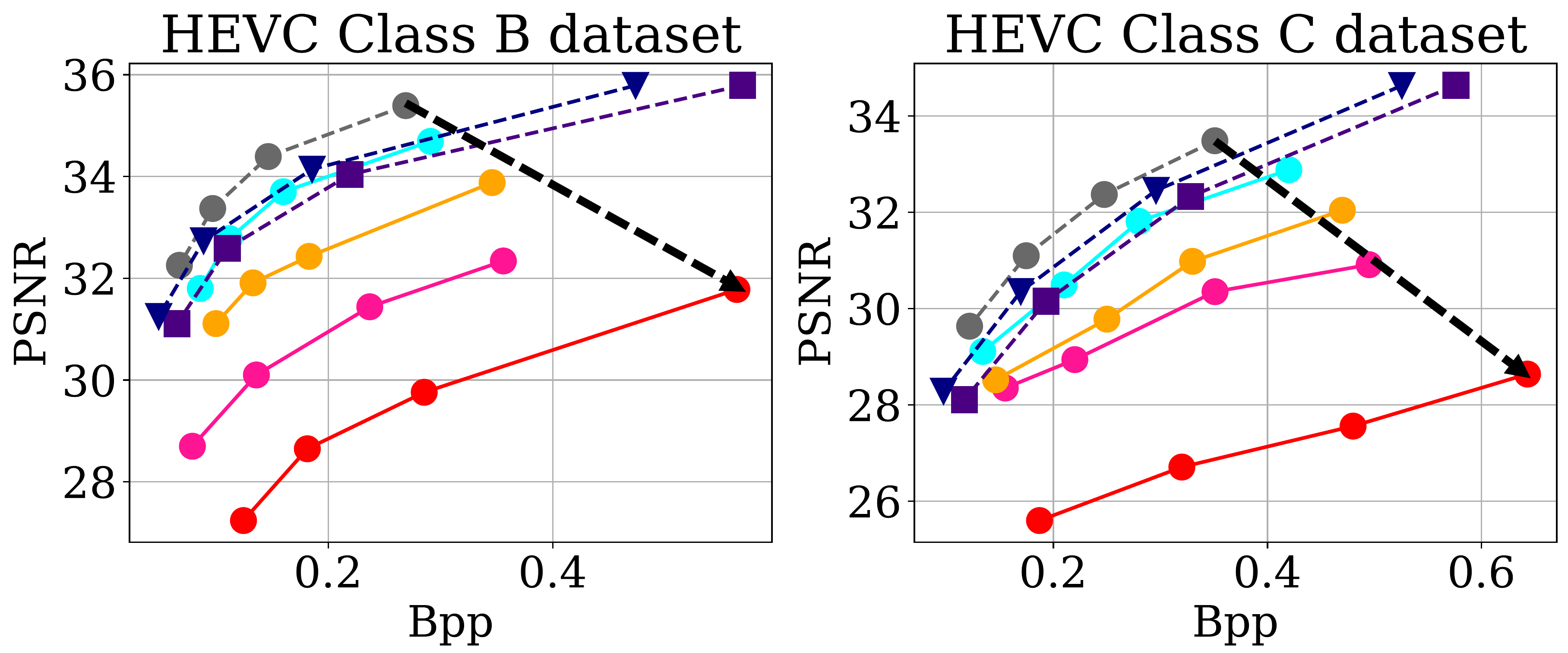} \\
    \footnotesize{Video Quality Attack}&
    \footnotesize{Bandwidth Attack}&
    \footnotesize{RD Attack} \\
    \end{tabular}
    \begin{tabular}{@{}c@{}}
    (c) FVC~\cite{Hu_2021_CVPR}\\
    \end{tabular}
    
  \caption{\edit{White-box} RoVISQ attacks applied to DVC~\cite{Lu_2019_CVPR}, HLVC~\cite{Yang_2020_CVPR}, and FVC~\cite{Hu_2021_CVPR} video compression frameworks. Each graph contains the results for the victim video compression model trained with $\lambda \in \{256,512,1024,2048\}$. \edit{The numerical \textit{R}-\textit{D} values of this Figure are enclosed in Tables~\ref{tab:qoe_table_dvc},~\ref{tab:qoe_table_hlvc},~\ref{tab:qoe_table_fvc} in Appendix~\ref{sec:extra_exps}.}
}
  \label{fig:Exp_qoe_attack}
  \vspace{-0.5cm}
\end{figure*}


\subsection{RoVISQ Attacks on Video Compression} \label{sec:qoe_attacks}
\edit{In this Section, we analyze the effect of our proposed attacks on the performance of video compression. Specifically, we analyze two attack scenarios, namely white-box and black-box settings. In the white-box attack scenario, the victim video compression model is open-source and the attacker has complete knowledge of the underlying DNNs used.}
Figure~\ref{fig:Exp_qoe_attack} demonstrates the PSNR-based \textit{R}-\textit{D} performance evaluated on the HEVC standard test sequences (class B, class C). \edit{The RoVISQ attacks enclosed in this Figure follow the white-box threat model explained above.} Each plot corresponds to a video compression model trained with various values of $\lambda\in\{256, 512,1024, 2048\}$. As $\lambda$ decreases, the bit-rate of the video compression model improves but the video quality deteriorates. We report the results for our video quality, bandwidth, and RD attacks in both offline and online settings. \edit{Intuitively, the online attacks are slightly less powerful than the image-specific offline attacks. It is worth noting that even though the training dataset used for generating the universal perturbation is entirely different from the dataset on which it is tested on, the online attack remains highly effective.} 

We further study the black-box attack scenario, i.e., when the attacker does not know the architecture or weights of the DNN used for video compression. In this setting, we show that our universal perturbations trained on a surrogate open-source video compression model are transferable to unseen systems. Table~\ref{tab:blackbox_qoe} encloses the \textit{R}-\textit{D} performance of the video compression systems before and after applying the black-box online attack. Below we analyze the effects of our various attacks in white-box and black-box scenarios in detail.

\begin{table}[b]
\centering
\caption{\edit{PSNR (dB) and Bpp of victim video compression models on the HEVC class C dataset after applying RoVISQ online perturbations in the black-box attack setting. Reported values are averaged across different encoding parameters $\lambda \in \{256,512,1024,2048\}$. Architectural details of the evaluated models are enclosed in the Appendix Table~\ref{tab:substitute_models}.}} \label{tab:blackbox_qoe}
\resizebox{\columnwidth}{!}{
\begin{tabular}{p{0.1cm}lccccc}
\toprule
&
& \begin{tabular}[c]{@{}c@{}}Video Quality \\ Attack\end{tabular} 
& \begin{tabular}[c]{@{}c@{}}Bandwith \\ Attack \end{tabular} 
& \begin{tabular}[c]{@{}c@{}}RD \\ Attack \end{tabular}
& \begin{tabular}[c]{@{}c@{}}Gaussian \\ Noise \end{tabular} 
\\ \hline
\multirow{2}{*}{M1} 
& PSNR (dB) 
& -2.37
& -0.87        
& -2.46         
& -1.57 \\ \cline{3-6}
& Bpp 
& +18.4\%    
& +32.5\%    
& +29.7\%         
& +17.3\%  \\ \hline
\multirow{2}{*}{M2} 
& PSNR (dB) 
& -2.31      
& -0.92       
& -2.48         
& -1.44 \\ \cline{3-6}
& Bpp
& +19.1\%   
& +30.4\%    
& +27.7\%         
& +17.8\% \\ \hline
\multirow{2}{*}{M3} 
& PSNR (dB) 
& -2.44      
& -0.91        
& -2.55         
& -1.68\\ \cline{3-6}
& Bpp
& +19.5\%    
& +31.7\%     
& +31.1\%         
& +14.8\% \\ \hline
\multirow{2}{*}{M4} 
& PSNR (dB)
& -2.47      
& -0.95   
& -2.51         
& -1.63\\ \cline{3-6}
& Bpp
& +18.6\%    
& +29.4\%    
& +30.2\%         
& +15.2\% \\ \hline
\multirow{2}{*}{M5} 
& PSNR (dB)  
& -2.49      
& -0.88   
& -2.53         
& -1.72\\ \cline{3-6}
& Bpp 
& +17.6\%    
& +32.8\%    
& +30.6\%         
& +17.4\%\\ \hline
\multirow{2}{*}{M6} 
& PSNR (dB)  
& -2.38      
& -0.98        
& -2.36         
& -1.65\\ \cline{3-6}
& Bpp 
& +18.3\%    
& +31.4\%    
& +32.1\%         
& +17.8\% \\ \bottomrule
\end{tabular}}
\end{table}

\begin{table}[t]
\centering
\caption{\cameraready{PSNR and Bpp of conventional compression methods after applying RoVISQ universal perturbations, evaluated on the HEVC class C dataset.}} \label{tab:conventional_attack}
\resizebox{1.0\columnwidth}{!}{
\begin{tabular}{clcccc}
\hline
& & \begin{tabular}[c]{@{}c@{}}Video Quality \\ Attack\end{tabular}
& \begin{tabular}[c]{@{}c@{}}Bandwith \\ Attack\end{tabular}
& \begin{tabular}[c]{@{}c@{}}RD \\ Attack\end{tabular} 
& \begin{tabular}[c]{@{}c@{}}Gaussian \\ Noise\end{tabular} 
\\ \hline
\multirow{2}{*}{\begin{tabular}[c]{@{}c@{}}PSNR\\  (dB)\end{tabular}}
& \multicolumn{1}{c}{H.265} 
& -3.47      
& -1.55        
& -3.62         
& -1.71 \\ \cline{2-6}
& \multicolumn{1}{c}{H.264} 
& -3.19      
& -1.03    
& -3.48         
& -1.31 \\ \hline
\multirow{2}{*}{Bpp}
& \multicolumn{1}{c}{H.265} 
& +45.5\%           
& +78.4\%        
& +73.8\%         
& +62.1\% \\ \cline{2-6}
& \multicolumn{1}{c}{H.264} 
& +34.7\%      
& +65.2\%    
& +61.8\%         
& +45.9\% \\ \hline
\end{tabular}}
\end{table}



\noindent\textbf{Analysis of the Video Quality Attack.} 
As shown in Figure~\ref{fig:Exp_qoe_attack}-left, RoVISQ attacks can break the \textit{R}-\textit{D} model of pre-trained video encoders by a large margin. Specifically, our attacks lower the video quality by up to 5.43dB when $\lambda=2048$ while maintaining the Bpp level. Our attack performance improves when $\lambda$ is bigger. This is because the bit-rate increases with $\lambda$ and therefore it is less likely for the perturbations to be removed during compression. Figure~\ref{fig:Exp_qoe_attack} also shows that video compression is vulnerable to our universal online attack. \edit{When applying the online perturbation in the white-box scenario, the PSNR drops by 2.30-3.24dB. When applying the universal perturbation to unseen victim models, i.e., in the black-box setting, the average PSNR drop is 2.41dB which is only 0.41dB lower than the average PSNR drop of the white-box attack. This shows that RoVISQ online perturbations can be used to effectively attack any arbitrary video compression model.}


\noindent\textbf{Analysis of the Bandwidth Attack.}
Figure~\ref{fig:Exp_qoe_attack}-middle demonstrates the performance of our proposed bandwidth attack in the white-box setting. As seen, our adversarial perturbations successfully reduce the compression rate while maintaining the PSNR of the benign models to hide the attack. Gaussian noise deviates a lot from the \textit{R}-\textit{D} curve of the original models and also increases the Bpp level much less than our bandwidth attack. In order to maintain $D$, the perturbed video should not be much different from before. However, Gaussian noise differs by up to 2.39dB from the original video quality, which is considerable. When compared with the benign compression models, our proposed offline bandwidth attack dramatically increases the Bpp level by up to 2.12$\times$, 2.03$\times$, and 2.37$\times$ on the HEVC class B, C, and D while reducing the PSNR by only 0.01dB. \edit{Similarly, our white-box online attack increases the Bpp level by up to 1.61$\times$ while maintaining the PSNR difference within 0.55dB. Our online attack increases the Bpp of unseen models by up to 1.52$\times$, which is on average only 5.6$\%$ lower than the Bpp of the white-box scenario.}

\noindent\textbf{Analysis of the RD Attack.}
Figure~\ref{fig:Exp_qoe_attack}-right shows the performance of our white-box \rdo. We confirm that \rdo can achieve all the objectives of the video quality and bandwidth attacks. When compared with the video quality attack, \rdo lowers the PSNR by 0.44dB more, on average. When compared to the bandwidth attack, the increase in the Bpp is on average 6.3$\%$ lower. This is due to the inherent trade-off between the bit-rate and distortion in the perturbation loss (Equation~\ref{eq:video_compression}), which is controlled by $\lambda$; a higher value of $\lambda$ shifts the optimization towards lower PSNR, thereby resulting in a higher Bpp compared to the bandwidth attack.  \edit{When applying our online RD perturbations to unseen models (black-box attack), the average PSNR drop and Bpp increase are -2.48dB and 30.2$\%$, compared to the -3.41dB and 37.5$\%$ achieved by the white-box online attack. }

\noindent\textbf{Comparison with Baselines.}
As shown in Figure~\ref{fig:Exp_qoe_attack}, our proposed perturbations outperform the baselines by a large margin, decreasing the PSNR by up to 4.52dB, or increasing the BPP by 2.12$\times$. Without considering the QoE factors in the optimization problem, the evaluated baselines cannot critically hurt video compression. Especially, while Gaussian Noise Case II outperforms case I, it contains twice the noise level compared to our perturbation in the I frames. Nevertheless, our perturbations achieve a much higher attack performance by leveraging the QoE factors of video compression.


\noindent\textbf{Attack on Conventional Compression Methods.}
\cameraready{We were curious to know whether our trained universal perturbations are also effective against conventional, non-DNN-based, compression methods, i.e., H.264~\cite{H.264} and H.265~\cite{H.265}. Therefore, in addition to DNN-based video compression models, we show that our online perturbations can also be used for black-box attack on traditional video compression methods, i.e., H.264~\cite{H.264} and H.265~\cite{H.265}. As shown in Table~\ref{tab:conventional_attack}, the attack performance decreases for conventional video-compression methods compared to DNN-based systems evaluated in Figure~\ref{fig:Exp_qoe_attack}. However, the attack performance still outperforms the noise baseline by a large margin. It can be concluded that RoVISQ perturbations automatically learn to target low-level and intrinsic properties of video compression which are common between DNN-based and conventional methods. Our perturbations are thus transferable and can downgrade video compression performance even when it is not performed by a DNN. }

\subsection{\edit{Effect of RoVISQ Attacks on User Experience}}\label{sec:user_study}

\begin{figure}[t]
\centering
    \begin{subfigure}[b]{\columnwidth}
        \begin{tabular}{@{}cc@{}}
        \includegraphics[width=0.48\linewidth]{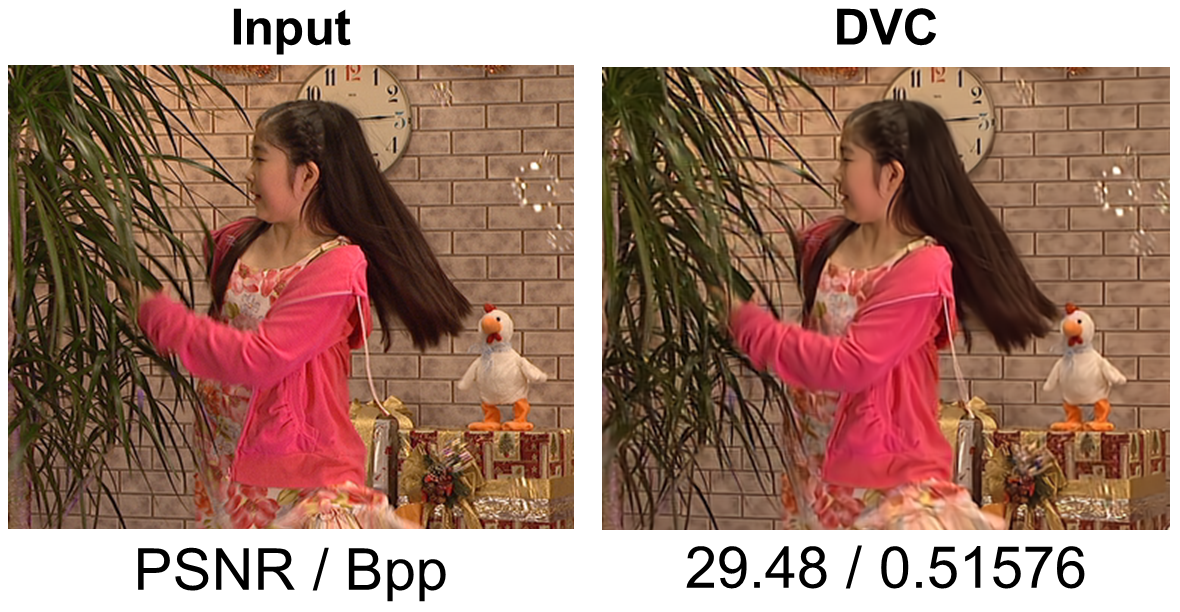} &
        \includegraphics[width=0.48\linewidth]{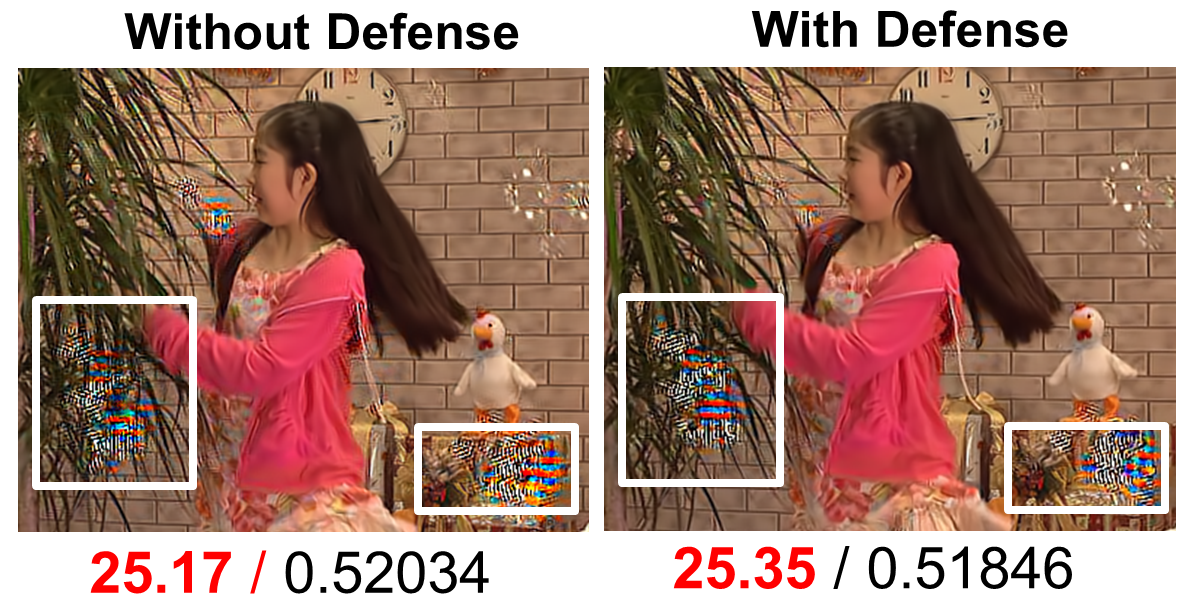} \\
        (a) \footnotesize{No Attack} &
        (b) \footnotesize{Video Quality Attack} \\
        \end{tabular}  
    \end{subfigure}
    \begin{subfigure}[b]{\columnwidth}
        \begin{tabular}{@{}cc@{}}
        \includegraphics[width=0.48\linewidth]{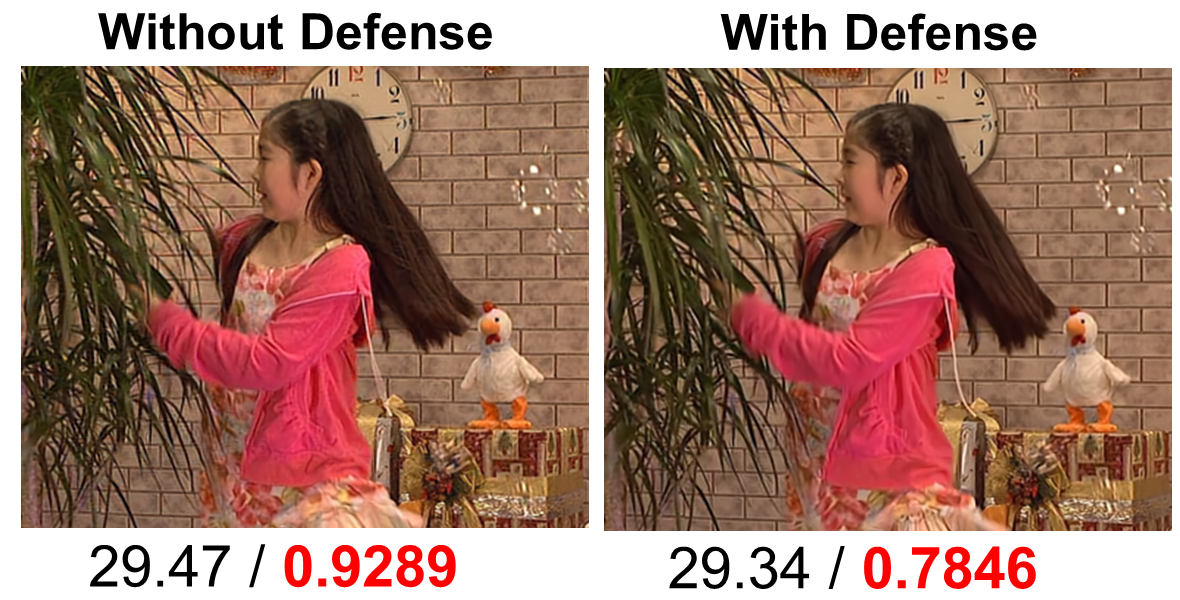}   &
        \includegraphics[width=0.48\linewidth]{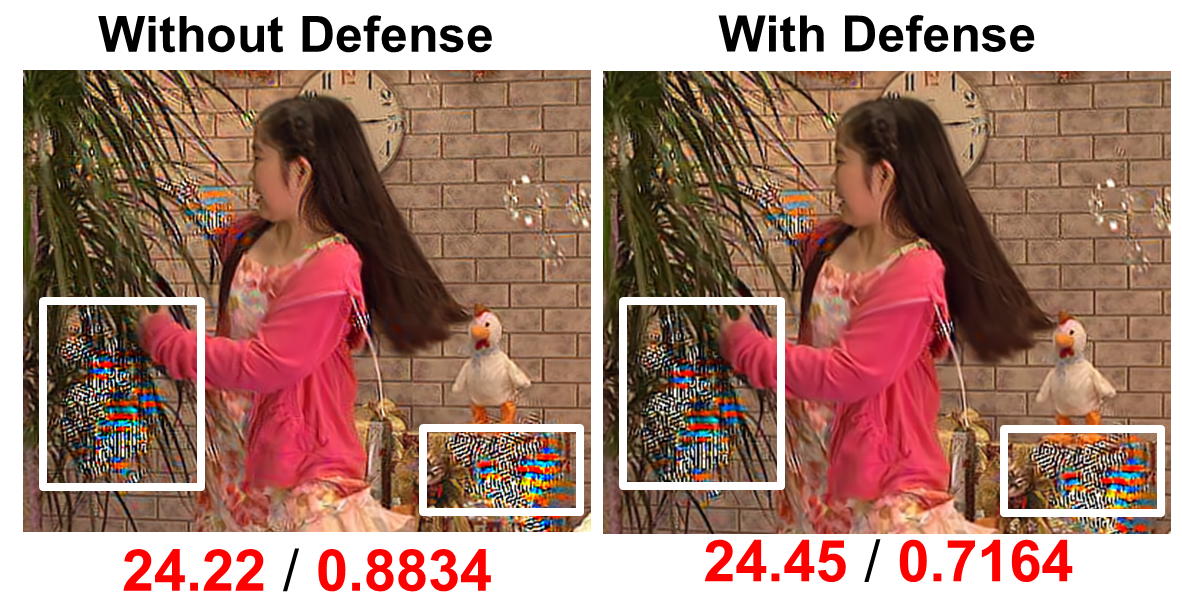} \\
        (c) \footnotesize{Bandwidth Attack} &
        (d) \footnotesize{RD Attack}\\
        \end{tabular}  
    \end{subfigure}
  \caption{Visual comparison of decoded frames when the input video is clean versus when it is perturbed by RoVISQ attacks. We further show how our perturbations are robust against a DNN-based video denoiser~\cite{claus2019videnn} when applied to the adversarial video as a defense.}
  \label{fig:Exp_qoe_pic}
\vspace{-0.2cm}
\end{figure}

\noindent\textbf{Attack Visualization.} Figure \ref{fig:Exp_qoe_pic} shows an example of RoVISQ attacks against DVC video compression. As seen, the performance of the video coder is significantly degraded in terms of either PSNR or Bpp, after compressing the perturbed video. Specifically, the benign compressed video and the adversarial video perturbed with the bandwidth attack show similar image quality after compression, although the compression ratio differs by 1.8$\times$. We can also see that the video quality attack keeps the Bpp constant but adds several noise-like artifacts to the compressed video that degrade the visual quality. \rdo generates more distortion in the decoded frame compared to the video quality attack while requiring a higher Bpp (although lower than the bit-rate achieved by the bandwidth attack). \edit{We note that the disturbance observed in the video is not uniform for all frames: the disturbance is higher for frames which have a low inter-frame correlation, such as those visualized in Figure\ref{fig:Exp_qoe_pic}-b,d. It is inherently difficult to perform spatiotemporal predictions on dynamic video sequences containing fast moving objects or sudden scene changes~\cite{zhou2020rate}. Due to inaccurate motion compensation, the magnitude of the generated residual is large and the contextual information of the residual frame is largely lost after video compression. When our perturbation is added to such dynamic videos, the quality of the decoded frame is further deteriorated due to error propagation and accumulation of the decoded error.}

\begin{figure}[t]
    \centering
    \includegraphics[width=\columnwidth]{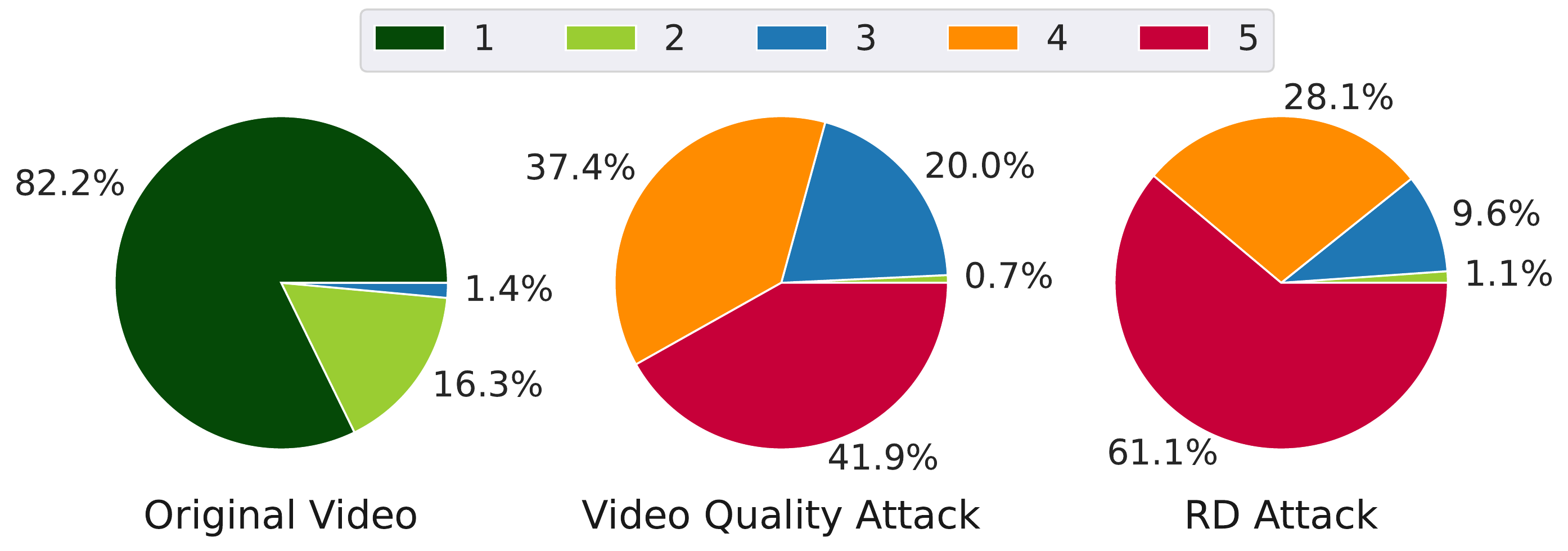}
    \caption{\edit{Distribution of scores received from participants in our survey for each category of videos. Here, a score of $1$ indicates no perceptible noise in the video while a score of $5$ corresponds to highly perceptible noise with drastic video quality reduction.}}
    \label{fig:survey}
\vspace{-0.2cm}
\end{figure}


\begin{figure*}[h]
\centering
  \begin{tabular}{@{}c@{}}
    \includegraphics[width=0.6\linewidth]{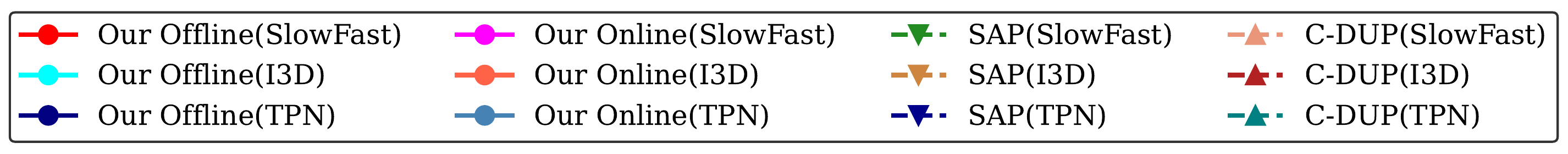} \\  
  \end{tabular}
    \begin{tabular}{@{}cc@{}}
    \includegraphics[width=0.42\linewidth]{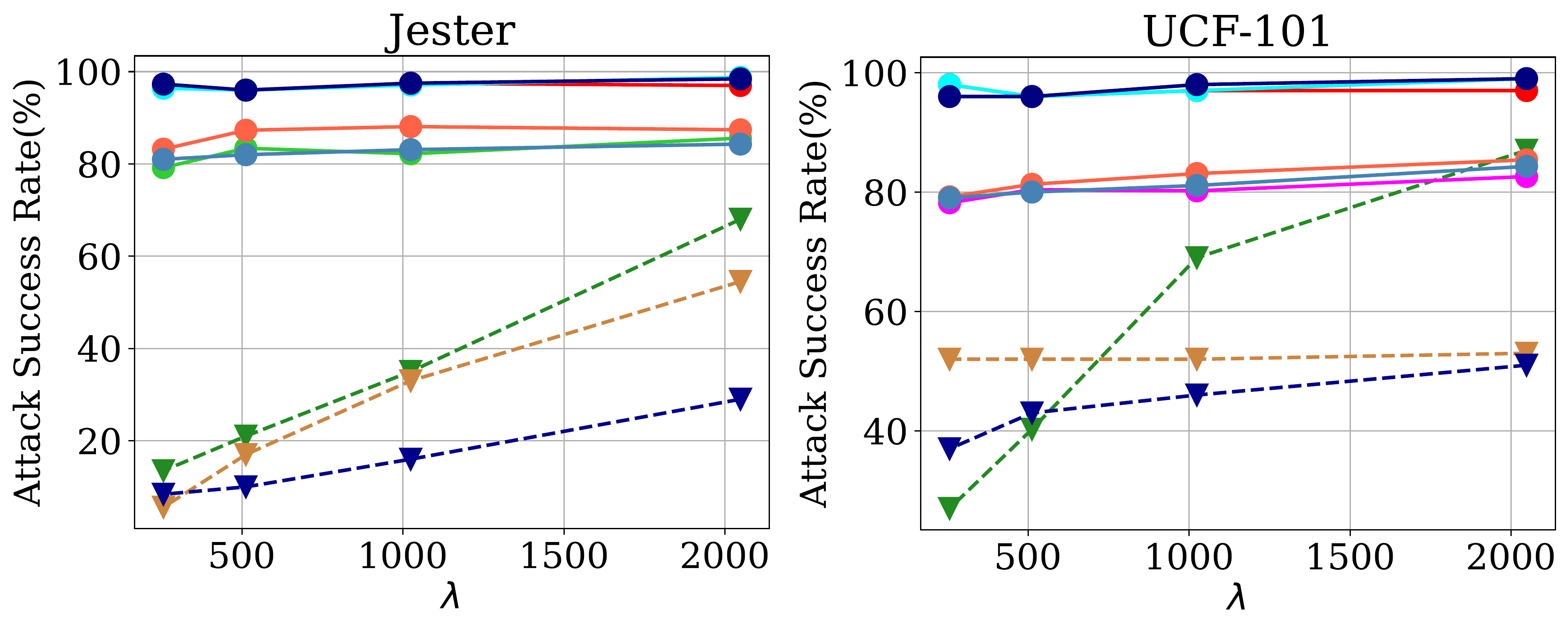}   &
    \includegraphics[width=0.42\linewidth]{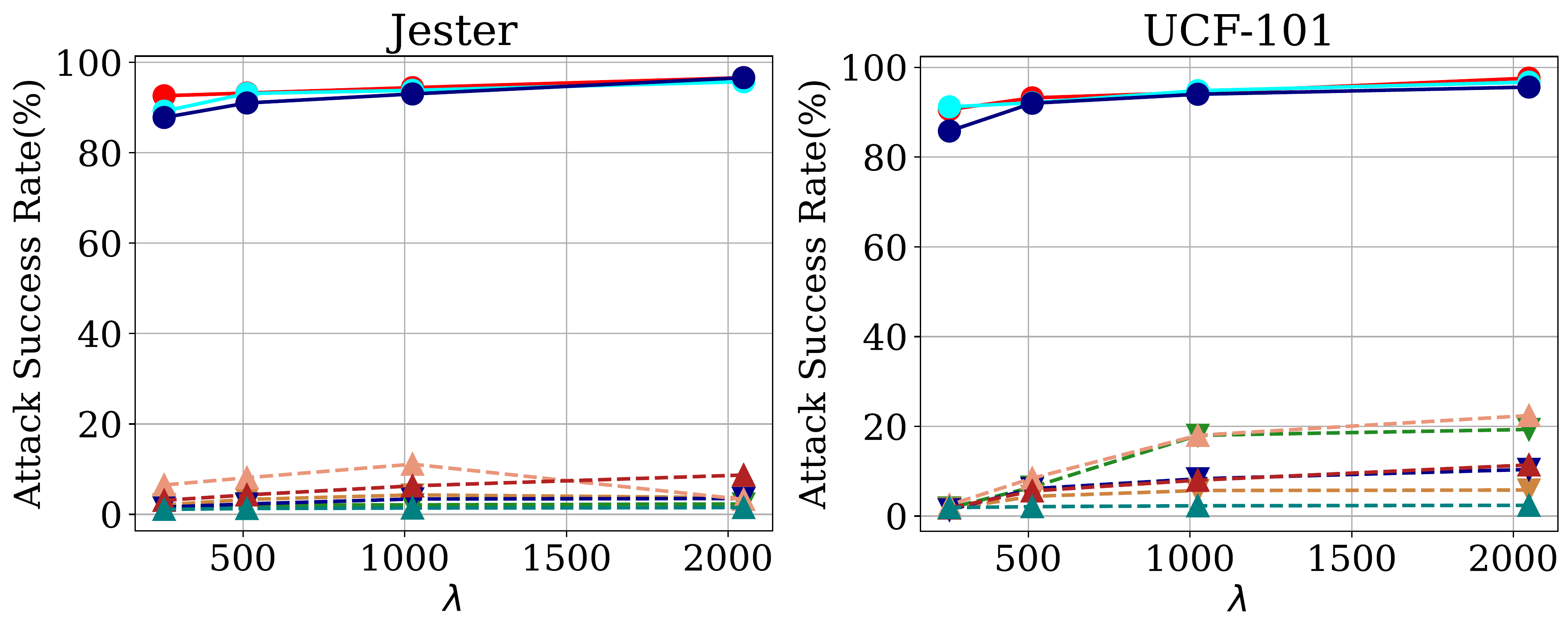} \\
    (a) Untargeted Attack&
    (b) Targeted Attack\\
    \end{tabular}  
  \caption{\edit{Attack success rate of our compression-robust classifier perturbations compared to state-of-the-art attacks on video classification, SAP~\cite{Wei_aaai_2019} and C-DUP~\cite{Li_2019_NDSS}. Reported values are for the bandwidth attack conducted in the white-box scenario. We benchmark three video classification models, I3D \cite{Carreira_2017_CVPR}, SlowFast \cite{Feichtenhofer_2019_ICCV}, and TPN \cite{TPN_2020_CVPR}. The perturbations in prior work cannot withstand compression. RoVISQ perturbations, however, maintain a high success rate even after video compression.}
}
  \label{fig:Exp_asr_vcs}
  \vspace{-0.5cm}
\end{figure*}

\noindent\textbf{User Study.} \edit{We conduct a public survey to study the effect of our attacks on users' QoE. We recruited 55 volunteers aged between 20-40 from different backgrounds, i.e., college students, graduate students, postdoctoral researchers, and industry professionals. We chose the background and age range such that the target demographic has relatively high exposure to various video streaming services, e.g., YouTube, on a daily basis.
Each user is shown video sequences in HEVC class B and class C. Each set contains three video clips: 1)~the original benign video, 2)~the video with RoVISQ's video quality attack, and 3)~the video attacked using RoVISQ's RD perturbations\footnote{\edit{The video clips used in the user survey are visible on the project website ((\url{https://sites.google.com/view/demo-of-rovisq/home})).}}.} 

\edit{The participants were asked to rate the noise level for each video clip based on a Likert scale~\cite{likert1932technique} from $1$ to $5$ where a score of $1$ corresponds to no perceptible noise and a score of $5$ shows highly perceptible noise. As clarified in the survey, a score of $5$ would mean the user is likely to abandon the streaming service due to high levels of noise disrupting the video quality. The category of each video clip, i.e., benign versus video quality or RD attacks, is not revealed to the users and the videos are not annotated in any way. To avoid any bias in the gathered answers, the participants are not aware of RoVISQ methodology and the overarching goal of the paper. Specifically, the users do not know which videos are generated by RoVISQ, whether RoVISQ is a defense or an attack, and what the nominal score for each video clip is.}

\edit{Figure~\ref{fig:survey} shows the distribution of scores selected by the participants for the original video clips as well as the attacked versions. As seen, participants assign low noise scores to benign videos, with a score of $1$ appearing in 82.2$\%$ of the responses. This result establishes the baseline for video quality: the users unanimously agree that the quality of the benign videos included in the survey are adequately high. Once RoVISQ attacks are applied to the same videos, however, we can see a clear shift in the distribution of scores to higher values (more visible noise and lower video quality). For the video quality attack, the distribution of the $4$ and $5$ scores is nearly the same, constituting 37.4$\%$ and 41.9$\%$ of the participant answers. For the \rdo, the portion of scores equal to $5$ increases to 61.1$\%$. This is intuitive since \rdo drops the PSNR by a larger margin compared to the video quality attack (see Figure~\ref{fig:Exp_qoe_attack}). Observing a noise score of $5$ in the majority of answers validates that our proposed attacks successfully degrade users' QoE, leading to denial or interruption of service.}

\subsection{RoVISQ Attacks on Video Classification}\label{sec:exps_classification}

We evaluate the success rate of our bandwidth attack\footnote{\edit{Results for video quality and \rdos are provided in Appendix~\ref{sec:appdx_classification_attack}.}} when directed towards a downstream video classifier and provide comparisons with state-of-the-art attacks on video classification. We define the attack success rate as the portion of misclassified test samples for the untargeted attack. For the targeted attack, success rate is the portion of samples mapped to the adversary's desired class. 
The convergence curves for generating targeted and untargeted perturbations using our losses are shown in Appendix~\ref{sec:convergence} Figure \ref{fig:Exp_qoe_on_vcs_appdx}.
The attacks in this section not only affect the downstream video classifier but also degrade the
performance of the video compression system. We provide an analysis of this effect in Appendix~\ref{sec:appdx_classification_attack}.

\edit{We first evaluate our attacks in the white-box scenario, i.e., when the attacker has complete knowledge of the underlying video classification model. Figure~\ref{fig:Exp_asr_vcs} demonstrates the attack success rate for white-box RoVISQ attacks, along with two baseline white-box attacks on video classification, namely, SAP~\cite{Wei_aaai_2019} and C-DUP~\cite{Li_2019_NDSS}. As seen, our attack consistently achieves the highest success rate.} In particular, we obtain over 90$\%$ success rate on the UCF-101 and Jester datasets. This is in contrast to prior works, which largely lose their performance once the video is compressed, especially in that targeted scenario. We additionally consider an online attack scenario, where video data is continuously generated and a video clip is classified in real time from the generated video sequence. Even in the complex real-time scenario, our universal perturbations can be injected into both UCF-101 and Jester to achieve a high success rate up to 85.4$\%$ and 87.5$\%$, respectively. 

\begin{figure}[h]
\centering
    \begin{subfigure}[b]{\columnwidth}
    \centering
    \includegraphics[width=\linewidth]{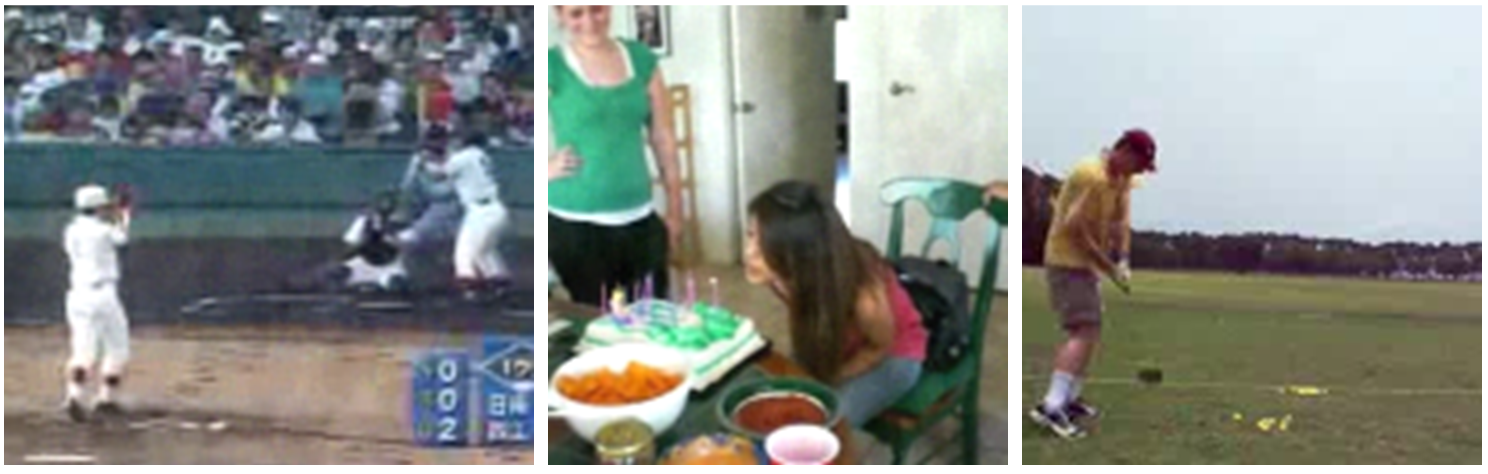}
    \caption{Adversarial Input Video Clips}
    \end{subfigure}
    \begin{subfigure}[b]{\columnwidth}
    \includegraphics[width=\linewidth]{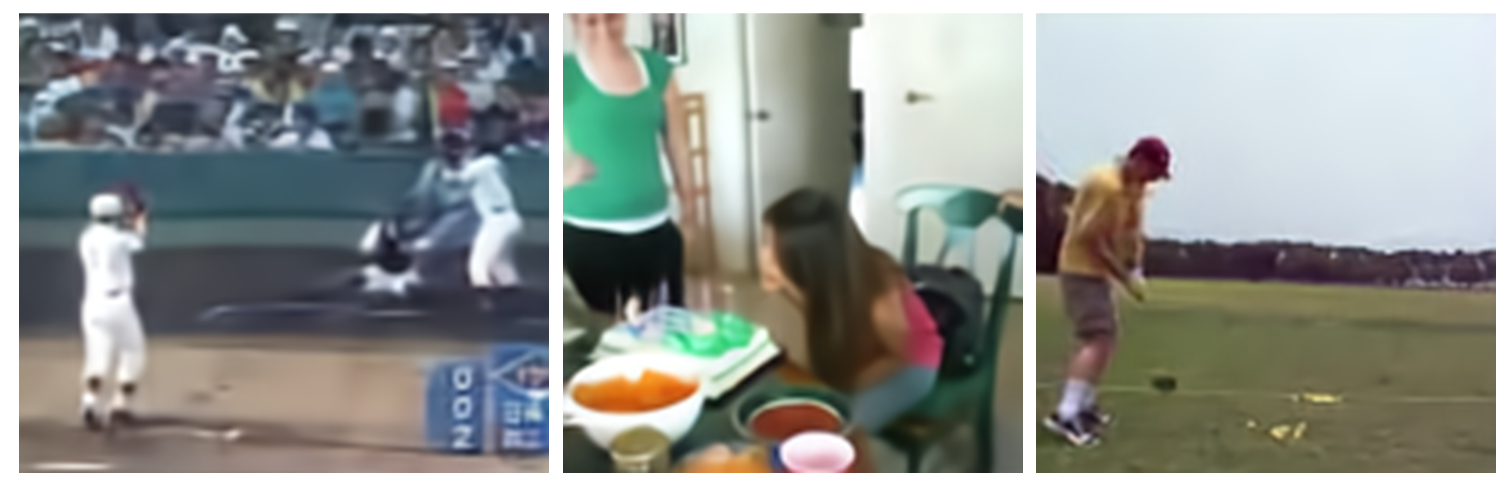}
    \caption{Adversarial Decoded Video Clips}
    \end{subfigure}
    \vspace{-0.5cm}
  \caption{Example video clips before and after applying compression. Compressed videos lose some texture information and high frequency components after coding.}
  \label{fig:Exp_asr_vcs_appdx}
\end{figure}

To explain why RoVISQ outperforms other baselines, we visualize some example video snapshots before and after compression in Figure~\ref{fig:Exp_asr_vcs_appdx}.
As observed in Figure \ref{fig:Exp_asr_vcs_appdx}-(b), the compressed video clips lose some texture information and high-frequency components when compared with the original perturbed video in Figure \ref{fig:Exp_asr_vcs_appdx}-(a). This effect hinders the attack success rate of prior works, causing the perturbations to be removed after compression.



\edit{We further analyze the performance of RoVISQ attacks in the black-box scenario, i.e., when the attacker does not have access to the video classifier. In this setting, we train our universal perturbations on a surrogate video classifier, and use the trained perturbations to attack an unseen victim model. The attack success rate obtained by RoVISQ and two state-of-the-art black-box attacks on video classification are summarized in Table~\ref{tab:classification_blackbox}. As shown, the proposed adversarial perturbations are transferable to unseen video classification models, outperforming previous attacks, namely, GeoTrap~\cite{Li_NIPS_2021} and U3D~\cite{Xie_SP_2022}, by 27.8-66.8$\%$ and 22.2-65.8$\%$, respectively. }

\setlength{\tabcolsep}{3.5pt}
\begin{table}[b]
\caption{\edit{Performance of RoVISQ bandwidth attack compared with prior work on adversarial video classification. Attacks are conducted in the black-box untargeted scenario on the Jester dataset. The names inside parentheses in the attack column are the surrogate video classifiers used to train the RoVISQ universal perturbations. }}\label{tab:classification_blackbox}
\resizebox{\columnwidth}{!}{
\begin{tabular}{clclll}
\toprule
\multicolumn{1}{c}{\multirow{2}{*}{\begin{tabular}[c]{@{}c@{}}Victim \\ Model\end{tabular}}}
& \multirow{2}{*}{Attack}
& \multicolumn{4}{c}{Attack Success Rate (\%)} \\ \cline{3-6}
\multicolumn{1}{c}{}                              
&                                   
& \multicolumn{1}{c}{$\lambda=256$} 
& \multicolumn{1}{c}{$512$} 
& \multicolumn{1}{c}{$1024$} 
& \multicolumn{1}{c}{$2048$} \\ \hline
\multirow{4}{*}{\begin{tabular}[c]{@{}c@{}}
TPN \\ \cite{TPN_2020_CVPR}\end{tabular}} 
& GeoTrap~\cite{Li_NIPS_2021}                                          
& 6.4
& 16.8
& 18.5
& 32.4  \\ \cline{2-6}
& U3D~\cite{Xie_SP_2022}                                          
& 7.4                                
& 17.5                                
& 19.4
& 36.1 \\ \cline{2-6}
& Bandwidth (I3D)                   
& 71.3                        
& 76.9                     
& 79.6                        
& \textbf{82.4}\\
& Bandwidth (SlowFast)
& \textbf{73.2}                      
& \textbf{77.8}                       
& \textbf{80.6}                      
& 81.5\\ \hline
\multirow{4}{*}{\begin{tabular}[c]{@{}c@{}}
SlowFast\\ \cite{Feichtenhofer_2019_ICCV}
\end{tabular}} 
& GeoTrap~\cite{Li_NIPS_2021}
& 11.2                                
& 22.2
& 38.9
& 54.6  \\ \cline{2-6}
& U3D~\cite{Xie_SP_2022}
& 10.2                                
& 24.1                                
& 37.0
& 60.2 \\ \cline{2-6}
& Bandwidth (I3D)                        
& 73.2                       
& \textbf{76.9}                    
& 78.7                   
& 81.5 \\
& Bandwidth (TPN)                 
& \textbf{74.1}                    
& 75.0                    
& \textbf{80.6}                    
& \textbf{82.4}\\ \hline
\multirow{4}{*}{\begin{tabular}[c]{@{}c@{}}
I3D\\ \cite{Carreira_2017_CVPR}
\end{tabular}} 
& GeoTrap~\cite{Li_NIPS_2021}
& 8.3                               
& 24.1
& 41.7
& 42.6  \\ \cline{2-6}
& U3D~\cite{Xie_SP_2022}
& 6.5                               
& 16.7                                
& 39.8
& 48.1 \\ \cline{2-6}
& Bandwidth (SlowFast)    
& 70.4             
& \textbf{76.9 }            
& \textbf{81.5}               
& \textbf{83.3} \\
& Bandwidth (TPN)                       
& \textbf{72.2}                        
& 74.1                       
& 76.9                    
& 80.6 \\ \bottomrule                             
\end{tabular}}
\end{table}


\vspace{-0.2cm}
\section{Resiliency to Defense Schemes}
\label{sec:experiments_uni}
\begin{figure}[t]
    \centering
    \includegraphics[width=\columnwidth]{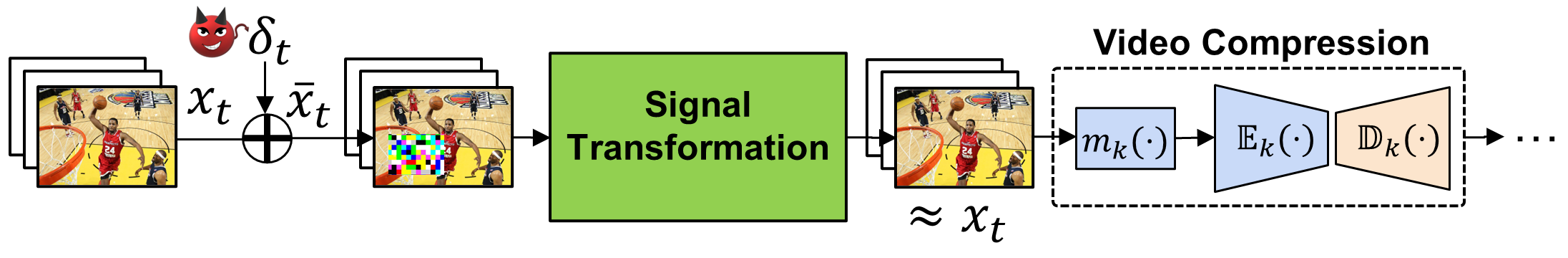}
    \vspace{-0.1cm}
    \caption{Overview of the evaluated defenses. The signal transformation aims at reconstructing the original signal while removing the adversarial perturbations.}
    \label{fig:defense}
\end{figure}

In this section, we comprehensively evaluate different defense mechanisms against RoVISQ attacks. There are very few defenses available for adversarial video classification. We evaluate the recent defense proposed in~\cite{lo2020defending} which relies on adversarial training (AT). We make slight modifications to~\cite{lo2020defending} to make it applicable to our attack scenario. In addition, we implement new defense mechanisms that rely on signal transformations to remove adversarial perturbations as shown in Figure~\ref{fig:defense}. For this set of defenses, we are inspired by the literature in image and audio domains~\cite{Jia_2019_CVPR,lin2019defensive, khalid2019qusecnets,hussain2021waveguard}. Signal transformation aims at removing the adversarial perturbations from the input video and obtain a clean signal that is similar to the original benign input. We evaluate two signal transformations, namely, JPEG compression \cite{Jia_2019_CVPR,aydemir2018effects,8953640,8416586} and the state-of-the-art video denoising method \cite{claus2019videnn}. We use DVC~\cite{Lu_2019_CVPR} compression and the HEVC (class C) and UCF-101 datasets for the defense evaluations.  

\subsection{Adversarial Training}\label{sec:AT}
The performance of AT depends on the adversary's knowledge regarding the attack algorithms. We assume that the defender has complete white-box knowledge about the perturbations crafted by our attacks. AT expands the training dataset by including all of the adversarial examples generated by the adversary and trains the video compression model on the augmented training dataset. Specifically, the defender adds perturbed videos from every $\lambda \in\{256, 512, 1024, 2048\}$ for each RoVISQ attack (bandwidth, video quality, and RD) to the clean training dataset. Due to space limitations, we present the results for selected benchmarks in what follows. 

\noindent\textbf{Defense for Video Compression.} As shown in Table~\ref{tab:qoe_defense_model_table}, the incorporation of adversarial images inside the training comes at a cost of $18.5\%$ higher Bpp and $1.64$dB lower PSNR even if the underlying victim model is not attacked. Moreover, the adversarial training cannot protect the video compression model against RoVISQ attacks. As an example, for the offline attack in Table~\ref{tab:qoe_defense_model_table}, the video quality attack to the benign model would drop the PSNR by $3.5$dB. This drop is reduced to $2.5$dB with the adversarial training. However, due to the original $1.64$dB drop of PSNR after adversarial training, the total effect on PSNR is even worse than the no defense scenario. For the bandwidth attack, the Bpp is increased by $99.5\%$ when no defense is present and $88.5\%$ with adversarial training. However, the original $18.5\%$ increase in Bpp due to adversarial training defeats the purpose of the defense. Similar trends can be observed for \rdo. We include additional variations of the datasets evaluated against AT in Appendix Table~\ref{tab:qoe_defense_dvc_table}.

\setlength{\tabcolsep}{3pt}
\begin{table}[h]
\centering
\caption{PSNR and Bpp of adversarially trained (AT)\cite{madry2017towards} DVC on the HEVC class C Dataset.} \label{tab:qoe_defense_model_table}
\resizebox{0.95\columnwidth}{!}{
\begin{tabular}[t]{lcccc}
\toprule
\multirow{2}{*}{Benchmark} 
& \multicolumn{2}{c}{w Defense}
& \multicolumn{2}{c}{w/o Defense} \\
& PSNR (dB) & Bpp
& PSNR (dB) & Bpp \\
\midrule
DVC~\cite{Lu_2019_CVPR}
& 29.22 & 0.34 
& 31.24 & 0.27 \\ \hline
\begin{tabular}[c]{@{}c@{}} Video Quality (Offline)\end{tabular} 
& -2.41 & +0.6$\%$
& -3.52 & +0.7$\%$\\
\begin{tabular}[c]{@{}c@{}} Video Quality (Online)\end{tabular}
& -2.51 & +16.4$\%$ 
& -3.05 & +19.9$\%$\\ \hline
\begin{tabular}[c]{@{}c@{}} Bandwidth (Offline)\end{tabular}
& -0.12 & +84.2$\%$ 
& -0.01 & +99.4$\%$\\
\begin{tabular}[c]{@{}c@{}} Bandwidth (Online)\end{tabular}
& -0.75 & +31.5$\%$ 
& -0.39 & +35.7$\%$\\ \hline
\begin{tabular}[c]{@{}c@{}} RD (Offline)\end{tabular}
& -2.88 & +71.5$\%$ 
& -4.21 & +85.3$\%$\\
\begin{tabular}[c]{@{}c@{}} RD (Online)\end{tabular}
& -2.41 & +25.6$\%$ 
& -3.10 & +33.5$\%$\\
\bottomrule
\end{tabular}}
\end{table}%

\begin{table}[h]
\centering
\caption{Effect of various defense methods on the accuracy (ACC) of video classifiers on clean data along with the targeted attack success rate (ASR). Results are gathered on UCF-101 dataset. } \label{tab:target_defense_model_table}
\resizebox{\columnwidth}{!}{
\begin{tabular}[t]{clcccc}
\toprule
{\begin{tabular}[c]{@{}c@{}}Video \\ Classifier\end{tabular}}
& Defense 
& \begin{tabular}[c]{@{}c@{}}ACC (\%)\\ w/o Defense\end{tabular}
& {\begin{tabular}[c]{@{}c@{}}ACC \\ Drop (\%)\end{tabular}}
& {\begin{tabular}[c]{@{}c@{}}ASR (\%)\\ w Defense\end{tabular}} 
& \begin{tabular}[c]{@{}c@{}}ASR (\%)\\ w/o Defense\end{tabular} \\
\midrule
{\multirow{3}{*}{\begin{tabular}[c]{@{}c@{}}
SlowFast\\ \cite{Feichtenhofer_2019_ICCV}
\end{tabular}}}
& AT~\cite{madry2017towards} 
& \multirow{3}{*}{85.4}
& -11.3 & 68.2 
& \multirow{3}{*}{93.2}\\
& JPEG~\cite{125072} & & -5.2 & 75.5 & \\
& Denoising~\cite{claus2019videnn} & & -7.5 & 76.9 &\\
\hline
{\multirow{3}{*}{\begin{tabular}[c]{@{}c@{}}
TPN\\ \cite{TPN_2020_CVPR}
\end{tabular}}}
& AT~\cite{madry2017towards} 
& \multirow{3}{*}{74.3}
& -10.1 & 63.1 
& \multirow{3}{*}{92.0} \\
& JPEG~\cite{125072} & & -2.5 & 74.8 & \\
& Denoising~\cite{claus2019videnn} & & -4.0 & 75.3 & \\
\hline
{\multirow{3}{*}{\begin{tabular}[c]{@{}c@{}}
I3D\\ \cite{Carreira_2017_CVPR}
\end{tabular}}}
& AT~\cite{madry2017towards} 
& \multirow{3}{*}{71.7}
& -8.0 & 76.2 
& \multirow{3}{*}{92.1} \\
& JPEG~\cite{125072} & & -7.4 & 80.1 & \\
& Denoising~\cite{claus2019videnn} & & -5.8 & 81.8 & \\
\bottomrule
\end{tabular}}
\end{table}%

\begin{table}[ht!]
\centering
\caption{Untargeted Attack success rate (ASR) on video classification in the presence of various defense techniques. Results are gathered on the UCF-101 dataset.} \label{tab:untarget_defense_model_table}
\resizebox{0.85\columnwidth}{!}{
\begin{tabular}[t]{clcccc}
\toprule
\multirow{2}{*}{\begin{tabular}[c]{@{}c@{}}Video\\ Classifier\end{tabular}} & \multirow{2}{*}{Defense} & \multicolumn{2}{c}{\begin{tabular}[c]{@{}c@{}}ASR (\%)\\ w Defense\end{tabular}} & \multicolumn{2}{c}{\begin{tabular}[c]{@{}c@{}}ASR (\%)\\ w/o Defense\end{tabular}} \\ \cline{3-6}
& & Offline & Online & Offline & Online \\ 
\midrule
{\multirow{3}{*}{\begin{tabular}[c]{@{}c@{}}
SlowFast\\ \cite{Feichtenhofer_2019_ICCV}
\end{tabular}}}
& AT~\cite{madry2017towards} & 67.1 & 53.2
& \multirow{3}{*}{96.1} & \multirow{3}{*}{80.4}\\
& JPEG~\cite{125072} & 72.3 & 64.6 & \\
& Denoising~\cite{claus2019videnn} & 73.3 & 64.1 & \\\hline
{\multirow{3}{*}{\begin{tabular}[c]{@{}c@{}}
TPN \\ \cite{TPN_2020_CVPR}\end{tabular}}}
& AT~\cite{madry2017towards} & 64.2 & 58.2
& \multirow{3}{*}{95.8} & \multirow{3}{*}{81.3} \\
& JPEG~\cite{125072} & 70.9 & 61.2 & \\
& Denoising~\cite{claus2019videnn} & 71.8 & 63.8 & \\\hline
{\multirow{3}{*}{\begin{tabular}[c]{@{}c@{}}
I3D \\ \cite{Carreira_2017_CVPR}\end{tabular}}}
& AT~\cite{madry2017towards} & 75.8 & 65.3
& \multirow{3}{*}{96.3} & \multirow{3}{*}{80.7} \\
& JPEG~\cite{125072} & 80.8 & 72.2 &\\
& Denoising~\cite{claus2019videnn} & 82.7 & 68.5 & \\
\bottomrule
\end{tabular}}
\end{table}%

\noindent\textbf{Defense for Video Classification.} We now present the defense results against RoVISQ attacks for the video compression and classification system. Here, we benchmark our bandwidth attack. Table \ref{tab:target_defense_model_table} shows the classification accuracy of the original classification models along with the AT-trained models. We note that AT degrades the accuracy of the benign model significantly which hinders its applicability. We summarize the attack success rate (ASR) of our targeted and untargeted perturbations in the presence of AT defense in Table~\ref{tab:target_defense_model_table} and Table~\ref{tab:untarget_defense_model_table}, respectively. In both tables, we observe that our proposed attack still achieves a high attack success rate, despite the defense. This is due to the fundamentally complex problem of training a model that is universally robust to multiple adversarial attacks and can simultaneously classify clean video clips correctly. Based on the obtained results, we conclude that our adversarial attacks against video compression and classification are resilient against the AT defense mechanism. We include additional results for our attacks on the Jester dataset in Appendix Table~\ref{tab:jester_at_defense}.
\subsection{JPEG Compression}\label{sec:jpeg}
 JPEG~\cite{125072} is a popular lossy compression that applies discrete cosine transform (DCT) on images with different frequencies. We assume that the defender executes JPEG compression as a preprocessing before performing video compression. To compress a perturbed image, the high (low) frequency DCT coefficients are usually scaled more (less). The coefficients are then rounded to the nearest integers by performing a quantization controlled by compression factor (CF). CF directly controls the trade-off between image quality and compression rate, as such, we perform our evaluations with two different values of CF, i.e., $20$ and $40$.

\noindent\textbf{Defense for Video Compression.}
The results of JPEG compression against RoVISQ attacks are summarized in Table~\ref{tab:qoe_defense_model_JPEG_table}. As seen, our attacks can bypass the widely used JPEG image compression scheme, even when the compression factor CF is set to a high value of $40$. Note that when CF=$40$ in Table~\ref{tab:qoe_defense_model_JPEG_table}, we can observe that the compression performance of the clean video is reduced by $\sim$2dB even in the absence of the attack. This is because the texture of the input video which contains high-frequency information is also removed after JPEG coding. As shown, the effect of JPEG compression on attack effectiveness is very negligible, i.e., our attacks can still degrade the QoE factors as intended. This trend can be observed for both our online and offline attacks.

\begin{table}[t]
\centering
\caption{PSNR and Bpp of DVC on videos preprocessed by JPEG~\cite{125072} compression on the HEVC class C Dataset, where CF means compression factor.} \label{tab:qoe_defense_model_JPEG_table}
\resizebox{\columnwidth}{!}{
\begin{tabular}[t]{lccccc}
\toprule
\multirow{2}{*}{Benchmark} 
& \multirow{2}{*}{CF} 
& \multicolumn{2}{c}{w Defense}
& \multicolumn{2}{c}{w/o Defense} \\
& & PSNR (dB) & Bpp
& PSNR (dB) & Bpp \\
\midrule
{\multirow{2}{*}{DVC~\cite{Lu_2019_CVPR}}}
& 20 & 31.14 & 0.28 & \multirow{2}{*}{31.24} & \multirow{2}{*}{0.27} \\
& 40 & 29.26 & 0.21 & & \\\hline
{\multirow{2}{*}{Video Quality (Offline) }}
& 20 & -3.35 & +0.7$\%$ & \multirow{2}{*}{-3.52} & \multirow{2}{*}{+0.8$\%$}\\
& 40 & -3.14 & +0.6$\%$ \\\cline{2-6}
{\multirow{2}{*}{Video Quality (Online) }}
& 20 & -2.86 & +19.1$\%$ & \multirow{2}{*}{-3.05} & \multirow{2}{*}{+19.9$\%$}\\
& 40 & -2.76 & +18.4$\%$ & & \\\hline
{\multirow{2}{*}{Bandwidth (Offline) }}
& 20 & -0.25 & +95.4$\%$ & \multirow{2}{*}{-0.01} & \multirow{2}{*}{+99.5$\%$} \\
& 40 & -0.45 & +86.7$\%$ & &\\\cline{2-6}
{\multirow{2}{*}{Bandwidth (Online) }}
& 20 & -1.45 & +34.2$\%$ & \multirow{2}{*}{-0.39} & \multirow{2}{*}{+35.7$\%$}\\
& 40 & -1.76 & +31.2$\%$ & & \\\hline
{\multirow{2}{*}{RD (Offline) }}
& 20 & -4.09 & +82.6$\%$ & \multirow{2}{*}{-4.21} & \multirow{2}{*}{+85.3$\%$}\\
& 40 & -3.71 & +70.5$\%$ & &\\\cline{2-6}
{\multirow{2}{*}{RD (Online) }}
& 20 & -2.95 & +31.8$\%$ & \multirow{2}{*}{-3.10} & \multirow{2}{*}{+33.5$\%$}\\
& 40 & -2.79 & +28.6$\%$ & &\\
\bottomrule
\end{tabular}}
\end{table}%

\noindent\textbf{Defense for Video Classification.} To evaluate RoVISQ attacks against JPEG compression, we apply our offline and online attacks on UCF-101 datasets. As shown in Table~\ref{tab:target_defense_model_table}, applying JPEG compression reduces the accuracy of the downstream video classifier by 2.5-7.4$\%$ even in the absence of any adversarial attack. However, JPEG compression cannot remove the effect of our perturbations. Table~\ref{tab:target_defense_model_table} summarizes the attack success rate on video classification when JPEG is used as preprocessing before video compression. We see that JPEG fails to defend against our targeted attacks in most cases. In addition, Table~\ref{tab:untarget_defense_model_table} shows that compared to the AT defense against untargeted attacks, the performance of JPEG compression is relatively worse. The main reason is that JPEG compression removes adversarial instances on a frame-by-frame basis but it cannot suppress error propagation using a time-series analysis.
\subsection{Video Denoising}\label{sec:denoising}
Denoising is a fundamental video processing algorithm that removes sensor imperfections. We assume that the defender has a noise removal algorithm in the front-end sources, e.g., cameras, before applying the video compression. We use a state-of-the-art DNN-based blind denoiser (ViDeNN)~\cite{claus2019videnn} that does not assume any prior environmental conditions, e.g., color and light. As such, ViDeNN can adapt to the external changes where there is a different distribution for each frame. 


\noindent\textbf{Defense for Video Compression.} As seen in Table~\ref{tab:qoe_defense_model_denoising_table}, the addition of the video denoising pipeline at the front-end sources causes a $1.5$dB drop in the PSNR of the benign compression model even when the attack is not present. When our proposed video quality attack is applied in the offline and online settings, the PSNR is dropped by $3.23$dB and $2.76$dB compared to a $3.5$dB and $3.05$dB drop when the defense is not present, respectively. It can thus be observed that the video quality attack is largely resilient against denoising. Additionally, our bandwidth and RD attacks, particularly in the offline setting, still significantly affect the video compression after video denoising. It is worth noting that the high computational cost of ViDeNN makes it impractical for real-time defense on video content. We conclude that transforming the input signal through video denoising cannot prevent RoVISQ attacks. 
As shown in Figure~\ref{fig:Exp_qoe_pic}, even if the defender denoises the perturbed input, the error is still propagated from the perturbed input to the compressed video. 

\begin{table}[t]
\centering
\caption{PSNR and Bpp of DVC on videos preprocessed by DNN-based denoiser \cite{claus2019videnn} on the HEVC class C Dataset.} \label{tab:qoe_defense_model_denoising_table}
\resizebox{0.95\columnwidth}{!}{
\begin{tabular}[t]{lcccc}
\toprule
\multirow{2}{*}{Benchmark} 
& \multicolumn{2}{c}{w Defense}
& \multicolumn{2}{c}{w/o Defense} \\
& PSNR (dB) & Bpp
& PSNR (dB) & Bpp \\
\midrule
DVC~\cite{Lu_2019_CVPR}
& 29.74 & 0.28
& 31.24 & 0.27 \\\hline
\begin{tabular}[c]{@{}c@{}} Video Quality (Offline)\end{tabular} 
& -3.23 & +0.5$\%$ 
&-3.52 & +0.8$\%$\\
\begin{tabular}[c]{@{}c@{}} Video Quality (Online)\end{tabular}
& -2.76 & +14.3$\%$ 
& -3.05 & +19.9$\%$\\ \hline
\begin{tabular}[c]{@{}c@{}} Bandwidth (Offline)\end{tabular}
& -0.12 & +64.8$\%$
& -0.01 & +99.5$\%$\\
\begin{tabular}[c]{@{}c@{}} Bandwidth (Online)\end{tabular}
& -0.43 & +21.8$\%$ 
& -0.39 & +35.7$\%$\\ \hline
\begin{tabular}[c]{@{}c@{}} RD (Offline)\end{tabular}
& -3.81 & +56.8$\%$ 
& -4.21 & +85.3$\%$\\
\begin{tabular}[c]{@{}c@{}} RD (Online)\end{tabular}
& -2.63 & +18.4$\%$ 
& -3.10 & +33.5$\%$\\
\bottomrule
\end{tabular}}
\end{table}%

\noindent\textbf{Defense for Video Classification.} The denoising operation, by nature, is a lossy transformation that reduces the accuracy of the downstream classifier. As shown in Table~\ref{tab:target_defense_model_table}, denoising the benign streaming video can lead to 4-7.5$\%$ lower classification accuracy even when the input is not adversarial. Tables~\ref{tab:target_defense_model_table} and~\ref{tab:untarget_defense_model_table} report the attack success rate for targeted and untargeted scenarios when the perturbed video is passed through the denoiser. As can be observed, the targeted attack still maintains a high success rate of 75.3-81.8$\%$ which is 10.3-16.7$\%$ lower than when no defense is present. On the untargeted attack scenarios, the denoising reduces the attack success rate by 12.2-24$\%$ but it remains well above 60$\%$, thus still posing a critical problem for the classification service. Our results show that denoising the video cannot successfully remove the effect of injected adversarial perturbations.

\section{Related Work}
\label{sec:related_work}


\smallskip
\noindent\textbf{DNN-based Video Compression.}
In the past decades, plenty of handcrafted image and video compression standards were proposed, such as JPEG \cite{125072}, JPEG 2000 \cite{952804}, H.264 \cite{H.264}, and H.265 \cite{H.265}. Most of these methods follow handcrafted algorithms to remove redundancies in both spatial and temporal domains. Recently, DNN-based video compression frameworks have attracted a lot of attention~\cite{Lu_2019_CVPR,Hu_2021_CVPR,Yang_2020_CVPR}. Especially, the video compression framework in~\cite{Lu_2019_CVPR} achieves impressive results by replacing all the components in the standard H.264/H.265 video compression codecs with DNNs. Deep learning-based video compression techniques rely on convolutional neural networks (CNNs) for their three main design components: (1)~motion estimation network for estimating the temporal motion, (2)~motion compensation network to generate the predicted frame, and (3)~auto-encoder style network for compressing the motion and residual data. MPEG experts have already started discussions about a new video codec that takes advantage of DNN technology~\cite{Mpeg}. It has also been adopted in industry~\cite{Qualcomm}. An important motivation is DNN’s superior performance compared to legacy methods. Another motivation seen in industry is using DNN accelerators on mobile devices, e.g., Qualcomm Snapdragon 888 processor~\cite{Qualcomm}.

Compared to handcrafted video compression standards~\cite{H.264, H.265}, CNN-based end-to-end optimized video compression significantly reduces the redundancies in video motion~\cite{Lu_2019_CVPR}.
Furthermore, the \textit{R}-\textit{D} optimization adopted in DNN-based video compression enables higher compression efficiency by directly using the number of bits in the optimization procedure. To estimate the bit-rates, context models~\cite{Mentzer_2018_CVPR, Toderici_2017_CVPR, Ball_2017_ICLR, Ball_2018_ICLR, liu2020unified} are learned for the adaptive arithmetic coding method which compress discrete-valued data to bit-rates closely approaching the entropy of the representation. More recently, Yang \textit{et al.} \cite{Yang_2020_CVPR} shows that hierarchical GOP structure can make use of the advantageous information from high-quality frames by improving the temporal predictive coding. Furthermore, Hu \textit{et al.} \cite{Hu_2021_CVPR} propose to perform all operations (motion estimation, motion compensation, motion compression, and residual compression) in the feature space, which demonstrates better video compression performance.

\smallskip
\noindent\textbf{DNN-based Video Classification.}
DNNs are widely adopted in live video analysis services such as video activity recognition for health care \cite{healthcare}, video surveillance \cite{surveillance}, and self-driving cars\cite{autonomous}. One of the widely used approaches for video activity recognition is an inflated three-dimensional (I3D) network \cite{Carreira_2017_CVPR}. This method builds upon a pre-trained image classification model by inflating the convolutional and pooling kernels with an additional temporal dimension. 
Doing so enables the proposed spatiotemporal convolutions to treat spatial structures
and temporal events separately, creating a two-stream approach. 
Slowfast \cite{Feichtenhofer_2019_ICCV} also adopts a two-stream approach and improves the accuracy for action detection by digesting the input video at different temporal resolutions. More recently, Yang \textit{et al.} \cite{TPN_2020_CVPR} propose a generic feature-level Temporal Pyramid Network (TPN) to model speed variations amid different actions.

\smallskip
\noindent\textbf{Adversarial Attack and Defense on Image Domain.}
The adversarial attack was first proposed by \cite{Szegedy_2013} aiming to fool a victim model with small perturbations. Goodfellow \textit{et al.} \cite{Goodfellow_2014} developed the fast gradient sign method (FGSM) to calculate the perturbation by following the direction of gradients. Kurakin \textit{et al.} \cite{kurakin2016adversarial} extended FGSM to an iterative approach, called iterative FGSM (I-FGSM), and showed higher attack success rates. Recently, adversarial attacks using image compression \cite{duan2021advdrop} have been proposed that drop critical information from images. This method optimizes over a trainable quantization table by minimizing the adversarial loss. 

In response to adversarial attacks, many prior works suggest adversarial training to improve the robustness of the victim DNN~\cite{tramer2017ensemble, escalante2009particle, shafahi2019adversarial, madry2017towards}. Nevertheless, adversarial training is known to negatively affect the accuracy on clean samples, thus leading to the much-debated trade-off between accuracy and robustness. Recently, compression-based defense algorithms~\cite{Jia_2019_CVPR, aydemir2018effects,8953640,8416586} have been proposed which use lossy compression techniques such as JPEG coding to remove the adversarial perturbations. In the video domain, research on defenses against adversarial attacks using video compression is not explored in prior methods.

\smallskip
\noindent\textbf{Adversarial Attack and Defense on Video Domain.}
There are only a few studies that propose adversarial attack on video action recognition models. Wei \textit{et al.} \cite{Wei_aaai_2019} propose an optimization-based method to generate adversarial perturbation for the CNN+RNN video classifier \cite{Donahue_2015_CVPR}.  Pony \textit{et al.} \cite{Pony_2021_CVPR} present a flickering attack that changes the color of each frame to obtain an adversarial effect and fool the video recognition model. Li \textit{et al.} \cite{Li_NIPS_2021} employ standard geometric transformations for query-efficient black-box attacks. The attacks proposed in~\cite{Li_NIPS_2021, Wei_aaai_2019,Pony_2021_CVPR, cao2022stylefool} are all performed offline. In this context, the video is decomposed into frames which then undergo adversarial attacks, similar to the image domain. 

Li \textit{et al.} \cite{Li_2019_NDSS} train a universal perturbation generator offline and use it to attack real-time video classification systems. The most recent study \cite{Xie_SP_2022} generates universal 3-dimensional perturbations to subvert real-time video classification systems using a surrogate DNN model. The proposed perturbations in the above studies \cite{Wei_aaai_2019, Pony_2021_CVPR, Xie_SP_2022, Li_2019_NDSS,Li_NIPS_2021} lose effect once the video is passed through the compression pipeline. As such, they cannot be applied in real-world live video streaming and classification scenarios where video compression is a crucial step, as shown in Figure~\ref{fig:intro}. The proposed attacks on video classification in this work, however, 
can withstand several lossy signal transformations, e.g., compression and denoising.

There are very few defenses that protect against adversarial attacks on videos~\cite{ lo2020defending}. Lo \textit{et al.}~\cite{lo2020defending} perform adversarial training with multiple independent batch normalization layers to learn different perturbation types. Denoising~\cite{claus2019videnn} is also a fundamental video processing method to remove noise from a camera sensor, such as mobile phones and surveillance cameras. In this paper, we show that RoVISQ attacks are robust to the proposed adversarial training and DNN-based denoiser. 



\section{\edit{Conclusion, Limitations, and Future Work}}
This paper presents the first systematic study on adversarial attacks to deep learning-based video compression systems. Our attacks, dubbed RoVISQ, are the first to manipulate the Rate-Distortion ($R$-$D$) relationship of the video compression model for influencing the user quality of experience (QoE). RoVISQ attacks are formalized as well-defined optimization problems, thus resulting in a dramatic degradation of video quality and compression ratio. We also propose novel targeted and untargeted attacks against a downstream deep learning-based video classification system. While the video compression framework inherently invalidates most adversarial examples, our attack is robust to compression by considering the temporal coding structure of the GOP and encoding parameters. Our comprehensive experiments show that our attacks outperform noise baselines and previously proposed attacks in both offline and online settings. Furthermore, our attacks still maintain high success rate in the presence of various defenses, such as adversarial training, video denoising, and JPEG coding.

\edit{While RoVISQ shows great success in attacking various video compression systems, the perturbations designed in this work are purely digital. An interesting future work is investigating physical sensor-based attacks where the video is perturbed during generation, e.g., using LED light manipulations at the camera. Another area of future studies is expanding factors that affect the user QoE in video compression and classification systems, potentially by performing large-scale user studies. Establishing effective defense strategies against our attacks is also a crucial and promising direction.}

\section*{Acknowledgement}
We would like to thank the anonymous NDSS reviewers for their valuable feedback. We also thank Mohammad Samragh for fruitful discussions. This work was supported by the U.S. Army/Department of Defense award number W911NF2020267.

\bibliographystyle{plain}
\bibliography{main}

\begin{thebibliography}{10}

\bibitem{Mpeg}
Explorations: Neural network-based video compression.
  https://www.mpeg.org/standards/{Explorations}/36/.

\bibitem{Qualcomm}
How ai research is enabling next-gen codecs.
  https://www.qualcomm.com/news/onq/2021/07/14/how-ai-research-enabling-next-gen-codecs.

\bibitem{Twitch}
Twitch official website. https://www.twitch.tv.

\bibitem{YouTube}
Youtube live streaming. https://www.youtube.com/live.

\bibitem{avgousti2018medical}
Sotiris Avgousti, Andreas~S Panayides, Eftychios~G Christoforou, Argyris
  Argyrou, Antonis Jossif, Panicos Masouras, Cyril Novales, and Pierre Vieyres.
\newblock Medical telerobotics and the remote ultrasonography paradigm over 4g
  wireless networks.
\newblock In {\em 2018 IEEE 20th International Conference on e-Health
  Networking, Applications and Services (Healthcom)}, pages 1--6. IEEE, 2018.

\bibitem{aydemir2018effects}
Ayse~Elvan Aydemir, Alptekin Temizel, and Tugba~Taskaya Temizel.
\newblock The effects of jpeg and jpeg2000 compression on attacks using
  adversarial examples.
\newblock {\em CoRR}, abs/1803.10418, 2018.

\bibitem{Ball_2018_ICLR}
Johannes {Ball{\'e}}, David {Minnen}, Saurabh {Singh}, Sung~Jin {Hwang}, and
  Nic {Johnston}.
\newblock Variational image compression with a scale hyperprior.
\newblock In {\em International Conference on Learning Representations (ICLR)},
  2018.

\bibitem{Ball_2017_ICLR}
Johannes Ballé, Valero Laparra, and Eero~P. Simoncelli.
\newblock End-to-end optimized image compression.
\newblock In {\em International Conference on Learning Representations (ICLR)},
  2017.

\bibitem{cao2022stylefool}
Yuxin Cao, Xi~Xiao, Ruoxi Sun, Derui Wang, Minhui Xue, and Sheng Wen.
\newblock Stylefool: Fooling video classification systems via style transfer.
\newblock {\em arXiv preprint arXiv:2203.16000}, 2022.

\bibitem{carlini2017adversarial}
Nicholas Carlini and David Wagner.
\newblock Adversarial examples are not easily detected: Bypassing ten detection
  methods.
\newblock In {\em Proceedings of the 10th ACM workshop on artificial
  intelligence and security}, pages 3--14, 2017.

\bibitem{carlini2017towards}
Nicholas Carlini and David Wagner.
\newblock Towards evaluating the robustness of neural networks.
\newblock In {\em 2017 ieee symposium on security and privacy (sp)}, pages
  39--57. Ieee, 2017.

\bibitem{8424625}
Nicholas Carlini and David Wagner.
\newblock Audio adversarial examples: Targeted attacks on speech-to-text.
\newblock In {\em 2018 IEEE Security and Privacy Workshops (SPW)}, pages 1--7,
  2018.

\bibitem{Carreira_2017_CVPR}
Joao Carreira and Andrew Zisserman.
\newblock Quo vadis, action recognition? a new model and the kinetics dataset.
\newblock In {\em Proceedings of the IEEE Conference on Computer Vision and
  Pattern Recognition (CVPR)}, July 2017.

\bibitem{chen2020hopskipjumpattack}
Jianbo Chen, Michael~I Jordan, and Martin~J Wainwright.
\newblock Hopskipjumpattack: A query-efficient decision-based attack.
\newblock In {\em 2020 ieee symposium on security and privacy (sp)}, pages
  1277--1294. IEEE, 2020.

\bibitem{AV1}
Yue Chen, Debargha Murherjee, Jingning Han, Adrian Grange, Yaowu Xu, Zoe Liu,
  Sarah Parker, Cheng Chen, Hui Su, Urvang Joshi, Ching-Han Chiang, Yunqing
  Wang, Paul Wilkins, Jim Bankoski, Luc Trudeau, Nathan Egge, Jean-Marc Valin,
  Thomas Davies, Steinar Midtskogen, Andrey Norkin, and Peter de~Rivaz.
\newblock An overview of core coding tools in the av1 video codec.
\newblock In {\em 2018 Picture Coding Symposium (PCS)}, pages 41--45, 2018.

\bibitem{claus2019videnn}
Michele Claus and Jan van Gemert.
\newblock Videnn: Deep blind video denoising.
\newblock In {\em Proceedings of the IEEE/CVF Conference on Computer Vision and
  Pattern Recognition Workshops}, pages 0--0, 2019.

\bibitem{concolato2017adaptive}
Cyril Concolato, Jean Le~Feuvre, Franck Denoual, Fr{\'e}d{\'e}ric Maz{\'e},
  Eric Nassor, Nael Ouedraogo, and Jonathan Taquet.
\newblock Adaptive streaming of hevc tiled videos using mpeg-dash.
\newblock {\em IEEE transactions on circuits and systems for video technology},
  28(8):1981--1992, 2017.

\bibitem{Donahue_2015_CVPR}
Jeffrey Donahue, Lisa Anne~Hendricks, Sergio Guadarrama, Marcus Rohrbach,
  Subhashini Venugopalan, Kate Saenko, and Trevor Darrell.
\newblock Long-term recurrent convolutional networks for visual recognition and
  description.
\newblock In {\em Proceedings of the IEEE Conference on Computer Vision and
  Pattern Recognition (CVPR)}, June 2015.

\bibitem{duan2021advdrop}
Ranjie Duan, Yuefeng Chen, Dantong Niu, Yun Yang, A~Kai Qin, and Yuan He.
\newblock Advdrop: Adversarial attack to dnns by dropping information.
\newblock In {\em Proceedings of the IEEE/CVF International Conference on
  Computer Vision}, pages 7506--7515, 2021.

\bibitem{escalante2009particle}
Hugo~Jair Escalante, Manuel Montes, and Luis~Enrique Sucar.
\newblock Particle swarm model selection.
\newblock {\em Journal of Machine Learning Research}, 10(2), 2009.

\bibitem{Feichtenhofer_2019_ICCV}
Christoph Feichtenhofer, Haoqi Fan, Jitendra Malik, and Kaiming He.
\newblock Slowfast networks for video recognition.
\newblock In {\em Proceedings of the IEEE/CVF International Conference on
  Computer Vision (ICCV)}, October 2019.

\bibitem{Goodfellow_2014}
Ian~J Goodfellow, Jonathon Shlens, and Christian Szegedy.
\newblock Explaining and harnessing adversarial examples.
\newblock {\em arXiv preprint arXiv:1412.6572}, 2014.

\bibitem{hassan2019high}
Ali Hassan, Mubeen Ghafoor, Syed~Ali Tariq, Tehseen Zia, and Waqas Ahmad.
\newblock High efficiency video coding (hevc)--based surgical telementoring
  system using shallow convolutional neural network.
\newblock {\em Journal of digital imaging}, 32(6):1027--1043, 2019.

\bibitem{healthcare}
M.~Shamim Hossain.
\newblock Patient state recognition system for healthcare using speech and
  facial expressions.
\newblock {\em Journal of medical systems 40}, pages 1--8, 2016.

\bibitem{pytorchvc}
Zhihao Hu.
\newblock Pytorch video compression,
  https://github.com/zhihaohu/pytorchvideocompression.
\newblock 2020.

\bibitem{Hu_2021_CVPR}
Zhihao Hu, Guo Lu, and Dong Xu.
\newblock Fvc: A new framework towards deep video compression in feature space.
\newblock In {\em Proceedings of the IEEE/CVF Conference on Computer Vision and
  Pattern Recognition (CVPR)}, pages 1502--1511, June 2021.

\bibitem{hussain2021waveguard}
Shehzeen Hussain, Paarth Neekhara, Shlomo Dubnov, Julian McAuley, and Farinaz
  Koushanfar.
\newblock Waveguard: Understanding and mitigating audio adversarial examples.
\newblock In {\em 30th USENIX Security Symposium (USENIX Security 21)}, 2021.

\bibitem{huszak2010analysing}
{\'A}rp{\'a}d Husz{\'a}k and S{\'a}ndor Imre.
\newblock Analysing gop structure and packet loss effects on error propagation
  in mpeg-4 video streams.
\newblock In {\em 2010 4th International Symposium on Communications, Control
  and Signal Processing (ISCCSP)}, pages 1--5. IEEE, 2010.

\bibitem{Jia_2019_CVPR}
Xiaojun Jia, Xingxing Wei, Xiaochun Cao, and Hassan Foroosh.
\newblock Comdefend: An efficient image compression model to defend adversarial
  examples.
\newblock In {\em Proceedings of the IEEE/CVF Conference on Computer Vision and
  Pattern Recognition (CVPR)}, June 2019.

\bibitem{kambar2022survey}
Mina Esmail Zadeh~Nojoo Kambar, Armin Esmaeilzadeh, Yoohwan Kim, and Kazem
  Taghva.
\newblock A survey on mobile malware detection methods using machine learning.
\newblock In {\em 2022 IEEE 12th Annual Computing and Communication Workshop
  and Conference (CCWC)}, pages 0215--0221. IEEE, 2022.

\bibitem{autonomous}
Hirokatsu Kataoka, Yutaka Satoh, Yoshimitsu Aoki, Shoko Oikawa, and Yasuhiro
  Matsui.
\newblock Temporal and fine-grained pedestrian action recognition on driving
  recorder database.
\newblock {\em Sensors}, 18(2):627, 2018.

\bibitem{khalid2019qusecnets}
Faiq Khalid, Hassan Ali, Hammad Tariq, Muhammad~Abdullah Hanif, Semeen Rehman,
  Rehan Ahmed, and Muhammad Shafique.
\newblock Qusecnets: Quantization-based defense mechanism for securing deep
  neural network against adversarial attacks.
\newblock In {\em 2019 IEEE 25th International Symposium on On-Line Testing and
  Robust System Design (IOLTS)}, pages 182--187. IEEE, 2019.

\bibitem{Kim_sigcomm_20}
Jaehong Kim, Youngmok Jung, Hyunho Yeo, Juncheol Ye, and Dongsu Han.
\newblock Neural-enhanced live streaming: Improving live video ingest via
  online learning.
\newblock In {\em Proceedings of the Annual Conference of the ACM Special
  Interest Group on Data Communication on the Applications, Technologies,
  Architectures, and Protocols for Computer Communication}, SIGCOMM '20, page
  107–125, New York, NY, USA, 2020. Association for Computing Machinery.

\bibitem{kohler2022signal}
Sebastian K{\"o}hler, Richard Baker, and Ivan Martinovic.
\newblock Signal injection attacks against ccd image sensors.
\newblock In {\em Proceedings of the 2022 ACM on Asia Conference on Computer
  and Communications Security}, pages 294--308, 2022.

\bibitem{kurakin2016adversarial}
Alexey Kurakin, Ian Goodfellow, Samy Bengio, et~al.
\newblock Adversarial examples in the physical world, 2016.

\bibitem{Li_NIPS_2021}
Shasha Li, Abhishek Aich, Shitong Zhu, Salman Asif, Chengyu Song, Amit
  Roy-Chowdhury, and Srikanth Krishnamurthy.
\newblock Adversarial attacks on black box video classifiers: Leveraging the
  power of geometric transformations.
\newblock {\em Advances in Neural Information Processing Systems}, 34, 2021.

\bibitem{Li_2019_NDSS}
Shasha Li, Ajaya Neupane, Sujoy Paul, Chengyu Song, Srikanth~V. Krishnamurthy,
  Amit K.~Roy Chowdhury, and Ananthram Swami.
\newblock Stealthy adversarial perturbations against real-time video
  classification systems.
\newblock In {\em Proceedings 2019 Network and Distributed System Security
  Symposium}, 2019.

\bibitem{Li_audio_ccs}
Zhuohang Li, Yi~Wu, Jian Liu, Yingying Chen, and Bo~Yuan.
\newblock {\em AdvPulse: Universal, Synchronization-Free, and Targeted Audio
  Adversarial Attacks via Subsecond Perturbations}, page 1121–1134.
\newblock ACM, 2020.

\bibitem{likert1932technique}
Rensis Likert.
\newblock A technique for the measurement of attitudes.
\newblock {\em Archives of psychology}, 1932.

\bibitem{lin2019defensive}
Ji~Lin, Chuang Gan, and Song Han.
\newblock Defensive quantization: When efficiency meets robustness.
\newblock {\em arXiv preprint arXiv:1904.08444}, 2019.

\bibitem{liu2020unified}
Jiaheng Liu, Guo Lu, Zhihao Hu, and Dong Xu.
\newblock A unified end-to-end framework for efficient deep image compression.
\newblock {\em CoRR}, abs/2002.03370, 2020.

\bibitem{8953640}
Zihao Liu, Qi~Liu, Tao Liu, Nuo Xu, Xue Lin, Yanzhi Wang, and Wujie Wen.
\newblock Feature distillation: Dnn-oriented jpeg compression against
  adversarial examples.
\newblock In {\em 2019 IEEE/CVF Conference on Computer Vision and Pattern
  Recognition (CVPR)}, pages 860--868, 2019.

\bibitem{lo2020defending}
Shao-Yuan Lo and Vishal~M Patel.
\newblock Defending against multiple and unforeseen adversarial videos.
\newblock {\em arXiv preprint arXiv:2009.05244}, 2020.

\bibitem{Lu_2019_CVPR}
Guo Lu, Wanli Ouyang, Dong Xu, Xiaoyun Zhang, Chunlei Cai, and Zhiyong Gao.
\newblock Dvc: An end-to-end deep video compression framework.
\newblock In {\em Proceedings of the IEEE/CVF Conference on Computer Vision and
  Pattern Recognition (CVPR)}, June 2019.

\bibitem{lu2020edge}
Sidi Lu, Xin Yuan, and Weisong Shi.
\newblock Edge compression: An integrated framework for compressive imaging
  processing on cavs.
\newblock In {\em 2020 IEEE/ACM Symposium on Edge Computing (SEC)}, pages
  125--138. IEEE, 2020.

\bibitem{madry2017towards}
Aleksander Madry, Aleksandar Makelov, Ludwig Schmidt, Dimitris Tsipras, and
  Adrian Vladu.
\newblock Towards deep learning models resistant to adversarial attacks.
\newblock {\em arXiv preprint arXiv:1706.06083}, 2017.

\bibitem{materzynska2019jester}
Joanna Materzynska, Guillaume Berger, Ingo Bax, and Roland Memisevic.
\newblock The jester dataset: A large-scale video dataset of human gestures.
\newblock In {\em Proceedings of the IEEE/CVF International Conference on
  Computer Vision Workshops}, 2019.

\bibitem{Mentzer_2018_CVPR}
Fabian Mentzer, Eirikur Agustsson, Michael Tschannen, Radu Timofte, and Luc
  Van~Gool.
\newblock Conditional probability models for deep image compression.
\newblock In {\em Proceedings of the IEEE Conference on Computer Vision and
  Pattern Recognition (CVPR)}, June 2018.

\bibitem{ortega1998rate}
Antonio Ortega and Kannan Ramchandran.
\newblock Rate-distortion methods for image and video compression.
\newblock {\em IEEE Signal processing magazine}, 15(6):23--50, 1998.

\bibitem{ou2014q}
Yen-Fu Ou, Yuanyi Xue, and Yao Wang.
\newblock Q-star: A perceptual video quality model considering impact of
  spatial, temporal, and amplitude resolutions.
\newblock {\em IEEE Transactions on Image Processing}, 23(6):2473--2486, 2014.

\bibitem{pingle2018real}
Bhargav Pingle, Aakif Mairaj, and Ahmad~Y Javaid.
\newblock Real-world man-in-the-middle (mitm) attack implementation using open
  source tools for instructional use.
\newblock In {\em 2018 IEEE International Conference on Electro/Information
  Technology (EIT)}, pages 0192--0197. IEEE, 2018.

\bibitem{poeplau2014execute}
Sebastian Poeplau, Yanick Fratantonio, Antonio Bianchi, Christopher Kruegel,
  and Giovanni Vigna.
\newblock Execute this! analyzing unsafe and malicious dynamic code loading in
  android applications.
\newblock In {\em NDSS}, volume~14, pages 23--26, 2014.

\bibitem{Pony_2021_CVPR}
Roi Pony, Itay Naeh, and Shie Mannor.
\newblock Over-the-air adversarial flickering attacks against video recognition
  networks.
\newblock In {\em Proceedings of the IEEE/CVF Conference on Computer Vision and
  Pattern Recognition (CVPR)}, pages 515--524, June 2021.

\bibitem{8416586}
Aaditya Prakash, Nick Moran, Solomon Garber, Antonella DiLillo, and James
  Storer.
\newblock Protecting jpeg images against adversarial attacks.
\newblock In {\em 2018 Data Compression Conference}, pages 137--146, 2018.

\bibitem{samelakjoint}
Jaros{\l}aw Samelak, Jakub Stankowski, and Marek Doma{\'n}ski.
\newblock Joint collaborative team on video coding (jct-vc) of itu-t sg 16 wp 3
  and iso/iec jtc 1/sc 29/wg 11 26th meeting: Geneva, ch, 12--20 january 2017.

\bibitem{shafahi2019adversarial}
Ali Shafahi, Mahyar Najibi, Amin Ghiasi, Zheng Xu, John Dickerson, Christoph
  Studer, Larry~S Davis, Gavin Taylor, and Tom Goldstein.
\newblock Adversarial training for free!
\newblock {\em arXiv preprint arXiv:1904.12843}, 2019.

\bibitem{Sharif_2020_CCS}
Mahmood Sharif, Sruti Bhagavatula, Lujo Bauer, and Michael~K Reiter.
\newblock Accessorize to a crime: Real and stealthy attacks on state-of-the-art
  face recognition.
\newblock In {\em Proceedings of the 2016 acm sigsac conference on computer and
  communications security}, pages 1528--1540, 2016.

\bibitem{952804}
A.~Skodras, C.~Christopoulos, and T.~Ebrahimi.
\newblock The jpeg 2000 still image compression standard.
\newblock {\em IEEE Signal Processing Magazine}, 18(5):36--58, 2001.

\bibitem{soomro2012ucf101}
Khurram Soomro, Amir~Roshan Zamir, and Mubarak Shah.
\newblock Ucf101: A dataset of 101 human actions classes from videos in the
  wild.
\newblock {\em arXiv preprint arXiv:1212.0402}, 2012.

\bibitem{H.265}
Gary~J Sullivan, Jens-Rainer Ohm, Woo-Jin Han, and Thomas Wiegand.
\newblock Overview of the high efficiency video coding (hevc) standard.
\newblock {\em IEEE Transactions on circuits and systems for video technology},
  22(12):1649--1668, 2012.

\bibitem{sullivan1998rate}
Gary~J Sullivan and Thomas Wiegand.
\newblock Rate-distortion optimization for video compression.
\newblock {\em IEEE signal processing magazine}, 15(6):74--90, 1998.

\bibitem{surveillance}
Waqas Sultani, Chen Chen, and Mubarak Shah.
\newblock Real-world anomaly detection in surveillance videos.
\newblock In {\em Proceedings of the IEEE conference on computer vision and
  pattern recognition}, pages 6479--6488, 2018.

\bibitem{Cisco_vc}
Cisco Systems.
\newblock Cisco visual networking index: Forecast and methodology.
\newblock 2010.

\bibitem{Szegedy_2013}
Christian Szegedy, Wojciech Zaremba, Ilya Sutskever, Joan Bruna, Dumitru Erhan,
  Ian Goodfellow, and Rob Fergus.
\newblock Intriguing properties of neural networks.
\newblock {\em arXiv preprint arXiv:1312.6199}, 2013.

\bibitem{Toderici_2017_CVPR}
George Toderici, Damien Vincent, Nick Johnston, Sung Jin~Hwang, David Minnen,
  Joel Shor, and Michele Covell.
\newblock Full resolution image compression with recurrent neural networks.
\newblock In {\em Proceedings of the IEEE Conference on Computer Vision and
  Pattern Recognition (CVPR)}, July 2017.

\bibitem{tramer2017ensemble}
Florian Tram{\`e}r, Alexey Kurakin, Nicolas Papernot, Ian Goodfellow, Dan
  Boneh, and Patrick McDaniel.
\newblock Ensemble adversarial training: Attacks and defenses.
\newblock {\em arXiv preprint arXiv:1705.07204}, 2017.

\bibitem{125072}
G.K. Wallace.
\newblock The jpeg still picture compression standard.
\newblock {\em IEEE Transactions on Consumer Electronics}, 38(1):xviii--xxxiv,
  1992.

\bibitem{wang2017cruise}
Xu~Wang, Jing Xiao, Ruimin Hu, and Zhongyuan Wang.
\newblock Cruise uav video compression based on long-term wide-range
  background.
\newblock In {\em 2017 Data Compression Conference (DCC)}, pages 466--466. IEEE
  Computer Society, 2017.

\bibitem{Wei_aaai_2019}
Xingxing Wei, Jun Zhu, Sha Yuan, and Hang Su.
\newblock Sparse adversarial perturbations for videos.
\newblock In {\em Proceedings of the AAAI Conference on Artificial
  Intelligence}, volume~33, pages 8973--8980, 2019.

\bibitem{H.264}
Thomas Wiegand, Gary~J Sullivan, Gisle Bjontegaard, and Ajay Luthra.
\newblock Overview of the h. 264/avc video coding standard.
\newblock {\em IEEE Transactions on circuits and systems for video technology},
  13(7):560--576, 2003.

\bibitem{Xie_SP_2022}
Shangyu Xie, Han Wang, Yu~Kong, and Yuan Hong.
\newblock Universal 3-dimensional perturbations for black-box attacks on video
  recognition systems.
\newblock In {\em 2022 IEEE Symposium on Security and Privacy (SP)}, 2022.

\bibitem{xue2019video}
Tianfan Xue, Baian Chen, Jiajun Wu, Donglai Wei, and William~T Freeman.
\newblock Video enhancement with task-oriented flow.
\newblock {\em International Journal of Computer Vision}, 127(8):1106--1125,
  2019.

\bibitem{TPN_2020_CVPR}
Ceyuan Yang, Yinghao Xu, Jianping Shi, Bo~Dai, and Bolei Zhou.
\newblock Temporal pyramid network for action recognition.
\newblock In {\em Proceedings of the IEEE/CVF Conference on Computer Vision and
  Pattern Recognition (CVPR)}, June 2020.

\bibitem{Yang_2020_CVPR}
Ren Yang, Fabian Mentzer, Luc~Van Gool, and Radu Timofte.
\newblock Learning for video compression with hierarchical quality and
  recurrent enhancement.
\newblock In {\em Proceedings of the IEEE/CVF Conference on Computer Vision and
  Pattern Recognition (CVPR)}, June 2020.

\bibitem{ye2019omnidirectional}
Yan Ye, Jill~M Boyce, and Philippe Hanhart.
\newblock Omnidirectional 360° video coding technology in responses to the
  joint call for proposals on video compression with capability beyond hevc.
\newblock {\em IEEE Transactions on Circuits and Systems for Video Technology},
  30(5):1241--1252, 2019.

\bibitem{zhou2020rate}
Mingliang Zhou, Xuekai Wei, Sam Kwong, Weijia Jia, and Bin Fang.
\newblock Rate control method based on deep reinforcement learning for dynamic
  video sequences in hevc.
\newblock {\em IEEE Transactions on Multimedia}, 23:1106--1121, 2020.

\end{thebibliography}

\begin{appendices}

\begin{figure}[h]
\centering
\includegraphics[width=\columnwidth]{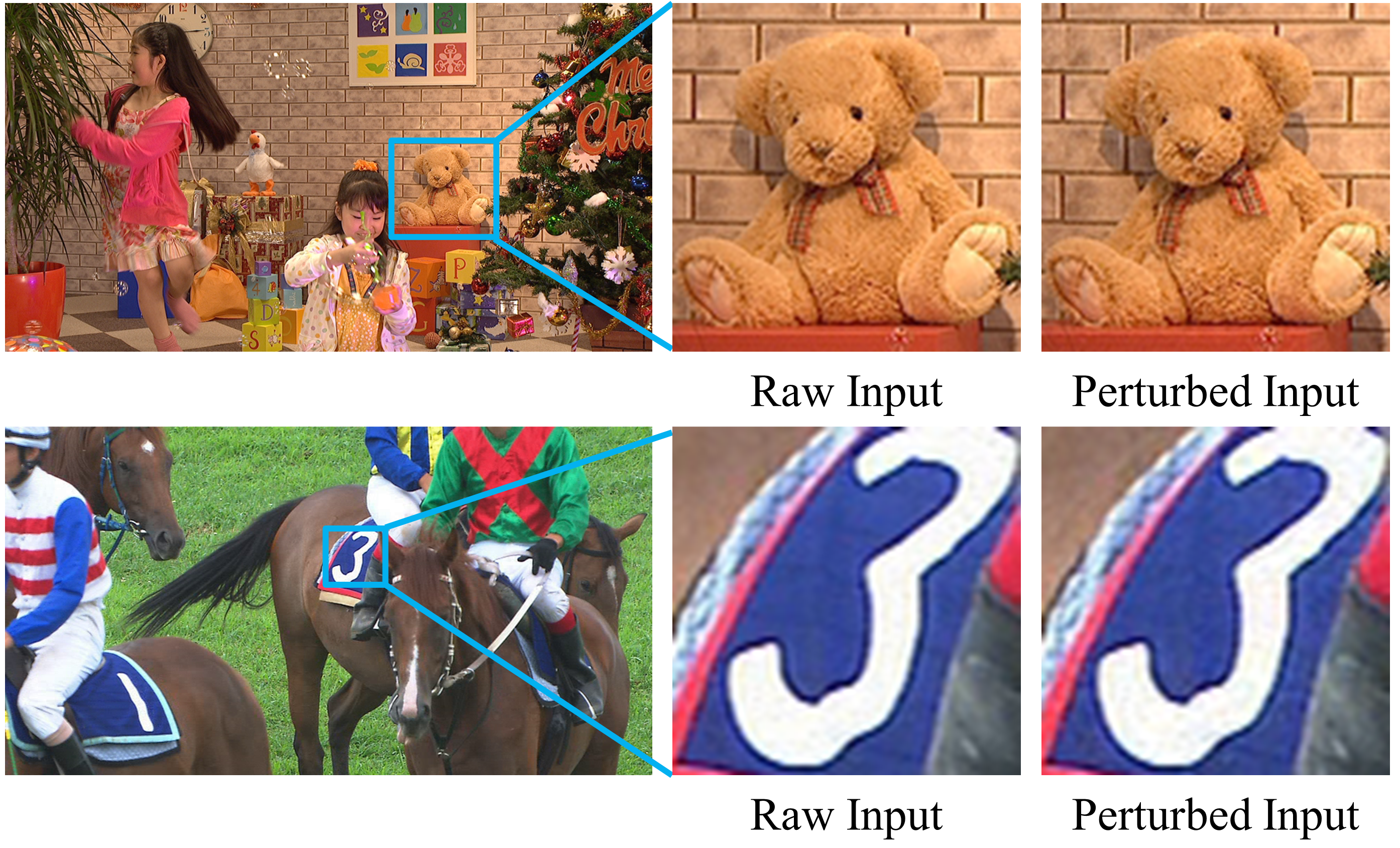}
\caption{\edit{Visual comparison between the original (raw) and adversarial input videos when \rdo is used to craft the perturbations. As seen, RoVISQ perturbations for the input video have extremely small noise magnitude that remains stealthy to human eye.
}}
\label{fig:adv_input_video}
\vspace{-0.5cm}
\end{figure}

\begin{table*}[t]
\begin{subtable}[c]{0.32\textwidth}
\centering
\begin{tabular}[t]{cccc}
\toprule
Attacks& Dataset & PSNR (dB) & Bpp\\
\midrule
{\multirow{3}{*}{\begin{tabular}[c]{@{}c@{}} Video Quality \\(Offline)\end{tabular}}}
& Class B & -3.85 & +0.9$\%$\\ 
& Class C & -3.52 & +0.7$\%$ \\
& Class D & -3.94 &  +0.7$\%$ \\ \hline
{\multirow{3}{*}{\begin{tabular}[c]{@{}c@{}} Video Quality \\(Online)\end{tabular}}}
& Class B & -2.30 & +13.6$\%$\\ 
& Class C & -3.05 & +19.9$\%$ \\
& Class D & -2.82 &  +20.3$\%$ \\ \hline
{\multirow{3}{*}{\begin{tabular}[c]{@{}c@{}} Bandwidth \\(Offline)\end{tabular}}}
& Class B & -0.00 & +87.5$\%$\\ 
& Class C & -0.01 & +99.4$\%$ \\
& Class D & -0.00 &  +87.5$\%$ \\ \hline
{\multirow{3}{*}{\begin{tabular}[c]{@{}c@{}} Bandwidth \\(Online)\end{tabular}}}
& Class B & -0.50 & +35.3$\%$\\ 
& Class C & -0.39 & +35.7$\%$ \\
& Class D & -0.18 &  +28.7$\%$ \\ \hline
{\multirow{3}{*}{\begin{tabular}[c]{@{}c@{}} RD \\(Offline)\end{tabular}}}
& Class B & -4.03 & +89.1$\%$\\ 
& Class C & -4.21 & +85.3$\%$ \\
& Class D & -4.38 &  +81.2$\%$ \\ \hline
{\multirow{3}{*}{\begin{tabular}[c]{@{}c@{}} RD \\(Online)\end{tabular}}}
& Class B & -2.38 & +30.3$\%$\\ 
& Class C & -3.10 & +33.5$\%$ \\
& Class D & -3.41 &  +25.3$\%$ \\ \hline
{\multirow{3}{*}{Gaussian (Case I)}}
& Class B & -1.22 & +12.7$\%$\\
& Class C & -1.53 & +16.9$\%$ \\
& Class D & -1.66 & +17.7$\%$ \\ \hline
{\multirow{3}{*}{Gaussian (Case II)}}
& Class B & -2.17 & +29.5$\%$\\
& Class C & -2.25 & +31.2$\%$ \\
& Class D & -2.39 & +32.6$\%$ \\ 
\bottomrule
\end{tabular}
\subcaption{\label{tab:qoe_table_dvc}}
\end{subtable}
\hfill
\begin{subtable}[c]{0.32\textwidth}
\centering
\begin{tabular}[t]{cccc}
\toprule
Attacks& Dataset & PSNR (dB) & Bpp\\
\midrule
{\multirow{3}{*}{\begin{tabular}[c]{@{}c@{}} Video Quality \\(Offline)\end{tabular}}}
& Class B & -3.94 & +0.8$\%$\\ 
& Class C & -4.02 & +0.7$\%$ \\
& Class D & -4.11 & +0.8$\%$ \\ \hline
{\multirow{3}{*}{\begin{tabular}[c]{@{}c@{}} Video Quality \\(Online)\end{tabular}}}
& Class B & -2.41 & +14.2$\%$\\ 
& Class C & -3.11 & +19.9$\%$ \\
& Class D & -2.97 & +20.3$\%$ \\ \hline
{\multirow{3}{*}{\begin{tabular}[c]{@{}c@{}} Bandwidth \\(Offline)\end{tabular}}}
& Class B & -0.00 & +112.3$\%$\\ 
& Class C & -0.01 & +102.6$\%$ \\
& Class D & -0.01 &  +94.2$\%$ \\ \hline
{\multirow{3}{*}{\begin{tabular}[c]{@{}c@{}} Bandwidth \\(Online)\end{tabular}}}
& Class B & -0.62 & +63.3$\%$\\ 
& Class C & -0.51 & +47.2$\%$ \\
& Class D & -0.22 & +39.6$\%$ \\ \hline
{\multirow{3}{*}{\begin{tabular}[c]{@{}c@{}} RD \\(Offline)\end{tabular}}}
& Class B & -4.15 & +107.6$\%$\\ 
& Class C & -4.19 & +96.2$\%$ \\
& Class D & -4.23 & +91.7$\%$ \\ \hline
{\multirow{3}{*}{\begin{tabular}[c]{@{}c@{}} RD \\(Online)\end{tabular}}}
& Class B & -2.67 & +51.5$\%$\\ 
& Class C & -3.76 & +42.8$\%$ \\
& Class D & -3.88 & +39.8$\%$ \\ \hline
{\multirow{3}{*}{Gaussian (Case I)}}
& Class B & -1.19 & +7.7$\%$\\
& Class C & -1.69 & +15.6$\%$ \\
& Class D & -1.66 & +17.7$\%$ \\ \hline
{\multirow{3}{*}{Gaussian (Case II)}}
& Class B & -1.80 & +31.3$\%$\\
& Class C & -2.37 & +33.7$\%$ \\
& Class D & -2.25 & +34.1$\%$ \\ 
\bottomrule
\end{tabular}
\subcaption{\label{tab:qoe_table_hlvc}}
\end{subtable}
\hfill
\begin{subtable}[c]{0.32\textwidth}
\centering
\begin{tabular}[t]{cccc}
\toprule
Attacks& Dataset & PSNR (dB) & Bpp\\
\midrule
{\multirow{3}{*}{\begin{tabular}[c]{@{}c@{}} Video Quality \\(Offline)\end{tabular}}}
& Class B & -4.52 & +1.1$\%$\\ 
& Class C & -3.57 & +0.9$\%$ \\
& Class D & -4.23 &  +0.5$\%$ \\ \hline
{\multirow{3}{*}{\begin{tabular}[c]{@{}c@{}} Video Quality \\(Online)\end{tabular}}}
& Class B & -2.48 & +15.4$\%$\\ 
& Class C & -3.24 & +20.3$\%$ \\
& Class D & -2.97 & +22.4$\%$ \\ \hline
{\multirow{3}{*}{\begin{tabular}[c]{@{}c@{}} Bandwidth \\(Offline)\end{tabular}}}
& Class B & -0.01 & +111.4$\%$\\ 
& Class C & -0.01 & +103.6$\%$ \\
& Class D & -0.00 &  +98.9$\%$ \\ \hline
{\multirow{3}{*}{\begin{tabular}[c]{@{}c@{}} Bandwidth \\(Online)\end{tabular}}}
& Class B & -0.59 & +45.8$\%$\\ 
& Class C & -0.62 & +42.4$\%$ \\
& Class D & -0.55 &  +30.4$\%$ \\ \hline
{\multirow{3}{*}{\begin{tabular}[c]{@{}c@{}} RD \\(Offline)\end{tabular}}}
& Class B & -4.49 & +94.5$\%$\\ 
& Class C & -4.12 & +90.3$\%$ \\
& Class D & -4.35 & +87.4$\%$ \\ \hline
{\multirow{3}{*}{\begin{tabular}[c]{@{}c@{}} RD \\(Online)\end{tabular}}}
& Class B & -2.61 & +38.1$\%$\\ 
& Class C & -3.36 & +36.2$\%$ \\
& Class D & -3.12 & +30.0$\%$ \\ \hline
{\multirow{3}{*}{Gaussian (Case I)}}
& Class B & -1.33 & +15.3$\%$\\
& Class C & -1.64 & +17.2$\%$ \\
& Class D & -1.51 & +19.2$\%$ \\ \hline
{\multirow{3}{*}{Gaussian (Case II)}}
& Class B & -2.23 & +31.7$\%$\\
& Class C & -2.11 & +32.6$\%$ \\
& Class D & -2.37 & +33.8$\%$ \\ 
\bottomrule
\end{tabular}
\subcaption{\label{tab:qoe_table_fvc}}
\end{subtable}
\vspace{-0.5cm}
\caption{PSNR and Bpp changes after applying RoVISQ attacks on the (a) DVC~\cite{Lu_2019_CVPR}, (b) HLVC~\cite{Yang_2020_CVPR}, and (c) FVC~\cite{Hu_2021_CVPR} compression models. Reported values are averaged over different encoding parameters $\lambda\in\{256, 512,1024, 2048\}$.}
\vspace{-0.7cm}
\end{table*}%

\section{RoVISQ Attacks on Video Compression}
\label{sec:extra_exps}
We provide the numerical results of our white-box QoE attacks on three state-of-the-art video compression models in Tables~\ref{tab:qoe_table_dvc},~\ref{tab:qoe_table_hlvc}~\ref{tab:qoe_table_fvc}. The provided numbers here correspond to the scatter plots in Figure~\ref{fig:Exp_qoe_attack}. \edit{We note that a successful bandwidth attack should increase Bpp while maintaining PSNR. Similarly, the video quality attack should maintain Bpp while reducing PSNR. Therefore, as seen in Tables~\ref{tab:qoe_table_dvc},~\ref{tab:qoe_table_hlvc}~\ref{tab:qoe_table_fvc}, RoVISQ outperforms Gaussian noise as the noise cannot target only one of the PSNR/Bpp metrics. When targeting both Bpp and PSNR, our RD attack still outperforms Gaussian noise by a large margin.} The overall better performance of RoVISQ attacks on HLVC and FVC compared to DVC for both offline and online scenarios can be attributed to the following reasons: (1)~HLVC and FVC use post-processing techniques based on the adversarial decoded frames at the end of the video decoder. (2)~HLVC and FVC employ hierarchical temporal coding structures as mentioned in Section~\ref{sec:system}. By exploiting temporal information form multiple perturbed reference frames, the hierarchical structure becomes more vulnerable to error than non-hierarchical structure.

\noindent\textbf{Unseen Victim Models for Black-box Attack.} \edit{Table~\ref{tab:substitute_models} encloses the architectural details of victim video compression models used in our black-box attack evaluations (see Section~\ref{sec:qoe_attacks}, Table~\ref{tab:blackbox_qoe}). New models are constructed by altering the architecture of a state-of-the-art video compression model, e.g., DVC~\cite{Lu_2019_CVPR}, denoted by ``Template Model'' in Table~\ref{tab:substitute_models}. Each model uses the GOP structure of the corresponding template model. The new models are then trained from scratch following the training setup of their template model. We train four variations for each new model using different encoding parameters $\lambda \in \{256,512,1024,2048\}$.}

\begin{table}[h]
\centering
\caption{\edit{Architecture details of victim video compression models used in the black-box evaluations of Table~\ref{tab:blackbox_qoe}. We use the $n_1 \Rightarrow n_2$ notation where $n_1$ is the number of layers/kernels for the corresponding module in the template model and $n_2$ is the altered number of layers/kernels in the new victim model.}}
\label{tab:substitute_models}
\resizebox{\columnwidth}{!}{
\begin{tabular}{llcccccc}
\toprule
& & M1 
& M2 
& M3 
& M4 
& M5 
& M6 \\
\midrule
\multicolumn{2}{r}{Template Model}
& DVC~\cite{Lu_2019_CVPR}
& DVC~\cite{Lu_2019_CVPR}
& HLVC~\cite{Yang_2020_CVPR}
& HLVC~\cite{Yang_2020_CVPR} 
& FVC~\cite{Hu_2021_CVPR} 
& FVC~\cite{Hu_2021_CVPR} \\
\midrule
\multirow{2}{*}{\begin{tabular}[c]{@{}l@{}}Motion \\ Estimation\end{tabular}}
& \# Layers & 20 $\Rightarrow$ 15 & 20 $\Rightarrow$ 25 & 25 $\Rightarrow$ 20 & 25 $\Rightarrow$ 30 & 2 & 2 $\Rightarrow$ 4 \\ 
& \# Kernels & 64 $\Rightarrow$ 56 & 64 $\Rightarrow$ 72& 64 $\Rightarrow$ 56 & 64 $\Rightarrow$ 72& 64 $\Rightarrow$ 56& 64 $\Rightarrow$ 72 \\ 
\midrule
\multirow{2}{*}{\begin{tabular}[c]{@{}l@{}}Motion \\ Compensation\end{tabular}}
& \# Layers & 12 $\Rightarrow$ 10 & 12 $\Rightarrow$ 14 & 12 $\Rightarrow$ 10 & 12 $\Rightarrow$ 14 & 3 & 3 $\Rightarrow$ 5 \\ 
& \# Kernels & 64 $\Rightarrow$ 56 & 64 $\Rightarrow$ 72& 64 $\Rightarrow$ 56 & 64 $\Rightarrow$ 72& 64 $\Rightarrow$ 56& 64 $\Rightarrow$ 72 \\
\midrule
\multirow{2}{*}{\begin{tabular}[c]{@{}l@{}}Motion \\ Compression\end{tabular}}
& \# Layers & 14 $\Rightarrow$ 12 & 14 $\Rightarrow$ 16 & 8 & 8 $\Rightarrow$ 10 & 42 $\Rightarrow$ 36 & 42 $\Rightarrow$ 48 \\ 
& \# Kernels & 128 $\Rightarrow$ 120 & 128 $\Rightarrow$ 136& 128 $\Rightarrow$ 120 & 128 $\Rightarrow$ 136& 128 $\Rightarrow$ 120& 128 $\Rightarrow$ 136 \\
\midrule
\multirow{2}{*}{\begin{tabular}[c]{@{}l@{}}Residual \\ Compression\end{tabular}}
& \# Layers & 8 $\Rightarrow$ 6 & 8 $\Rightarrow$ 10 & 8 & 8 $\Rightarrow$ 10 & 42 $\Rightarrow$ 36 & 42 $\Rightarrow$ 48 \\ 
& \# Kernels & 64 $\Rightarrow$ 56 & 64 $\Rightarrow$ 72& 128 $\Rightarrow$ 120 & 128 $\Rightarrow$ 136& 128 $\Rightarrow$ 120& 128 $\Rightarrow$ 136 \\
\bottomrule
\end{tabular}}
\end{table}

\vspace{-0.3cm}
\section{RoVISQ Attacks on Video Classification}\label{sec:appdx_classification_attack}
\begin{figure*}[h]
\centering
  \begin{tabular}{@{}c@{}}
    \includegraphics[width=0.6\linewidth]{Fig/Fig_QoE/Exp_asu_bar.png} \\  
  \end{tabular}
    \begin{tabular}{@{}cc@{}}
    \includegraphics[width=0.4\linewidth]{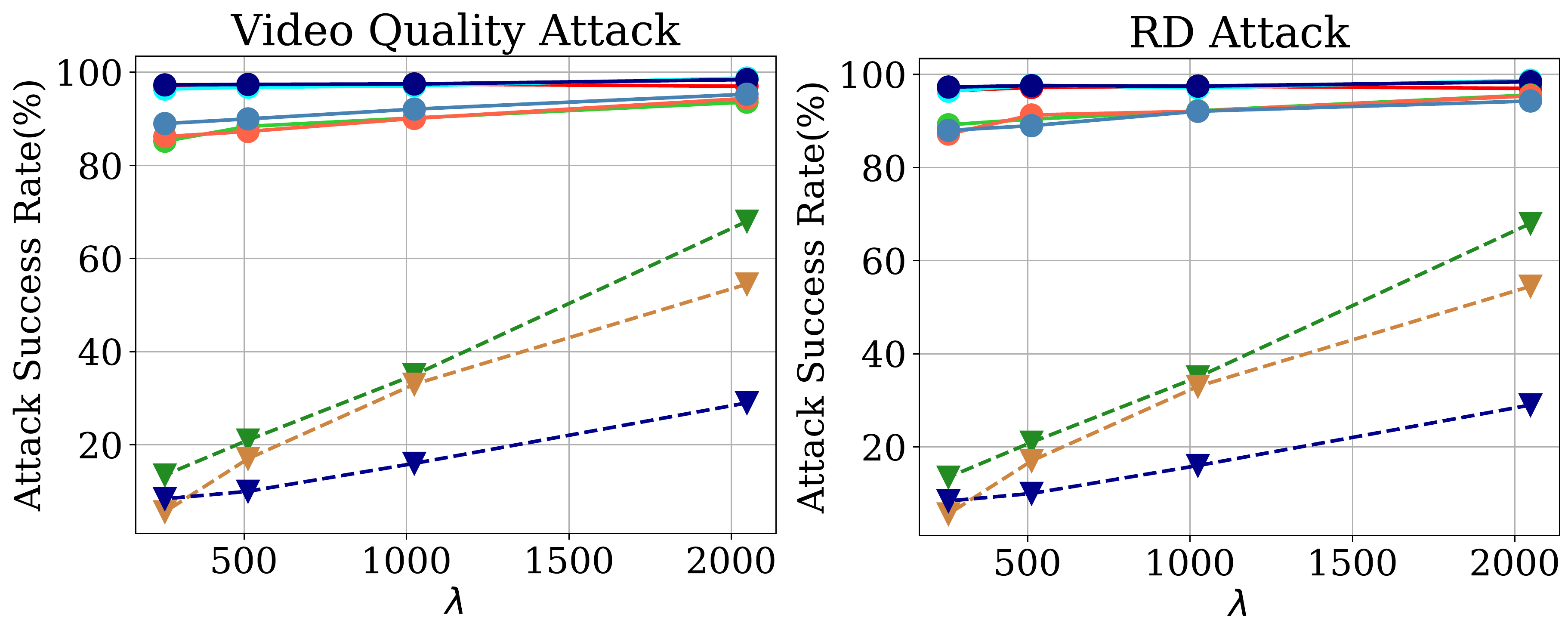} & 
    \includegraphics[width=0.4\linewidth]{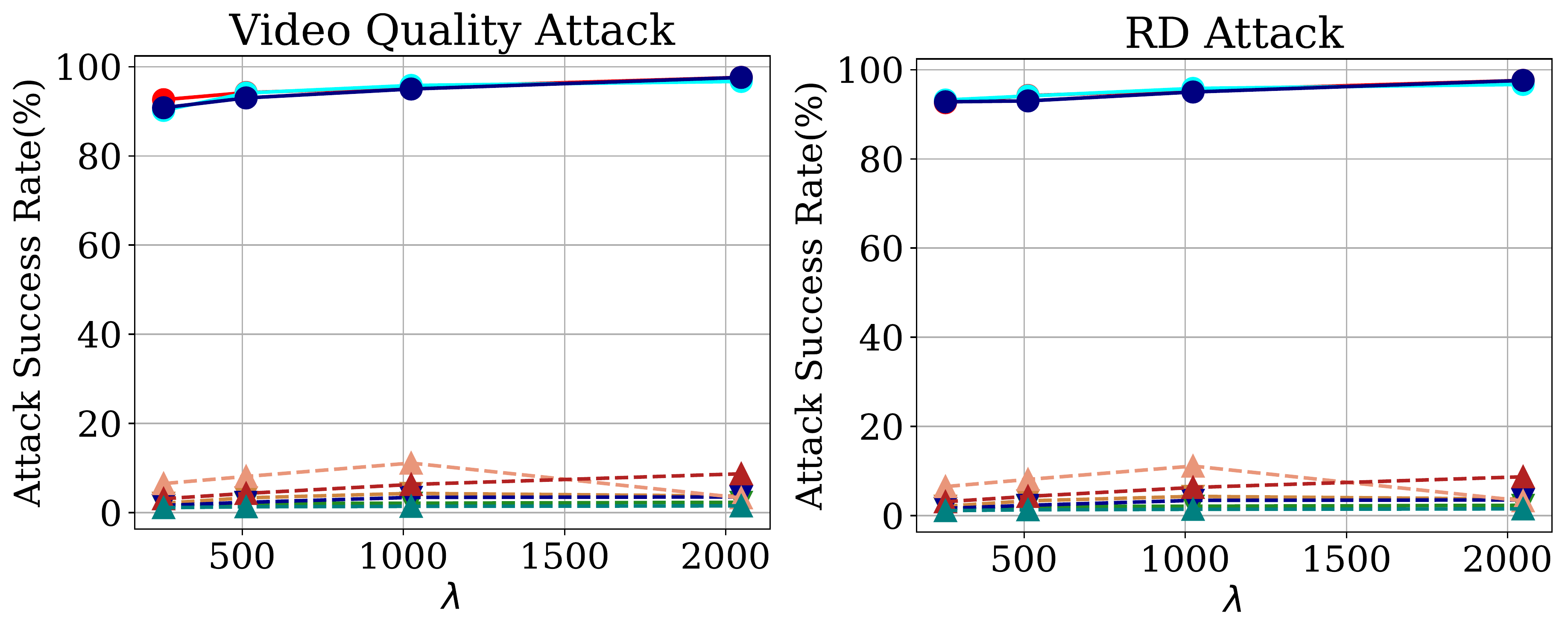} \\
    (a) Untargeted Attack&
    (b) Targeted Attack \\
    \end{tabular}
  \caption{\edit{Attack success rate of our compression-robust classifier perturbations compared to state-of-the-art adversarial attacks on video classification, i.e., SAP~\cite{Wei_aaai_2019} and C-DUP~\cite{Li_2019_NDSS}. Attacks are conducted in the white-box scenario on three video classification models, i.e., I3D \cite{Carreira_2017_CVPR}, SlowFast \cite{Feichtenhofer_2019_ICCV}, and TPN \cite{TPN_2020_CVPR} using the Jester dataset.}
}
  \label{fig:Exp_asr_vcs_vqrd}
\end{figure*}

\edit{Below we provide the success rate of our video quality and \rdos when applied to a downstream video classifier. Similar to the analysis performed in Section~\ref{sec:exps_classification}, we evaluate our attacks in the white-box and black-box settings in Figure~\ref{fig:Exp_asr_vcs_vqrd} and Table~\ref{tab:appdx_classification_blackbox}, respectively. All attacks in this section are conducted on the Jester activity recognition dataset. For each attack scenario, we compare our attack success rate with the prior state-of-the-art adversarial attacks for video classification. Overall, similar trends can be observed as those in Section~\ref{sec:exps_classification}: RoVISQ perturbations maintain a high attack success rate in both white-box and black-box scenarios while the effect of other adversarial attacks is removed after video compression. Notably, \rdo obtains the best attach success rate. This is intuitive since our formulation of the bandwidth and video quality attacks enforce constraints on the perturbations to ensure only one QoE factor is changed (see Equations~\ref{eq:video_quality_optimization_problem},~\ref{eq:bandwidth_optimization_problem}). Our formulation for \rdo, however, does not have such constraints (see Equation~\ref{eq:rdo_optimization_problem}) and therefore has more flexibility in finding a more effective perturbations.}

\begin{figure*}[h]
\centering
  \begin{tabular}{@{}c@{}}
    \includegraphics[width=\linewidth]{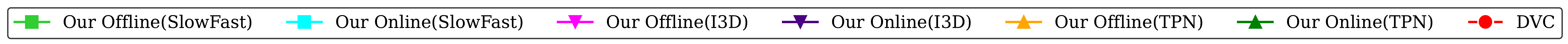} \\  
  \end{tabular}
    \begin{tabular}{@{}ccc@{}}
    \includegraphics[width=0.31\linewidth]{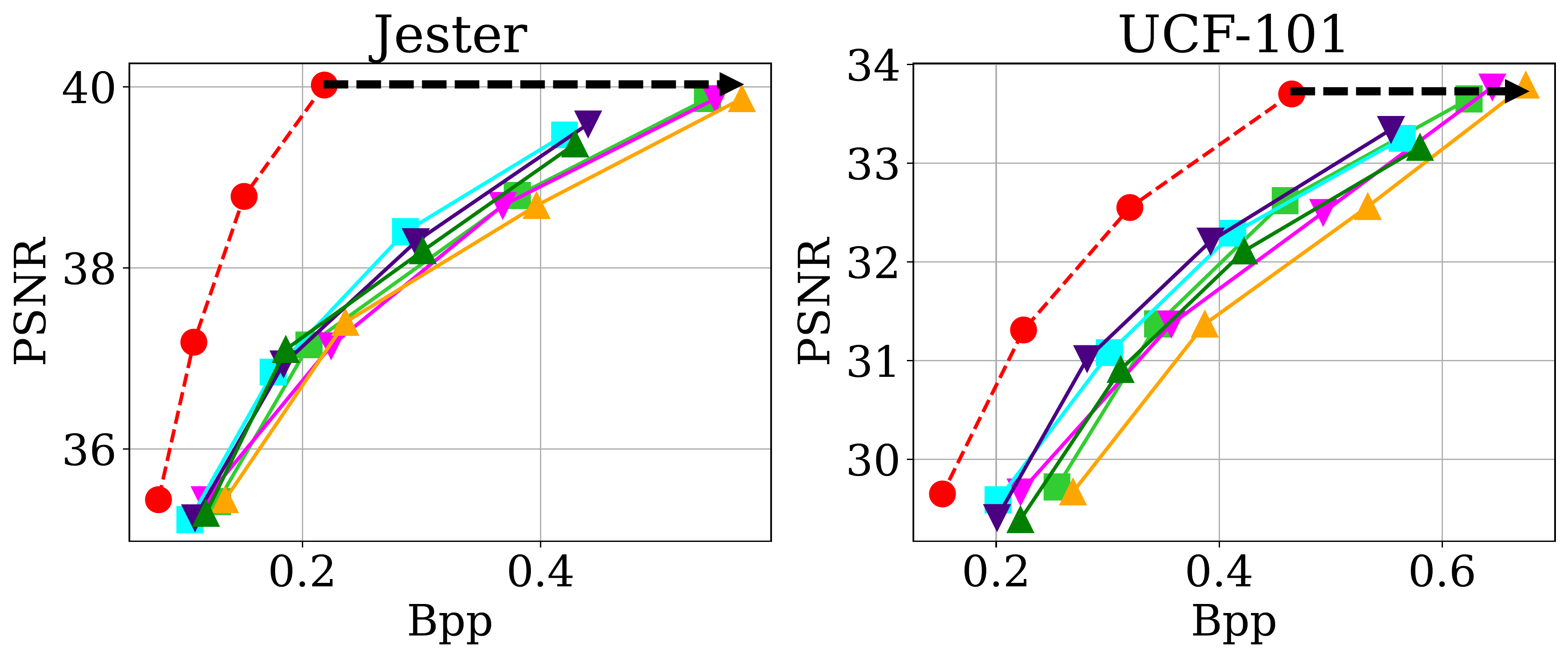}   &
    \includegraphics[width=0.31\linewidth]{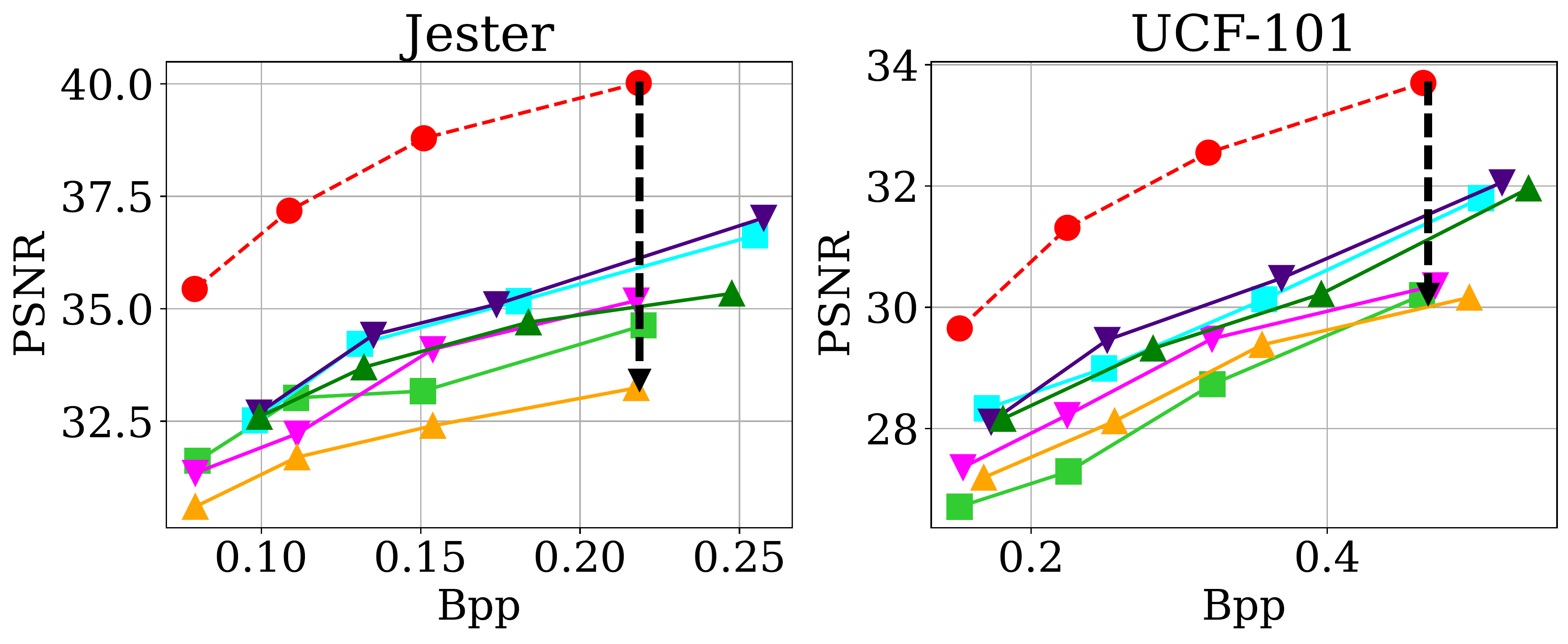} &
    \includegraphics[width=0.31\linewidth]{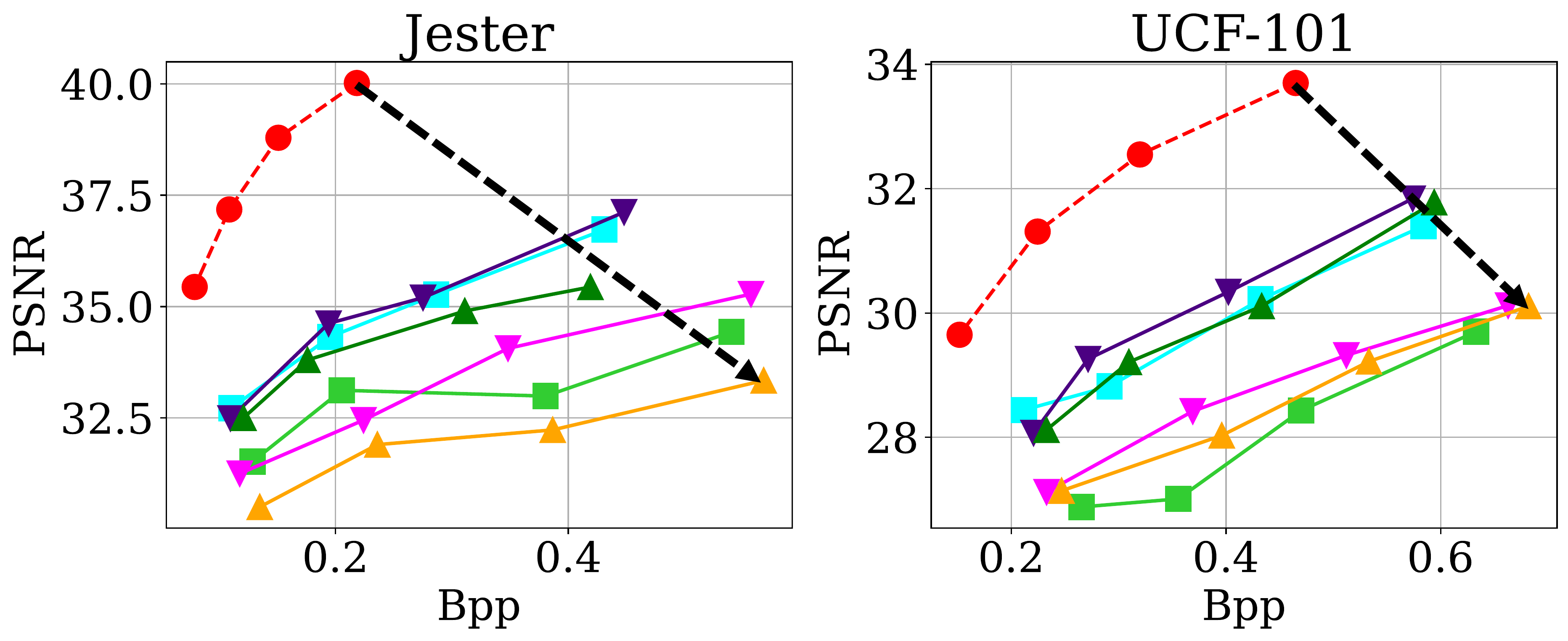} \\
    \footnotesize{Bandwidth Attack}&
    \footnotesize{Video Quality Attack}&
    \footnotesize{RD Attack} \\
    \end{tabular}
  \caption{Experimental results of RoVISQ attacks on video compression and classification systems when tested on the UCF-101 and Jester test sets. Each graph contains the results of video compression for the case where $\lambda \in \{256,512,1024,2048\}$.}
  \label{fig:Exp_qoe_on_vcs}
 \vspace{-0.5cm}
\end{figure*}

\noindent\textbf{Effect on QoE.} Our proposed attacks on downstream video classifiers also affect the \textit{R}-\textit{D} curve of the video compression system. In Figure~\ref{fig:Exp_qoe_on_vcs} we show the effect of our attacks for video compression and classification on the QoE of the DVC video coder. We observe that our adversarial attacks on video compression and classification can selectively deteriorate the benign compressor's $R$ and $D$ as explained in Section \ref{sec:qoe_attacks}. In the example of the Jester dataset, our bandwidth attack increases the Bpp by up to $2.4\times$ while changing the video quality by only $0.1$dB. Our video quality attack lowers the PSNR by $4.84$dB while changing the Bpp by only $1.47\%$. Our \rdo reduces the PSNR by $4.96$dB and increases Bpp by $2.2\times$. Although the online universal attack has lower performance than our offline attack, it lowers PSNR by up to $3.54$dB and increases Bpp by up to $101.42\%$ on Jester dataset.

\begin{table}[h]
\centering
\caption{\edit{RoVISQ video quality and \rdos compared with prior work on adversarial video classification. Attacks are conducted in the black-box untargeted scenario on the Jester dataset. The names inside parentheses in the attack column are the surrogate video classifiers used to train the RoVISQ universal perturbations.}}\label{tab:appdx_classification_blackbox}
\resizebox{0.9\columnwidth}{!}{
\begin{tabular}{clclll}
\toprule
\multicolumn{1}{c}{\multirow{2}{*}{\begin{tabular}[c]{@{}c@{}}Victim\\ Model\end{tabular}}}
& \multicolumn{1}{c}{\multirow{2}{*}{Attack}}
& \multicolumn{4}{c}{Attack Success Rate (\%)} \\ \cline{3-6}
&                                   
& \multicolumn{1}{c}{$\lambda=256$} 
& \multicolumn{1}{c}{$512$} 
& \multicolumn{1}{c}{$1024$} 
& \multicolumn{1}{c}{$2048$} \\ \hline
\multirow{6}{*}{\begin{tabular}[c]{@{}c@{}}
TPN\\ \cite{TPN_2020_CVPR}
\end{tabular}}  
& GeoTrap~\cite{Li_NIPS_2021}
& 6.5                                
& 16.7
& 18.5
& 32.4  \\ \cline{2-6}
& U3D~\cite{Xie_SP_2022}
& 7.4                                
& 17.6                                
& 19.4
& 36.1 \\ \cline{2-6}
& Video Quality (I3D)                   
& 74.1                       
& 77.8                        
& 82.4                        
& \textbf{87.0}\\
& Video Quality (SlowFast)                 
& \textbf{75.9}                       
& \textbf{78.7}                        
& \textbf{83.3}                        
& 85.1\\ \cline{2-6}
& RD (I3D)  
& 77.8                      
& \textbf{83.3}                       
& \textbf{86.1 }                     
& 90.7\\ 
& RD (SlowFast)
& \textbf{78.7}                      
& 80.6                       
& 82.4                      
& \textbf{91.6}\\ \hline
\multirow{6}{*}{\begin{tabular}[c]{@{}c@{}}
SlowFast\\ \cite{Feichtenhofer_2019_ICCV}
\end{tabular}} 
& GeoTrap~\cite{Li_NIPS_2021}
& 11.1                                
& 22.2
& 38.9
& 54.6  \\ \cline{2-6}
& U3D~\cite{Xie_SP_2022}
& 10.2                                
& 24.1                                
& 37.0
& 60.2 \\ \cline{2-6}
& Video Quality (I3D)                        
& 73.2                        
& \textbf{79.6}                      
& 81.5                      
& 85.2\\
& Video Quality (TPN)
& \textbf{75.9}                        
& \textbf{79.6}                      
& \textbf{83.3}                      
& \textbf{86.1}\\ \cline{2-6}
& RD (I3D)                                          
& \textbf{79.6}                    
& 81.5                    
& 84.3                    
& 88.0\\
& RD (TPN)                 
& 78.7                    
& \textbf{83.3}                    
& \textbf{86.1}                    
& \textbf{89.8}\\ \hline
\multirow{6}{*}{\begin{tabular}[c]{@{}c@{}}
I3D\\ \cite{Carreira_2017_CVPR}
\end{tabular}} 
& GeoTrap~\cite{Li_NIPS_2021}
& 8.3                                
& 23.1
& 41.7
& 43.5  \\ \cline{2-6}
& U3D~\cite{Xie_SP_2022}
& 6.5                                
& 16.7                                
& 19.4
& 36.1 \\ \cline{2-6}
& Video Quality (SlowFast)    
& \textbf{76.9}           
& \textbf{83.3}              
& \textbf{87.0}               
& \textbf{89.8}\\
& Video Quality (TPN) 
& 74.1             
& 78.7              
& 80.6               
& 84.3\\ \cline{2-6}
& RD (SlowFast)                          
& 78.7                        
& \textbf{82.4}                       
& \textbf{87.0}                     
& \textbf{90.7} \\
& RD (TPN)                       
& \textbf{80.6}                        
& \textbf{82.4}                       
& 84.3                     
& 88.9 
\\ \bottomrule    
\end{tabular}}
\end{table}

\section{Convergence Analysis of the Proposed Attacks}\label{sec:convergence}
\begin{figure}[h]
\centering
    \begin{subfigure}[b]{\columnwidth}
        \centering
        \includegraphics[width=0.93\linewidth]{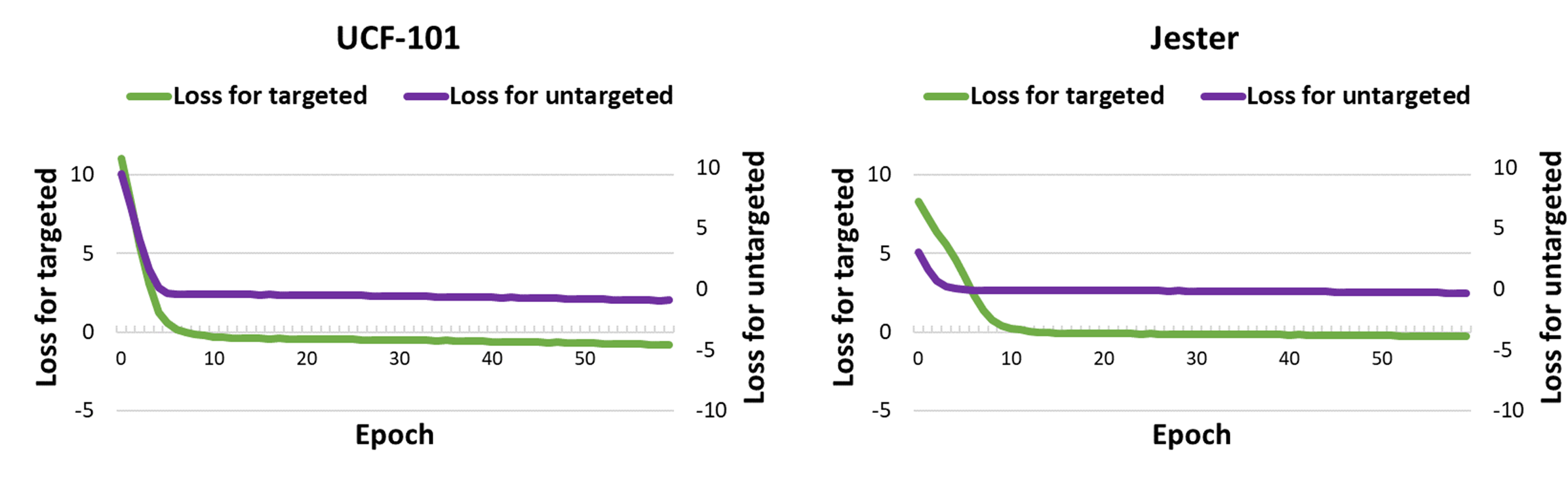}
        \vspace{-0.2cm}
        \caption{Video Quality Attack}
    \end{subfigure}
    \vspace{-0.2cm}
    \begin{subfigure}[b]{\columnwidth}
        \centering
        \includegraphics[width=0.93\linewidth]{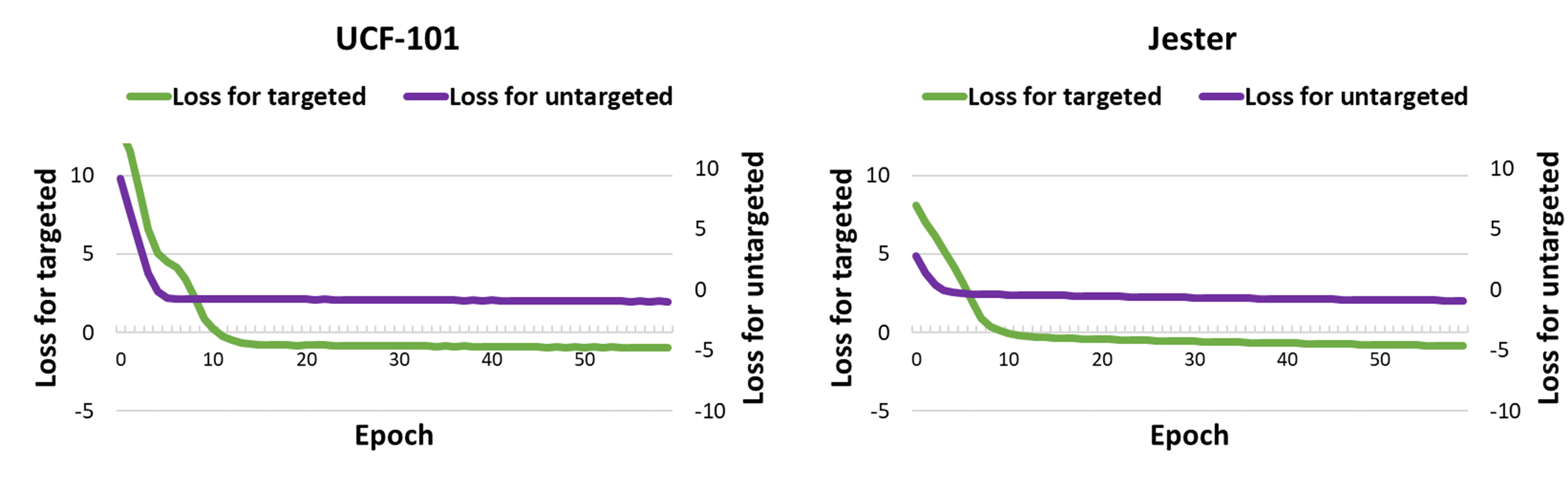}
        \vspace{-0.2cm}
        \caption{Bandwidth Attack}
    \end{subfigure}
    \vspace{-0.2cm}
    \begin{subfigure}[b]{\columnwidth}
        \centering
        \includegraphics[width=0.93\linewidth]{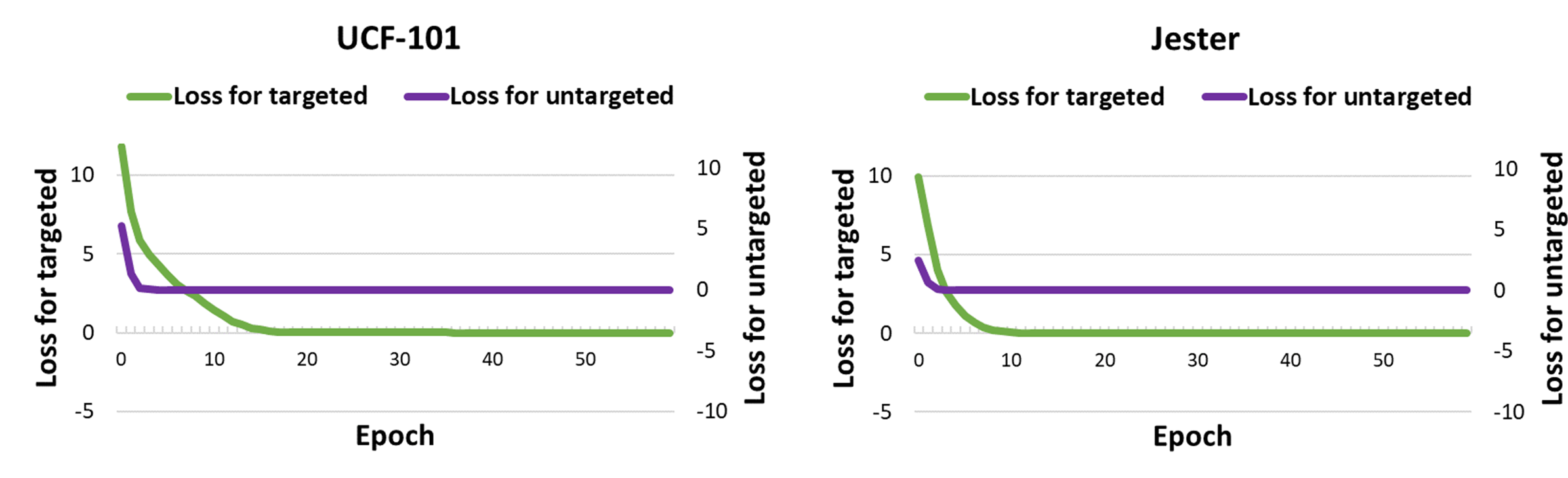}
        \vspace{-0.2cm}
        \caption{RD Attack}
    \end{subfigure}
    \vspace{-0.5cm}
  \caption{Convergence curves of RoVISQ attacks on video classification when optimizing the loss in Section~\ref{sec:white-box_construction}.}
  \label{fig:Exp_qoe_on_vcs_appdx}
\end{figure}

The convergence curves of perturbation generation using our losses for targeted and untargeted attacks are demonstrated in Figure~\ref{fig:Exp_qoe_on_vcs_appdx}. As seen, both losses converge rapidly, and achieve $\sim 100\%$ attack success rate with $20$ FGSM iterations. There is a slight difference in the convergence speed among the bandwidth, video quality, and RD attacks. Since the video quality and RD attacks inject perturbations that downgrade image quality, their overall convergence speed is a bit faster compared to the bandwidth attack.

\begin{table}[h]
\centering
\caption{PSNR and Bpp of adversarially trained (AT)\cite{madry2017towards} DVC on the HEVC class B and C Datasets.} \label{tab:qoe_defense_dvc_table}
\resizebox{0.9\columnwidth}{!}{
\begin{tabular}[t]{cccccc}
\toprule
\multirow{2}{*}{Benchmark} 
& \multirow{2}{*}{Dataset} 
& \multicolumn{2}{c}{w Defense}
& \multicolumn{2}{c}{w/o Defense} \\
& & PSNR (dB) & Bpp
& PSNR (dB) & Bpp \\
\midrule
{\multirow{2}{*}{DVC~\cite{Lu_2019_CVPR}}}
& Class B & 31.14 & 0.20 & 33.52 & 0.16 \\
& Class D & 29.28 & 0.33 & 31.32 & 0.28 \\
\midrule
{\multirow{2}{*}{\begin{tabular}[l]{@{}c@{}} Video Quality \\(Offline)\end{tabular}}}
& Class B & -2.61 & +0.8$\%$ & -3.85 & +0.9$\%$ \\ 
& Class D & -2.88 & +0.7$\%$ & -3.94 & +0.7$\%$ \\ \hline
{\multirow{2}{*}{\begin{tabular}[c]{@{}c@{}} Video Quality \\(Online)\end{tabular}}}
& Class B & -1.83 & +14.8$\%$ & -2.30 & +13.6$\%$ \\ 
& Class D & -2.13 & +17.7$\%$ & -2.82 & +20.3$\%$ \\ \hline
{\multirow{2}{*}{\begin{tabular}[c]{@{}c@{}} Bandwidth \\(Offline)\end{tabular}}}
& Class B & -0.13 & +73.9$\%$ & -0.00 & +87.5$\%$ \\ 
& Class D & -0.16 & +77.6$\%$ & -0.00 & +87.5$\%$ \\ \hline
{\multirow{2}{*}{\begin{tabular}[c]{@{}c@{}} Bandwidth \\(Online)\end{tabular}}}
& Class B & -0.68 & +32.7$\%$ & -0.50 & +35.3$\%$ \\ 
& Class D & -0.58 & +28.3$\%$ & -0.18 & +28.7$\%$ \\  \hline
{\multirow{2}{*}{\begin{tabular}[c]{@{}c@{}} RD \\(Offline)\end{tabular}}}
& Class B & -3.13 & +72.4$\%$ & -4.03 & +89.1$\%$ \\ 
& Class D & -3.38 & +72.5$\%$ & -4.38 & +81.2$\%$ \\ \hline
{\multirow{2}{*}{\begin{tabular}[c]{@{}c@{}} RD \\(Online)\end{tabular}}}
& Class B & -1.81 & +26.2$\%$ & -2.38 & +30.3$\%$ \\ 
& Class D & -2.76 & +21.7$\%$ & -3.41 & +25.3$\%$ \\ 
\bottomrule
\end{tabular}}
\end{table}%

\begin{table}[h]
\centering
\caption{Accuracy (ACC) of video classification on clean videos, along with the attack success rate (ASR) of compression-robust classifier perturbations. The defense is adversarial training on the Jester dataset. Here, ``T'' and ``U'' denote targeted and untargeted attacks, respectively. } \label{tab:jester_at_defense}
\resizebox{\columnwidth}{!}{
\begin{tabular}[t]{ccccc|cc}
\toprule
\begin{tabular}[c]{@{}c@{}}Video\\ Classifier\end{tabular}  
& Type
& Attack
& \begin{tabular}[c]{@{}c@{}}ASR (\%)\\ w Defense\end{tabular}
& \begin{tabular}[c]{@{}c@{}}ASR (\%)\\ w/o Defense\end{tabular}
& \begin{tabular}[c]{@{}c@{}}ACC (\%)\\ w Defense\end{tabular} 
& \begin{tabular}[c]{@{}c@{}}ACC (\%)\\ w/o Defense\end{tabular}
\\
\midrule
\multirow{3}{*}{\begin{tabular}[c]{@{}c@{}}SlowFast\\ \cite{Feichtenhofer_2019_ICCV}\end{tabular}}
& T & Offline & 71.1 & 94.2 & \multirow{3}{*}{73.3} & \multirow{3}{*}{89.5}\\
& U & Offline & 72.5 & 96.8 & & \\
& U & Online & 64.3 & 83.6 & & \\ \hline
\multirow{3}{*}{\begin{tabular}[c]{@{}c@{}}TPN\\ \cite{TPN_2020_CVPR}\end{tabular}}
& T & Offline & 70.7 & 93.0 & \multirow{3}{*}{80.3} & \multirow{3}{*}{90.5} \\ 
& U & Offline & 75.5 & 97.3 & & \\
& U & Online & 68.3 & 82.6 & & \\ \hline
\multirow{3}{*}{\begin{tabular}[c]{@{}c@{}}I3D\\ \cite{Carreira_2017_CVPR}\end{tabular}}
& T & Offline & 65.3 & 92.1 & \multirow{3}{*}{82.9} & \multirow{3}{*}{91.2} \\ 
& U & Offline & 66.7 & 97.0 & & \\
& U & Online & 56.8 & 86.5 & & \\
\bottomrule
\end{tabular}}
\end{table}

\end{appendices}



%

\end{document}